\def\PatasExtended{}
\documentclass[conference]{IEEEtran}

\usepackage{array}

\usepackage{textcomp}

\usepackage{url}
\usepackage{verbatim}
\usepackage{graphicx}
\usepackage{cite}
\usepackage{booktabs}
\usepackage{todonotes}
\usepackage{amsmath,amssymb,amsthm}
\usepackage{graphicx}
\usepackage{caption}
\usepackage{subcaption}
\usepackage{enumitem}
\usepackage{algorithm}
\usepackage{algpseudocode}
\algrenewcommand\algorithmicfunction{\textbf{Function}}
\usepackage{tikz}
\usetikzlibrary{arrows.meta,positioning}
\newcommand{\tnopfigure}{%
\begin{tikzpicture}[font=\footnotesize,
  box/.style={fill=black!10, minimum width=1.55cm, minimum height=0.85cm, align=center},
  node circle/.style={circle, fill=black!5, draw=black!30, minimum size=1.05cm, align=center},
  flow/.style={-{Triangle[width=5pt,length=5pt]}, line width=2.2pt, black!20}]
  \node[box] (in) {Trust in\\ $\boldsymbol{x^{(l)}}$};
  \node[node circle, right=0.55cm of in] (tn) {\textbf{Trust}\\ \textbf{Node}};
  \node[box, right=0.55cm of tn] (out) {Trust in\\ $\boldsymbol{x^{(l+1)}}$};
  \draw[flow] (in) -- (tn);
  \draw[flow] (tn) -- (out);
  \node[below=0.22cm of tn, align=left] {\textbf{TF:} $T_{z^{(l)}} = \bigoplus_{i=1}^{d} T_{\theta_i^{(l)}} \otimes\, T_{x_i^{(l)}}$\\[1pt]
  \phantom{\textbf{TF:} }$T_{x^{(l+1)}} = T_{\phi^{(l)}}(T_{z^{(l)}})$};
\end{tikzpicture}}
\newif\ifextended
\ifdefined\PatasExtended\extendedtrue\fi
\ifextended\else
  \usepackage{xr}
\fi
\usepackage{hyperref}
\usepackage[capitalise,nameinlink]{cleveref}
\crefname{figure}{Fig.}{Figs.}
\Crefname{figure}{Fig.}{Figs.}
\crefname{table}{Table}{Tables}
\Crefname{table}{Table}{Tables}
\crefname{equation}{Eq.}{Eqs.}
\Crefname{equation}{Eq.}{Eqs.}
\crefname{algorithm}{Alg.}{Algs.}
\Crefname{algorithm}{Alg.}{Algs.}
\crefname{section}{Section}{Sections}
\Crefname{section}{Section}{Sections}
\crefname{appendix}{Appendix}{Appendices}
\Crefname{appendix}{Appendix}{Appendices}

\newcommand{\appref}[1]{\ifextended Appendix~\ref{#1}\else Appendix~\ref*{ext-#1} of~\cite{patasext}\fi}
\newcommand{\apprefs}[2]{\ifextended Appendices~\ref{#1} and~\ref{#2}\else Appendices~\ref*{ext-#1} and~\ref*{ext-#2} of~\cite{patasext}\fi}
\newcommand{\xcref}[1]{\ifextended\cref{#1}\else\cref*{ext-#1} of~\cite{patasext}\fi}
\newcommand{\xcrefin}[2]{\ifextended\cref{#1} in Appendix~\ref{#2}\else\cref*{ext-#1} in Appendix~\ref*{ext-#2} of~\cite{patasext}\fi}
\newcommand{\xref}[1]{\ifextended\ref{#1}\else\ref*{ext-#1} of~\cite{patasext}\fi}

\newcommand{\pf}{\mathit{Pf}}

\newcommand{\suppp}{Supplementary Materials}

\newtheorem{theorem}{Theorem}
\newtheorem{definition}{Definition}
\newtheorem{lemma}{Lemma}

\theoremstyle{remark}

\theoremstyle{definition}

\setlist{noitemsep,topsep=2pt,leftmargin=*}
\begin{document}

\title{PaTAS: A Framework for Trust Propagation in Neural Networks Using Subjective Logic}

\author{
\IEEEauthorblockN{Koffi Ismael Ouattara\IEEEauthorrefmark{1},
Ioannis Krontiris\IEEEauthorrefmark{2},
Theo Dimitrakos\IEEEauthorrefmark{3},
Dennis Eisermann\IEEEauthorrefmark{4},
Houda Labiod\IEEEauthorrefmark{5},
and Frank Kargl\IEEEauthorrefmark{4}}
\IEEEauthorblockA{\IEEEauthorrefmark{1}Huawei Heisenberg Research Center, Munich, Germany\\
\IEEEauthorrefmark{2}Bavarian University of Business and Technology, Munich, Germany\\
\IEEEauthorrefmark{3}School of Computing, University of Kent, Canterbury, UK\\
\IEEEauthorrefmark{4}Institute of Distributed Systems, Ulm University, Ulm, Germany\\
\IEEEauthorrefmark{5}Huawei Fourier Research Center, Paris, France}
}

\maketitle

\begin{abstract}
\ifextended% IEEE notice for the posted accepted version (arXiv build only)
\begin{tikzpicture}[remember picture,overlay]
\node[anchor=north,yshift=-0.35cm,text width=0.86\paperwidth,align=center,font=\normalfont\scriptsize] at (current page.north)
{\copyright~2026 IEEE. Personal use of this material is permitted. Permission from IEEE must be obtained for all other uses, in any current or future media, including reprinting/republishing this material for advertising or promotional purposes, creating new collective works, for resale or redistribution to servers or lists, or reuse of any copyrighted component of this work in other works. Extended version, with appendices, of the paper accepted at IEEE TPS 2026.};
\end{tikzpicture}%
\fi
Trustworthiness has become a key requirement for deploying artificial intelligence in safety-critical applications, yet conventional metrics such as accuracy fail to capture uncertainty or the reliability of predictions, particularly under adversarial or degraded conditions. This paper introduces the \emph{Parallel Trust Assessment System (PaTAS)}, a framework for modeling and propagating trust in neural networks using Subjective Logic (SL). PaTAS operates in parallel with standard neural computation through \emph{Trust Nodes} and \emph{Trust Functions} that propagate input, parameter, and activation trust across the network, refining parameter trust during training (\emph{Parameter Trust Update}) and returning a per-inference trust opinion, quantifying belief, disbelief, and uncertainty, on each prediction (\emph{Inference-Path Trust Assessment}). Experiments on real-world and adversarial datasets show that these estimates are interpretable and stable and complement accuracy: PaTAS flags predictions that remain accurate yet rely on parameters learned from corrupted data, detects adversarially patched inputs without knowledge of the trigger, and audits training-label corruption without any test data. Compared with output-only uncertainty baselines and dedicated input- and feature-space detectors, out-of-distribution and corruption detection is carried by the conformity-derived input opinions PaTAS consumes, within a regime that a per-dataset diagnostic identifies in advance; trust propagation adds model-side signals these detectors lack, and the composed trust degrades gracefully where any single signal fails. The mechanism extends to convolutional networks. PaTAS thereby provides a foundation for transparent, quantifiable trust reasoning across the AI lifecycle.
\end{abstract}


\vspace{-0.5em}
\section{Introduction}

Artificial Intelligence (AI) systems, particularly Neural Networks (NNs), are employed in critical sectors such as healthcare and autonomous driving. While highly performant, they often operate as black boxes offering limited insight into the reliability of their outputs, an opacity that becomes critical under adversarial, uncertain, or degraded conditions. Conventional metrics such as accuracy or precision do not capture uncertainty or reliability and can mislead decision-makers~\cite{10.1145/3290605.3300509}: a model may report 95\% accuracy while evaluated on mislabeled data. Reliability measures must therefore go beyond predictive confidence and consider the quality and provenance of inputs, the reliability of training data, and the fit of learned parameters.

As AI systems become integral to critical applications, \emph{trustworthiness} has emerged as a fundamental requirement: a trustworthy system must be lawful, ethical, and robust~\cite{ec2019ethics}, properties partly measurable as accuracy, robustness, fairness, and explainability~\cite{10.3389/fdata.2024.1467222}, and embedding them across the AI pipeline is essential for preventing failures caused by poor data quality, bias, or unstable training. Yet most models still represent confidence poorly: trained under the assumption of clean, unambiguous labels, neural networks produce predicted probabilities that reflect similarity to seen patterns rather than genuine uncertainty, and that presuppose clean, in-distribution inference inputs, assumptions rarely satisfied in real-world or adversarial settings. Existing approaches, including attribution-based methods such as SmoothGrad~\cite{smilkov2017smoothgradremovingnoiseadding}, lack mechanisms to evaluate how trustworthiness in training inputs, activations, and parameters jointly affects output reliability, motivating tools that reason explicitly about trustworthiness across the entire model lifecycle.

\paragraph*{Problem Statement}
Existing methods for trustworthiness and uncertainty estimation in neural networks \emph{neglect data provenance and quality}, assuming the training data are fully reliable; offer only \emph{limited holistic propagation}, assessing reliability at the output layer alone; and \emph{lack interpretability}, rarely yielding estimates both faithful to the model's reasoning and understandable to end users. Together these hinder reliable trustworthiness assessment in safety-critical or adversarial contexts.

\paragraph*{Contributions}
To address these challenges, we propose the \emph{Parallel Trust Assessment System (PaTAS)}, a framework for modeling and propagating trustworthiness in NNs using Subjective Logic (SL). Our contributions are:
\begin{enumerate}
    \item \emph{Parallel Trust Computation:} \textit{Trust Nodes} and \textit{Trust Functions} mirroring neural computations to propagate trust through SL discounting and fusion.
    \item \emph{Parameter Trust Update:} an algorithm deriving trust in learned parameters from gradient values, input trust, and label trust, aligned with the learning dynamics.
    \item \emph{Inference-Path Trust Assessment (IPTA):} a context-aware trust function using activation-path information for per-instance trust scores.
    \item \emph{Empirical Validation:} evaluation on real-world and adversarial datasets showing interpretable trust estimates that expose reliability gaps under poisoned, biased, and uncertain data.
\end{enumerate}

\vspace{-0.5em}
\section{Background}\label{sec:back}
Trust assessment in NNs requires a formalization that can represent ambiguity, inadequate evidence, and source reliability.
\vspace{-0.5em}
\subsection{Subjective Logic Fundamentals}\label{sec:backsl}
Classical probability cannot represent missing knowledge (ignorance) or contradictory information. Subjective Logic~\cite{josang2016subjective} extends Dempster–Shafer theory~\cite{dempster1967upper,shafer1976mathematical} by representing trust as opinions whose belief and uncertainty are separate components, thereby distinguishing epistemic from aleatoric uncertainty.

A subjective opinion \(\omega_X^A\) expresses the beliefs of an agent \(A\) (e.g., a sensor, a human, or an external observer) about states of a variable $X$ over a domain $\mathbb{X}$. For a binary variable $X \in \mathbb{X} = \{x, \overline{x}\}$, a \emph{binomial} opinion is the quadruple \(\omega_x = \omega_{X=x} = (b_x, d_x, u_x, a_x)\) satisfying $b_x + d_x + u_x = 1$, where $b_x$ denotes belief in $x$, $d_x$ disbelief in $x$ (belief in \(\overline{x}\)), $u_x$ the uncertainty mass, and $a_x$ the base rate (prior probability of $x$ absent evidence). Its projected probability \(P(x) = b_x + a_x u_x\) reduces an opinion to an equivalent classical probability, enabling compatibility with standard probabilistic reasoning.

Binomial opinions can be derived from evidence: with $r_x$ and $s_x$ the amounts of positive and negative evidence for $x$, this work uses \emph{Baseline-Prior Quantification},
\begin{equation}
(b_x,\, d_x,\, u_x) = \frac{(r_x,\; s_x,\; W)}{W + r_x + s_x},
\label{eq:q1}
\end{equation}
where the weight \(W>0\) guarantees residual uncertainty; evidence-weighted and constant-uncertainty quantification~\cite{10.1007/978-3-032-19096-3_17} are alternatives.

In this work, trust is a binomial opinion $(t, d, u)$, with $t$ trust (belief), $d$ distrust (disbelief), and $u$ uncertainty; the proposition it refers to is fixed in \cref{sec:method}.

\vspace{-0.5em}
\subsection{Subjective Logic Operators}\label{sec:backslop}

SL provides three families of reasoning operators for manipulating opinions~\cite{8009820,10706345,josang2016subjective}, all used in this work.

\begin{definition}[SL Operators]\label{def:fusion}
Let \(A\) be an agent reasoning about a variable \(X\).

\emph{Fusion} combines opinions held by two sources \(P\) and \(Q\) about the same proposition, \(\omega^A_{x} = \omega^P_{x} \oplus \omega^Q_{x}\), with variants (consensus, averaging, weighted, cumulative) chosen by whether the sources are correlated or independent.

\emph{Trust discounting} propagates an opinion along a referral chain: if \(A\) holds referral trust \(\omega^A_B\) in \(B\), who holds \(\omega^B_x\), then \(\omega^{[A;B]}_{x} = \omega^A_B \otimes \omega^B_{x}\), so that evidence from a partially reliable source is weakened and its uncertainty increased.

\emph{Inferential operators} generalize Bayesian reasoning through a conditional opinion \(\omega^A_{Y \mid X}\): \emph{deduction} derives \(\omega^A_Y\) from \(\omega^A_X\), and \emph{abduction} derives \(\omega^A_X\) from \(\omega^A_Y\).
\end{definition}
\noindent Operator details and symbols appear in \apprefs{app:operators}{app:operator}.

Building on these operators, \emph{Subjective Trust Networks}~\cite{4622580,10.1007/978-3-032-08887-1_4} model trust relationships as subjective opinions propagated through trust discounting and fusion. 

\vspace{-0.5em}
\subsection{Trustworthiness in the dataset}\label{sec:backtwdt}
% \todo[inline]{Paper Trustworthiness in Dataset}

Training-data quality largely determines the performance, robustness, and fairness of a model: sampling biases, mislabeled instances, or lack of diversity degrade learned representations~\cite{arp2021dosdontsmachinelearning}, and even small fractions of mislabeled samples can change model rankings~\cite{northcutt2021pervasivelabelerrorstest}. Subjective Logic has been applied to model dataset trustworthiness~\cite{10.1007/978-3-032-19096-3_17,trust-aidatatrust} at three granularities: the \emph{dataset level} (e.g., class imbalance or sampling bias), the \emph{instance level} (mislabeled or corrupted samples), and the \emph{input feature level}, as when an adversarial patch renders only part of an image untrustworthy~\cite{Vargas2020} or features originate from sources of differing reliability.

\vspace{-0.5em}
\section{Related Work}\label{sec:rw}

Quantifying trust and uncertainty in neural networks is central in safety-critical domains, where incorrect yet confident predictions can have severe consequences; under data poisoning, a model may score well on standard metrics while having learned erroneous patterns.
\vspace{-0.5em}
\subsection{Uncertainty Quantification in Neural Networks}

Uncertainty quantification estimates predictive reliability by modeling epistemic and aleatoric uncertainty~\cite{kendall2017uncertainties}, through Bayesian neural networks~\cite{neal1996bayesian,blundell2015weightuncertaintyneuralnetworks,pmlr-v80-depeweg18a}, Monte Carlo dropout~\cite{gal2016dropoutbayesianapproximationrepresenting}, and ensembles~\cite{lakshminarayanan2017simplescalablepredictiveuncertainty}. Yet calibration remains difficult in complex models~\cite{Gawlikowski2023}, confidence degrades under dataset shift~\cite{begoli2019need,2506.07461,1906.02530}, and post-hoc calibration such as temperature scaling~\cite{guo2017calibrationmodernneuralnetworks} acts only at the output layer and assumes clean data, so it can still mislead when inputs or training data are corrupted.
\vspace{-0.5em}
\subsection{Subjective Logic Approaches to Trust}

Evidential deep learning (EDL)~\cite{sensoy2018evidentialdeeplearningquantify} applies SL principles by representing class predictions as Dirichlet-parameterized subjective opinions, jointly predicting outcomes and quantifying confidence so as to flag high-uncertainty regions such as out-of-distribution inputs; although effective, it assumes clean data and provides no input-level trust assessment. A calibration-based approach~\cite{11124121} instead clusters model outputs into subjective opinions to derive per-prediction trust scores without accessing internal parameters; while simple and model-agnostic, it likewise assumes a trustworthy dataset and neglects input evidence.

Two further lines relate directly to our evaluation. Input- and feature-space detectors flag distribution violations without touching training dynamics: Mahalanobis distances on hidden features~\cite{lee2018mahalanobis}, deep nearest-neighbor distances~\cite{sun2022knnood}, energy scores~\cite{liu2020energy}, and conformal prediction~\cite{angelopoulos2023conformal} wrap point predictions with distribution-level evidence; \cref{exp:4} includes the first three as baselines and quantifies what trust propagation adds beyond them. Data-centric methods attribute unreliability to training points: influence functions~\cite{koh2017influence} and confident learning~\cite{northcutt2021confidentlearning} estimate which labels are erroneous and grade corruption severity, and we compare against such signals in the training-quality audit. PaTAS differs from both lines by composing input-, parameter-, and path-level trust into a single propagating calculus with an interpretable opinion output, rather than producing a task-specific score.

The DeepTrust framework by Cheng et~al.~\cite{10.3389/frai.2020.00054} adopts a white-box perspective, integrating dataset evidence during training to assess global model trustworthiness. Although holistic, its mapping of neural-network addition to SL fusion and multiplication to SL opinion multiplication combines operators of different algebraic types (fusion is defined over agents' opinions, multiplication over variables), and its trust computation is specified for a single layer, so it is not clear how the scheme extends to deeper architectures. PaTAS addresses both through a propagation mechanism defined layer-wise, hence at arbitrary depth, with consistency properties established analytically, convergence proven for the idealized fixed-revision iteration, and training-time stability assessed empirically (\cref{sec:ptastheory}).
\vspace{-0.5em}
\subsection{Trust and Uncertainty Propagation}

Work on uncertainty propagation improves accuracy and efficiency: sampling-free variance propagation for dropout networks~\cite{MAE2021394}, Gaussian-mixture propagation of input uncertainty~\cite{pmlr-v202-monchot23a}, and extensions to multi-layer perceptrons~\cite{astudillo2011propagation}. Beyond neural models, the Appleseed model propagates trust in social networks via spreading activation~\cite{ziegler2004spreading}, showing the value of treating trust as a structural, context-dependent quantity.

These approaches typically treat reliability as a property of the output layer or of the model as a whole, and none propagates trust jointly from input, training data, and parameters. PaTAS differs in three specific respects rather than in its use of Subjective Logic alone: its trust computation \emph{structurally mirrors} the network, with one Trust Node per neuron, so trust propagates along the same topology as the prediction; it derives \emph{parameter-level} trust during training from gradient evidence conditioned on label trust, which existing evidential and calibration methods do not model; and its Inference-Path Trust Assessment yields \emph{per-instance, path-specific} trust at deployment. The individual ingredients are known; the contribution is their composition into a single mechanism assigning and propagating trust at input, parameter, and path granularity simultaneously, with proven consistency properties.

\vspace{-0.5em}
\section{Proposed Method}\label{sec:method}
\paragraph*{What trust denotes} PaTAS attaches a binomial opinion to every quantity of the computation (an input feature, a parameter, a neuron output, and ultimately the prediction), always about the same proposition: \emph{the quantity is supported by reliable evidence}. Data noise, adversarial manipulation, and the training history of a parameter are therefore not distinct notions of trust but distinct \emph{sources of evidence} about this proposition: input opinions carry evidence on how an input was acquired or how well it conforms to the training data, parameter opinions carry evidence on the data and gradient corrections that shaped the parameter (\cref{sec:ptasfunc}), and the inference path selects which parameter opinions bear on a given prediction. For the output opinion, $t$ is evidence that the prediction rests on reliable inputs and parameters, $d$ evidence that it rests on distrusted inputs, and $u$ the absence of evidence either way.\footnote{Parameter opinions act as referral trust, so unreliable parameters (e.g., shaped by distrusted labels) lower $t$ and raise $u$ but never create $d$ (\cref{thm:infvac}).} It is neither a probability of correctness nor an out-of-distribution score, although it can correlate with both; \cref{exp:4} measures where it does.

\paragraph*{What PaTAS provides and what it assumes} PaTAS provides the propagation calculus, the parameter opinions derived during training by the Parameter-Trust Update, and the per-inference output opinion. It consumes opinions on input features and training labels but does not prescribe how they are obtained; the available routes, and what PaTAS still provides without a reliable external assessment, are discussed in \cref{sec:acquisition}.

\vspace{-0.5em}
\subsection{Foundations of Trust-Aware Neural Inference for PaTAS}
% \todo[inline]{General Representation of NN}
Given a NN represented by \(\Theta = \left(\mathbf{W}, \mathbf{b}, \phi\right)\), where \(\mathbf{W} = \{\mathbf{W}^{(l)}\}\) collects the weight matrices \(\mathbf{W}^{(l)} \in \mathbb{R}^{n_l \times n_{l-1}}\) of each layer \(l\) (with \(n_l\) the number of neurons in layer \(l\)), \(\mathbf{b} = \{\mathbf{b}^{(l)}\}\) the bias vectors \(\mathbf{b}^{(l)} \in \mathbb{R}^{n_l}\), and \(\phi = \{\phi^l(\cdot)\}\) the activation functions, which may vary across the neurons of a layer,
with standard feedforward output \(\mathbf{y\prime} = f_\Theta(x)\). Given a training dataset \( \mathcal{D} = \{ (\mathbf{x}_i, \mathbf{y}_i) \}_{i=1}^{N} \) and a trust assessment function \( T(\cdot) \) over features and labels, our objective is to compute a trust opinion on the output \( \mathbf{y\prime} = f_\Theta(\mathbf{x}) \). We formalize this through a \emph{Parallel Function}.

\begin{definition}[Parallel Function]
The \emph{Parallel Function} of a Neural Network with parameters \(\Theta\), denoted \( \pf_\Theta \), is a function that mirrors the structure of the network’s feedforward computation \( f_\Theta \) to propagate trust assessments from input to output. Given a trust evaluation \( T(\mathbf{x}) \) over an input \(\mathbf{x}\), \( \pf_\Theta(T(\mathbf{x})) \) returns a trust opinion on the network’s output \(\mathbf{y\prime} = f_\Theta(\mathbf{x})\), reflecting trust in the input as well as the architecture and how the parameters of \( f_\Theta \) were learned.
\end{definition}

The parameters follow from the usual empirical-risk minimization \(\mathbf{W},\mathbf{b} = \arg\min \sum_i \mathcal{L}(\mathbf{y\prime}_i,\mathbf{y}_i)\), which optimizes accuracy but says nothing about how input trust influences output trust. We therefore use \emph{Subjective Logic} as the propagation calculus and build \(\pf_\Theta\) for general networks, with the perceptron as a tractable illustration.
 
\vspace{-0.5em}
\subsection{Perceptron Case: SL Formulation}\label{sec:perceptron}

Consider an observer \(A\) forming a trust opinion \(\omega_{y\prime}^A\) on the output \(y\prime\) of a perceptron. Not observing the internal computation directly, \(A\) relies on the opinion \(\omega_{y\prime}^{N_O}\) held by the output neuron \(N_O\) and on a referral trust \(\omega_{N_O}^A\) in \(N_O\) itself; trust in the output is thus determined by trust in the input and trust in the perceptron, the latter following from its design and the trustworthiness of the training data. Applied to a two-feature perceptron \(y\prime = \theta_1 s + \theta_2 n_r\) with input trust opinion \(T_x\)\footnote{We write \(T_x\) rather than \(\omega_x\) since, as a framework input, only the opinion on the variable is required and no specific agent need be represented.}, this reasoning yields the following, derived in full in \appref{app:perceptron}.

\begin{lemma}[Perceptron Trust Propagation]\label{lem:perceptron}
Let \(T_s\) and \(T_{n_r}\) be trust opinions on the input features of the perceptron \(y\prime = \theta_1 s + \theta_2 n_r\), and let \(T_{\theta_1}, T_{\theta_2}\) be trust opinions on its parameters. Then
\begin{equation}
\label{eq:sol}
    \pf_\Theta\!\left(\begin{pmatrix} T_s & T_{n_r}\end{pmatrix}\right) =  (T_{\theta_1}\otimes T_s)\oplus (T_{\theta_2}\otimes T_{n_r}),
\end{equation}
where \(\otimes\) is a trust discounting operator and \(\oplus\) a fusion operator whose variant is determined by the semantics of the features (\cref{def:fusion}). For fully trusted parameters \(T_\theta = (1,0,0)\) this reduces to \(T_s \oplus T_{n_r}\).
\end{lemma}

\Cref{eq:sol} shows that parameter trust discounts the contribution of each feature before the contributions are fused, so that the output assessment reflects trust in both the data and the learned model. How the parameter-trust opinions \(T_{\theta_i}\) are themselves computed is the subject of \cref{sec:patas}.

\subsection{Trust Nodes and Trust Functions}

To extend \(\pf_\Theta\) from a single perceptron to larger networks, we introduce two fundamental concepts, the \emph{Trust Node} and the \emph{Trust Function}, which provide the basic mechanisms for propagating trust through the network structure; both are illustrated alongside the corresponding neuron operations in \xcrefin{fig:trustnodefull}{app:trustnode}.

\begin{definition}[Trust Node]
A \emph{Trust Node} is an abstract computational unit associated with a neuron: it receives trust opinions on the neuron’s inputs and produces a trust opinion on its output, mirroring the neuron’s structure while computing over trust opinions with operators such as discounting and fusion.
\end{definition}

\begin{definition}[Trust Function]\label{def:trustfunc}
A \emph{Trust Function} models the transformation of trust through a Trust Node. It defines how trust opinions on the inputs of a neuron are combined to produce a trust opinion on the output. 
For a neuron that computes
\(
z^{(l)} = \sum \theta^{(l)}_i \cdot x^{(l)}_i, \quad x^{(l+1)} = \phi^{(l)}(z^{(l)}),
\)
the corresponding trust computation is given by:
\begin{equation}
T_{z^{(l)}} = \bigoplus_{i=1}^{d} T_{\theta_{i}^{(l)}} \otimes T_{x_{i}^{(l)}}, \quad
T_{x^{(l+1)}} = T_{\phi^{(l)}}(T_{z^{(l)}}),
\label{eq:layerprop}
\end{equation}
where \( \otimes \) is a trust discounting operator, \(\bigoplus\) a trust fusion operator, and \( T_{\phi}^{(l)} \) the trust-equivalent of the activation function, set to the identity in this work since we use ReLU activations. Since \(\otimes\) is non-commutative, the parameter opinion stands first: it plays the role of referral trust in the virtual source holding the input opinion, exactly as in \cref{eq:sol}. The weight and activation magnitudes do not enter \cref{eq:layerprop}: the Trust Function tracks the reliability of what enters a computation, not its functional influence on the output (see \cref{sec:limitations}).
\end{definition}

As in \cref{eq:sol}, the discount operator \( \otimes \) models how trust in an input is modulated by trust in the associated parameter, while \( \bigoplus \) combines these contributions across inputs; fusion operators should be associative or generalizable~\cite{vanderheijden2018multisourcefusionoperationssubjective} to support multiple inputs. Reading the fused terms as opinions of distinct virtual sources about the same node output keeps this mapping algebraically consistent (\appref{app:operators}), a consistency that variable-level fusion, as in DeepTrust, does not guarantee. Averaging fusion is used for feedforward summation because it is fan-in invariant, keeping trust magnitudes comparable across layer widths, whereas cumulative fusion accumulates evidence with fan-in and drives wide layers toward dogmatic opinions; cumulative (evidence-additive) fusion is reserved for pooling genuinely accumulating evidence, as in the input-conformity construction of \cref{exp:4}.

Two challenges remain, quantifying the trustworthiness of parameters learned from datasets of variable quality and generalizing trust propagation to deeper architectures; they motivate the Parallel Trust Assessment System (PaTAS), presented next.
\section{PaTAS For Neural Networks}\label{sec:patas}

When parameters such as \( \theta_1 \) and \( \theta_2 \) are learned by backpropagation, their trustworthiness depends on the training data: mislabeled or biased samples compromise parameter trust. The SL opinion representation of \cref{sec:backsl} further distinguishes the causes of unreliability, with distrust arising from systematic issues such as mislabeled or poisoned data and uncertainty reflecting variability or noise in the assessment. A central challenge is how to exploit this distinction in practice.

\vspace{-0.5em}
\subsection{PaTAS General Description}

To address these challenges, we introduce the \emph{Parallel Trust Assessment System (PaTAS)}, which propagates trust assessments through a neural network by operating in parallel with the standard architecture and maintaining a structure mirroring its topology. It computes trust in the output from two sources: the trust in the input features at inference time, and the trust in the parameters, derived from the trustworthiness of the training dataset (covering both the features and the labels of the training samples). To extend the perceptron principles of the previous section, PaTAS associates a Trust Node with each neuron and connects the nodes according to the network's topology, forming a Trust Nodes Network (TNN).
\begin{definition}[Trust Nodes Network]
The \emph{Trust Nodes Network} is a structured composition of Trust Nodes, one per neuron of the underlying network, mirroring its architecture; it propagates trust across layers by applying trust-specific reasoning operations aligned with the neural computation flow.
\end{definition}

PaTAS continuously evaluates trustworthiness during inference: an input \(x\) may be trustworthy (high \(T_x\)) while the output \(y\prime\) is unreliable because of biased or poorly calibrated parameters \(\theta\) (low \(T_{y\prime}\)). Embedding trust operations inside the network would alter its computations, risk harming accuracy, and raise cost; PaTAS therefore performs trust assessment in parallel (\appref{app:flow}), so that trust reasoning never interferes with the model's predictions.

\vspace{-0.5em}
\subsection{PaTAS Design}
PaTAS is composed of four main modules (high-level overview and functional flow in \appref{app:flow}). \emph{Trust Feedforward} propagates trust through the Trust Node Network in alignment with the model's inference flow, combining input and parameter trust at each layer through the Trust Function's discounting and fusion; it produces a trust opinion per output neuron and stores the intermediate trust scores later used by the Parameter-Trust Update. \emph{Output-Trust Aggregation} combines the per-neuron output opinions into a single trust score via an SL fusion operator; alternatively, the decision trust can be read directly off the predicted class's output Trust Node. \emph{GenIPTA} dynamically constructs, for a single inference, an Inference-Path Trust Assessment (IPTA) reflecting the trustworthiness of the exact computational path taken: the activation trace instantiates a temporary subnetwork containing only the Trust Nodes of the activated neurons; since all neurons are typically activated to some degree while only a subset is strongly relevant, GenIPTA can also operate on truncated activation sets or weight the assessment by activation values. The fourth module, the \emph{Parameter-Trust Update}, parallels the role of backpropagation and receives a dedicated subsection.

\paragraph*{Convolutional networks}\label{sec:convipta} Under weight sharing an activation path cannot select weights, only \emph{applications} of a filter, so binary neuron pruning generalizes to a participation degree: each channel's contribution to the next layer's fusion is weighted by the fraction of spatial positions where its post-ReLU output was positive, and the bias and input rows keep weight one. For binary participation this weighted propagation coincides exactly with the row-pruned propagation of the fully connected case, so conv IPTA is a strict generalization rather than a new mechanism; average pooling is fusion over the window and a residual connection fuses the block input with the block output. \Cref{exp:4} evaluates this per-sample conv IPTA on three convolutional networks.

\vspace{-0.5em}
\subsection{Parameter-Trust Update}\label{sec:ptasfunc}
The Parameter-Trust Update adapts the trust of model parameters during training so that it reflects both the model's observed behavior and the reliability of the training data. To ground the design, recall the standard gradient-descent update for a parameter $\theta$, \(\theta \leftarrow \theta - l_r \, \partial \mathcal{L} / \partial \theta\), with $l_r$ the learning rate and $\mathcal{L}$ the loss. The gradients are computed recursively using the chain rule:
\begin{equation}
\label{eq:gd2}
\frac{\partial \mathcal{L}}{\partial \theta^{(l)}_{i,j}}
= \delta^{(l)}_i x^{(l-1)}_j,
\end{equation}
where $\theta^{(l)}_{i,j}$ denotes the weight connecting neuron $j$ in layer $(l-1)$ to neuron $i$ in layer $l$, $\delta^{(l)}_i$ is the error term of neuron $i$, and $x^{(l-1)}_j$ is the activation from the previous layer.

\Cref{eq:gd2} shows that gradients factor into an error term and an input activation, so parameter updates depend jointly on gradient signals, input activations, and learning dynamics; PaTAS mirrors this by using gradient information, input feature trust, and label trust as evidence in the Parameter-Trust Update. Notation is summarized in \appref{app:symbols}; the complete procedure is \cref{alg:trustupdate}.

\begin{algorithm}
\footnotesize
\caption{Parameter-Trust Update: revises each parameter's trust from gradient evidence, label trust, and neuron input trust under mini-batch training (helper functions in \appref{app:algorithm}).}
\begin{algorithmic}[1]
    \Function{ParameterTrustUpdate}{$g, T_y, \epsilon$}
        \State $T_{y_\text{batch}}  \longleftarrow  \bigwedge_{y \in y_\text{batch}} T_y$ \Comment{aggregate label trust}
        \For{each layer $l$}
            \For{each neuron $n_i^{(l)}$}
                \State $g^{(l)}_i  \longleftarrow  \{\, g^{(l)}_{i,j} \;|\; j \in \mathcal{N}(i)\,\}$
                \State $T_{n_i \mid y_\text{batch}}  \longleftarrow $ \Call{NodeTrust}{$g^{(l)}_i, T_{y_\text{batch}}, \epsilon$}
                \State $T_{n_i \parallel Y_\text{batch}}  \longleftarrow $ \Call{DeduceTrust}{$T_{n_i \mid y_\text{batch}}, T_{n_i \mid \overline{y_\text{batch}}}, T_{y_\text{batch}}$}
                \For{each incoming edge $j$ to $n_i^{(l)}$}
                    \State $T_{\theta_{i,j}^{(l)}}  \longleftarrow $ \Call{Revise}{$T_{\theta_{i,j}^{(l)}}, T_{n_i \parallel Y_\text{batch}}$}
                    \State $T_{\theta_{i,j}^{(l)}} \longleftarrow $ \Call{Update}{$T_{\theta_{i,j}^{(l)}}, T_{x^{(l-1)}_j}, T_{y_\text{batch}}$}
                \EndFor
            \EndFor
        \EndFor
    \EndFunction
\end{algorithmic}
\label{alg:trustupdate}
\end{algorithm}
\Cref{alg:trustupdate} runs once per mini-batch over all layers and neurons. It first aggregates label trust, \(T_{y_\text{batch}} = \bigwedge_{y \in y_\text{batch}} T_y\), which conditions the trust in every parameter, then derives for each neuron \(n_i^{(l)}\) the opinion \( T_{n_{i}^{(l)} \mid y_\text{batch}} \) from gradient evidence, counting each incoming weight gradient with \( |g^{(l)}_{i,j}| < \epsilon\)\footnote{The threshold $\epsilon$ is a scale parameter used only in \textsc{NodeTrust} to separate weak from strong gradients; it must be set relative to the prevailing gradient scale of the task (see the sweep in \appref{app:ablations} and the selection rule in \cref{sec:discussion}) and never affects model predictions or training dynamics.} as positive evidence \(r\) and the remainder as negative evidence \(s\), and mapping \((r,s)\) to a binomial opinion via \cref{eq:q1}. Since no evidence is available for the complementary case, \( T_{n_{i}^{(l)} \mid \overline{y_\text{batch}}} \) is vacuous, and the deduction operator \( \circledcirc \) combines the two conditionals with \( T_{y_\text{batch}} \) to give \( T_{n_{i}^{(l)} \parallel Y_\text{batch}} \).

For each incoming edge \(j\), the parameter trust is then revised with the deduced neuron trust, \(T_{\theta_{i,j}^{(l)}} \leftarrow T_{\theta_{i,j}^{(l)}} \circleddash T_{n_{i}^{(l)} \parallel Y_\text{batch}}\), and adjusted for the auxiliary factors that also drive the gradient update in \cref{eq:gd2}, namely the input feature and the batch labels (learning-rate trust is a possible third factor; it is not modeled in this work):
\begin{equation}
    T_{\theta_{i,j}^{(l)}} \longleftarrow T_{\theta_{i,j}^{(l)}} \odot (T_{x_j^{(l-1)}} \oslash T_{y_\text{batch}}),
\end{equation}
where $\odot$ is binomial multiplication and $\oslash$ is the conservative combination \((b,d,u)\) with \(b = \min(b_1,b_2)\), \(d = \max(d_1,d_2)\), \(u = 1-(b+d)\). This reflects that both unreliable features and mislabeled data can strongly bias parameter updates: trust cannot exceed the weakest evidence and distrust must reflect the strongest warning. Bias terms are handled as one additional parameter row whose input carries a vacuous opinion and whose trust is revised like any weight.

The mapping from gradient magnitude to evidence deserves explicit justification, since a large gradient does not, in isolation, imply an untrustworthy parameter. Two points make the interpretation well founded. First, evidence is \emph{relative}: gradients are thresholded against the prevailing gradient scale through $\epsilon$, so the update reacts to a parameter that keeps receiving atypically large corrections \emph{after} the model has otherwise settled, not to the large gradients normal early in training. Second, evidence is \emph{conditional on label trust}: under trusted labels, a parameter the network must repeatedly and strongly correct is one whose learned value the trusted data does not support, precisely the signal PaTAS accumulates as reduced trust, while under distrusted or uncertain labels the deduction through $T_{y_\text{batch}}$ discounts that same signal proportionally (asymptotic cases in \appref{app:asymptotic}). Gradient magnitude is nonetheless a proxy and can be confounded by learning-rate schedules, mini-batch noise, or saddle-point regions; the sensitivity of the trust output to $\epsilon$ is measured in \appref{app:ablations} and discussed in \cref{sec:discussion}, the level at which this signal has been validated is stated in \cref{sec:limitations}, and a more direct evidence model for parameter reliability is left to future work.

PaTAS mirrors the network's computational flow over four-component opinions rather than scalars, giving an analytical \(3\times\) overhead for the propagation arithmetic; measured costs and the memory footprint are reported in \cref{sec:discussion}. Because training requires no value from PaTAS, the neural network can proceed while trust assessment completes.

\vspace{-0.5em}
\subsection{Theoretical Properties of PaTAS}\label{sec:ptastheory}
We first establish fundamental properties describing the stability and consistency of PaTAS.
\subsubsection{Convergence of PaTAS}
% \todo[inline]{Discuss convergence of PaTAS in case of convergence of input}
\label{sec:ptas_convergence}

Theorem~\ref{thm:convergence} establishes that the recursive update of \cref{alg:trustupdate} converges under an explicitly stated idealization of stable training; the training-time behavior under mini-batch noise is assessed empirically.

% Before delving into the specifics of PaTAS convergence, we first present a key theorem that will serve as a foundational tool in establishing the convergence properties of PaTAS.

\begin{theorem}[Convergence of PaTAS Creation]\label{thm:convergence}
\footnote{All proofs of the theorems in this section, as well as additional theoretical results, are provided in \appref{app:proof}.}

Assume (i) the parameter opinions are initialized non-dogmatically (PaTAS initializes them as vacuous), and (ii) the per-batch revision opinion applied to each parameter stabilizes to a fixed opinion $q$: an idealization of a network trained to convergence with stationary gradient statistics and stable $T_x$, $T_y$, and $\epsilon$.
Let $T_\theta^{(n)}$ denote the trust opinion at iteration $n$ for parameter $\theta$. If the revision is performed as specified in \cref{alg:trustupdate}, then $T_\theta^{(n)}$ converges; for averaging-fusion revision the limit is $q$, at the geometric rate of Theorem~\xref{thm:arithm}.
\end{theorem}
Under actual mini-batch training the revision opinion fluctuates from batch to batch, so the theorem guarantees the consistency of the revision dynamics rather than convergence of the stochastic process itself; empirically, after accuracy convergence the trust trajectories oscillate within a narrow band around a stable level (\appref{app:expres}).
\subsubsection{Symmetry and Invariance Properties of PaTAS}\label{sec:ptassyminv}

\begin{theorem}[Consistency Properties of PaTAS]\label{thm:infvac}
Let \(\omega^\emptyset=(0,0,1,a)\) be a vacuous binomial opinion and let \(\bar{x}=(d,b,u,a)\) denote the symmetric counterpart of \(x=(b,d,u,a)\). Then, for any parameter trust configuration \(T_\theta\):
\begin{enumerate}
    \item \emph{Vacuity is preserved.} Discounting a vacuous opinion is vacuous, \(\omega^A_B \otimes \omega^\emptyset = \omega^\emptyset\), and hence \(\mathrm{IPTA}(\omega^\emptyset) = \omega^\emptyset\).
    \item \emph{Inference is symmetric.} The outputs \(y = T_\theta(x)\) and \(\bar y = T_\theta(\bar x)\) satisfy \(u_y = u_{\bar y}\), \(b_y = d_{\bar y}\), and \(d_y = b_{\bar y}\). In particular a fully trusted input \((1,0,0)\) yields \(d_y = 0\), and a fully distrusted input \((0,1,0)\) yields \(b_{\bar y} = 0\).
\end{enumerate}
\end{theorem}

These properties are consistency guarantees rather than the paper's headline result: they establish that the trust computation is well behaved and justify reasoning about only the trusted-input case in the evaluation; stronger security-specific guarantees, such as provable bounds on trust degradation under a bounded poisoning adversary, are not claimed and remain future work.

\vspace{-0.5em}
\section{Evaluation and Results}\label{sec:evaluation}

Our evaluation shows that PaTAS produces trust estimates that converge during training, respect the properties of \cref{sec:ptastheory}, and remain interpretable under noisy features, corrupted labels, and adversarial perturbations. We conduct four experiments of increasing complexity: (1)~\emph{Breast Cancer Classification} on a small tabular medical dataset; (2)~\emph{MNIST Digit Classification} across architectures under controlled uncertainty; (3)~\emph{Poisoned MNIST} under adversarial corruption; and (4)~\emph{Comparative Trust Evaluation} against established uncertainty baselines on out-of-distribution and corruption detection.

\vspace{-0.5em}
\subsection{Experimental Setup }
\subsubsection{Experiment 1 - Breast Cancer Classification}\label{exp:1}
We use the \href{https://archive.ics.uci.edu/dataset/17/breast+cancer+wisconsin+diagnostic}{Breast Cancer Wisconsin (Diagnostic) Dataset}~\cite{breastcancer-wisconsin}, 569 samples with 30 numeric features (e.g., radius, area, symmetry) derived from breast mass fine needle aspirates, classified into benign or malignant. The network has 30 input, 16 hidden (ReLU), and 2 output (Softmax) neurons, trained for 15 epochs with batch size 64 and learning rate 0.2, reaching 97\% test accuracy on unmodified data.

We degrade the training data in controlled ways and assign corresponding oracle trust assessments, which isolate the propagation mechanism from the problem of estimating opinions, combining three extreme profiles for features and labels alike, fully trusted \((1,0,0)\), fully distrusted \((0,1,0)\), and fully uncertain \((0,0,1)\), into nine cases; these dogmatic and vacuous opinions are the canonical SL boundary cases (\emph{fully uncertain} and \emph{vacuous} denote the same opinion \((0,0,1)\) and are used interchangeably). Degradation is simulated with controlled perturbation functions (\appref{app:perturb}): uncertain features receive additive uniform noise clipped to the valid range, distrusted features are resampled uniformly, and labels are degraded analogously through random noise (uncertainty) or complete replacement (distrust). Two intermediate scenarios complement the boundary cases (last two rows of \cref{tab:cancer_results}): the fully uncertain features-and-labels setting re-assessed as the mixed opinion \((0.25,0.25,0.5)\), and a mild degradation where features are perturbed with probability 0.15 and assessed as \((0.25,0,0.75)\).

\subsubsection{Experiment 2 - MNIST}\label{exp:2}

We evaluate PaTAS on MNIST~\cite{lecun1998mnist} (60{,}000 training and 10{,}000 test \(28\times28\) images, ten classes) with four architectures of increasing width, 784-16-10 through 784-128-10; not state of the art, but sufficient to expose how trust propagates across model sizes. Each uses ReLU and Softmax, trained for 20 epochs with batch size 128 and learning rate 0.05, reaching 99\% test accuracy on fully trusted data; the gradient-evidence threshold is $\epsilon = 0.05$ here and in Experiments 3 and 4. Feature and label trust assessments are fully uncertain, corresponding to no knowledge of the dataset (the vacuous route~(iii) of \cref{sec:acquisition}); for contrast we repeat the smallest architecture with a fully trusted assessment.

\subsubsection{Experiment 3 - Poisoned MNIST}\label{exp:3}
We poison one third of the MNIST training images by flipping the labels of digits 6 and 9 while adding a visible fixed-size patch at the top-left corner, following common backdoor practice~\cite{chen2017targetedbackdoorattacksdeep}; the remaining two thirds stay clean. Using the 784-128-10 architecture, patch pixels and the labels of patched 6s and 9s are assigned distrust and everything else trust, so PaTAS focuses on the corrupted regions while trusting the unaffected data. This oracle assignment is replaced by estimated opinions in \cref{exp:4}.

\subsubsection{Experiment 4 - Comparative Trust Evaluation}\label{exp:4}
We compare the per-sample PaTAS assessment against maximum softmax probability, MC~Dropout~\cite{gal2016dropoutbayesianapproximationrepresenting} ($T{=}30$, $p{=}0.2$), and Evidential Deep Learning~\cite{sensoy2018evidentialdeeplearningquantify} (ranking metrics score its negated Dirichlet total uncertainty $1-u$, $u = K/S$; calibration metrics its maximum expected probability $\max_k \alpha_k/S$), each trained with the modifications its method requires under the shared architecture, optimizer, and epoch budget (DeepTrust~\cite{10.3389/frai.2020.00054} has no public implementation; see \cref{sec:rw}). To attribute mechanisms rather than merely outperform output-only scores, we further include three input- and feature-space detectors, the class-conditional Mahalanobis distance on penultimate features~\cite{lee2018mahalanobis}, the deep nearest-neighbor distance~\cite{sun2022knnood}, and the energy score~\cite{liu2020energy}, together with the conformity score $P_{\mathrm{input}}$ alone, i.e., the route-(ii) input opinions with all propagation removed. Primary datasets are MNIST and Fashion-MNIST~\cite{xiao2017fashionmnist} (784-512-10, 20 epochs); GTSRB~\cite{GTSRB} (1024-128-43, train-standardized) serves as an applicability-boundary case. The battery is additionally run on convolutional networks with the per-sample conv IPTA of \cref{sec:convipta}: ResNet-style CNNs on MNIST and Fashion-MNIST ($99.3\%$ and $88.7\%$ test accuracy) and a seven-layer residual CNN on CIFAR-10 ($84.5\% \pm 0.8$); the networks use no normalization layers, so trust propagation covers every parameter (architectures and training details in \appref{app:conv}).

The PaTAS assessment instantiates route~(ii), estimation from data, of \cref{sec:acquisition}: each input feature receives a trust opinion from its conformity to the training-data statistics, mapped through the same Baseline-Prior Quantification as parameter trust (\cref{eq:q1}) with per-feature reliability weights (full construction in \appref{app:conformity}). The trust factor is the serial discount $P = P_{\mathrm{model}} \cdot P_{\mathrm{input}}$ of the IPTA path trust and the pooled input conformity, and the reported score anchors the discounted confidence at the class prior, $P \cdot \hat{p} + (1-P)/K$. Every method is evaluated identically on clean test data, under Bernoulli-uniform feature corruption, on a mixed batch pooling clean and corrupted inputs 1:1, and as an OOD detector (Fashion-MNIST for MNIST and vice versa, identical preprocessing); post-hoc isotonic calibration is fit per method on a held-out clean split. Results are mean~$\pm$~std over three evaluation seeds.

\paragraph*{Implementation Details for All Experiments}
Feedforward multiplications are implemented as SL trust discounting and additions as generalized SL averaging fusion (\cref{def:trustfunc}; operational flow in \xcref{fig:ptasoperation}); trust revision uses the same fusion. Baseline-Prior Quantification (\cref{eq:q1}) uses the non-informative prior weight \(W = 2\) throughout, for parameter trust and input conformity alike. The full Python/NumPy implementation and all experiment scripts are available as \suppp.

\vspace{-0.5em}
\subsection{Results and Analysis}
Evaluation is performed during training: after each feedforward/backpropagation iteration we assess the output trustworthiness under three boundary input profiles, fully trusted, fully uncertain, and fully distrusted, tracking trust, uncertainty, and distrust mass over training (Experiments 1--3)\footnote{The base rate is set to $0.5$, the maximum-entropy choice, and held constant; the score anchoring in \cref{exp:4} uses the class prior $1/K$ where a prior probability of correctness is the natural reference. A systematic base-rate sensitivity study is left to future work.}. Results are collected in \appref{app:expres} and summarized in \cref{tab:cancer_results,tab:mnist,tab:mnistpoisipta}; they confirm the properties proven in \cref{sec:ptastheory}: \emph{stability} (after accuracy converges, the trust assessment settles into a narrow band, the empirical counterpart of \cref{thm:convergence}), \emph{vacuity preservation} (a fully uncertain input always produces a fully uncertain output), and \emph{symmetry} (symmetric input trust assessments yield symmetric inference results).

\begin{table*}[ht]
\begin{minipage}{0.62\textwidth}
\centering\footnotesize
\begin{tabular}{cccccc}
\hline
X Trust Assessment & Y Trust Assessment& \(\epsilon = 0.01\) & \(\epsilon = 0.1\) & Train (\%) & Test (\%) \\
\hline
fully distrusted & any & 0 & 0 & 47--58 & 41--55 \\
any & fully distrusted & 0 & 0 & 96--98 & 3--4 \\
fully uncertain & fully uncertain & 0.17 & 0.28 & 66 & 85 \\ % noise-noise
% \hline
fully uncertain & fully trusted & 0.15 & 0.32 & 96 & 96 \\ % noise-clean
% \hline
 fully trusted & fully uncertain & 0.12 & 0.29 & 69 & 94 \\ % clean-noise
% \hline
fully trusted & fully trusted & 0.62 & 0.88 & 98 & 97 \\ % clean-clean
(0.25, 0.25, 0.5) & (0.25, 0.25, 0.5)& 0.06 & 0.11 & 66 & 85 \\
% \hline
(0.25, 0, 0.75) & (0.25, 0, 0.75)& 0.20 & 0.35 & 75 & 94\\
% \hline

\hline
\end{tabular}
\caption{Breast Cancer: final trust mass and train/test accuracy (\%). Rows with a fully distrusted assessment collapse to zero trust and are grouped.}
\label{tab:cancer_results}
\end{minipage}
\hfill
\begin{minipage}{0.36\textwidth}
\centering\footnotesize
\begin{tabular}{cccc}
\hline
Hidden Neurons & Trust \(t\) & Train (\%) & Test (\%) \\
\hline
16 & 0.306 & 65 & 72\\
% \hline
32 & 0.325 & 66 & 89 \\
% \hline
64 & 0.333 & 67 & 90\\
% \hline
128 & 0.337 & 68 & 93\\
% \hline
16 (trusted) & 0.778 & 96 & 95\\
\hline
\end{tabular}
\caption{MNIST: final trust mass per architecture under fully uncertain feature and label assessments; the last row repeats the 16-neuron network under a fully trusted assessment.}
\label{tab:mnist}
\end{minipage}

\end{table*}

We therefore report only the fully trusted input case: uncertain inputs yield uncertain outputs, symmetry makes the distrusted case a mirror of the trusted one (\cref{thm:infvac}), and under trusted inputs the distrust mass stays near zero, so the trust mass alone determines the opinion; the tables report its converged value.

\subsubsection*{Experiment 1 -- Breast Cancer Classification}
Corrupted labels mislead rather than prevent learning: training accuracy stays high (96--98\%) while test accuracy collapses to near zero; corrupted features eliminate learning in every setting (41--55\% test, near chance), while label noise depresses training accuracy (69\%) yet still permits generalization (94\%).

We evaluated two values of $\epsilon$ (0.1 and 0.01), the \textsc{NodeTrust} threshold of \cref{alg:trustupdate}, as summarized in \cref{tab:cancer_results}. For $\epsilon = 0.1$ the trust mass tracks the assessments ($0.88$ for trusted $X$ and $Y$, $0.28$--$0.32$ under partial trust), and for either $\epsilon$ a distrusted $X$ or $Y$ drives it rapidly to 0. For $\epsilon = 0.01$ the system reads more gradients as large and all partial-trust configurations drop (0.12--0.17); notably, under degraded features the trusted-$Y$ case falls \emph{below} the uncertain-$Y$ case (0.15 vs.\ 0.17), since trusted labels turn persistent gradient corrections into negative evidence, the mechanism analyzed in \cref{sec:acquisition}.
Overall, trust mass correlates strongly with test accuracy among the non-distrusted configurations. Two tendencies stand out (\cref{tab:cancer_results,tab:mnist} report single training runs; both orderings are stable across the two $\epsilon$ settings, but we state them qualitatively). First, \emph{label trust matters more than feature trust}: fully uncertain $X$ with trusted $Y$ yields a higher trust mass (0.32) than trusted $X$ with uncertain $Y$ (0.29). Second, \emph{distrust is more damaging than uncertainty}: replacing the fully uncertain opinion $(0,0,1)$ on both $X$ and $Y$ with the mixed assessment $(0.25,0.25,0.5)$ reduces the trust mass from $0.28$ to $0.11$; the operator-level component of this asymmetry is discussed in \cref{sec:discussion}.

\subsubsection*{Experiment 2 -- MNIST Dataset}
PaTAS produces ten output trust opinions, one per class; as they are almost identical we record only the aggregated trust mass. As shown in \cref{tab:mnist}, under noisy features and labels accuracy improves with model size, and test accuracy consistently exceeds training accuracy, since training accuracy is measured against the noisy labels while the test set is clean. Trust also increases with model size, but with diminishing gains (0.306 $\to$ 0.337), contrasting sharply with the fully trusted evaluation on the smallest architecture (0.778): \emph{training a smaller architecture on trusted data benefits trust far more than training a larger architecture on uncertain data}. Because gradient scale itself shrinks with width at fixed $\epsilon$, we verified the trend at additional thresholds: the ordering persists at every $\epsilon \in \{0.02, 0.05, 0.1\}$ (\xcrefin{tab:abl-width}{app:ablations}), so it is not an artifact of the fixed threshold, although the trust magnitudes are $\epsilon$-dependent.

\subsubsection*{Experiment 3 – Poisoned MNIST Dataset}
PaTAS again produces ten output trust opinions, now informative because the reliability of the inference path varies across labels when some are patched and others are not. \xcrefin{tab:mnistpois}{app:pois} reports the trust mass for label 3 (clean) and label 6 (poisoned; see \cref{exp:3}) across patch sizes, with clean and patched test accuracy for both digits. At every patch size the trust mass of label 3 remains higher than that of label 6 by a wide margin (e.g., 0.90 vs.\ 0.58 for the $4\times4$ patch), confirming that PaTAS reliably distinguishes clean from poisoned inference paths; both values decrease as the patch grows, since larger patches affect more neurons along the inference path, yet the clean--poisoned separation remains interpretable at every size (0.70 vs.\ 0.49 even at $27 \times 27$).

To evaluate the IPTA, \cref{tab:mnistpoisipta} reports accuracy and trust opinions obtained by feedforwarding a fully trusted opinion through the IPTA derived from clean digit-3, clean digit-6, and patched digit-6 images. The backdoor is near-total (patched 6s are misclassified in 99\% of cases), and the assessment exposes it sharply: clean digit-6 samples retain high accuracy ($97.6\%$) yet carry a markedly lower trust mass ($0.503 \pm 0.035$, mean $\pm$ std over ten independent trainings) than clean digit-3 samples ($0.884 \pm 0.001$), because the parameters along the digit-6 inference path were learned from poisoned data: an accurate prediction whose supporting evidence PaTAS refuses to endorse, precisely the reliability gap that confidence scores do not reveal. Patched samples inherit the same reduced trust and elevated uncertainty, and explicitly distrusting the patch pixels additionally surfaces a non-zero distrust mass ($0.493, 0.011, 0.496$). PaTAS thus warns interpretably about poisoned outputs even when accuracy alone looks healthy; \cref{exp:4} shows the same flagging emerges without oracle patch knowledge.

\begin{table}[ht]
\centering\footnotesize
\setlength{\tabcolsep}{4pt}
\begin{tabular}{ccccc}
\hline
 & Accuracy (\%) & Trust  & Distrust & Uncertainty \\
\hline
Clean 3 & 98.0 & $0.884{\scriptstyle\pm.001}$ & 0.0 & 0.116\\

Clean 6  & 97.6 & $0.503{\scriptstyle\pm.035}$ & 0.0 & 0.497\\

6 with patch& 1.0 & $0.503{\scriptstyle\pm.035}$ & 0.0 & 0.497\\

6 with patch & -- & $0.493{\scriptstyle\pm.035}$ & 0.011 & 0.496\\
(patch distrusted) &&&&\\
\hline
\end{tabular}
\caption{IPTA on the poisoned model ($4\times 4$ patch; mean $\pm$ std over ten trainings, accuracy std $\leq 0.4$), feedforwarding a fully trusted opinion; in the last row patch pixels are explicitly distrusted. The poisoned class carries reduced trust even on clean, correctly classified inputs.}
\label{tab:mnistpoisipta}
\end{table}

\subsubsection*{Experiment 4 -- Comparative Trust Evaluation}
\Cref{tab:comparative} reports the two detection tasks in which trust and confidence should differ: recognizing out-of-distribution inputs, and recognizing corrupted inputs inside a mixed batch. The attribution is stated plainly: violation detection is carried by the conformity-based input opinions, not by trust propagation. The conformity score alone is the strongest detector on both tasks, the composed PaTAS score inherits most of this, and the ablation row confirms the direction: with the input opinions replaced by the constant fully-trusted opinion, PaTAS collapses to the output-only level. The dedicated detectors show why the construction still matters: the feature-space methods are strong on OOD (kNN $0.994$) but largely miss input-space corruption ($0.588$), since the penultimate features of a corruption-robust network wash the perturbation out. Across the four violation tasks, the route-(ii) conformity opinions are the only signal strong on all of them where the diagnostics of \xcref{tab:diagnostics} predict informativeness.

\begin{table}[ht]
\centering\footnotesize
\setlength{\tabcolsep}{3.5pt}
\resizebox{\columnwidth}{!}{%
\begin{tabular}{lccccc}
\toprule
 & \multicolumn{2}{c}{OOD M$\to$F} & \multicolumn{2}{c}{Corruption det.} & OOD F$\to$M \\
Method & AUROC & FPR@95 & MNIST & F-MNIST & AUROC \\
\midrule
PaTAS (composed) & $0.975{\scriptstyle\pm.001}$ & $0.145{\scriptstyle\pm.014}$ & $0.972{\scriptstyle\pm.002}$ & $0.835{\scriptstyle\pm.003}$ & $0.781{\scriptstyle\pm.005}$ \\
\;\;w/o input trust & $0.896$ & $0.415$ & $0.639$ & -- & -- \\
Conformity alone & $\mathbf{1.000}{\scriptstyle\pm.000}$ & $\mathbf{0.000}{\scriptstyle\pm.000}$ & $\mathbf{1.000}{\scriptstyle\pm.000}$ & $\mathbf{0.994}{\scriptstyle\pm.002}$ & $0.481{\scriptstyle\pm.004}$ \\
\midrule
Mahalanobis & $0.978{\scriptstyle\pm.001}$ & $0.090{\scriptstyle\pm.003}$ & $0.743{\scriptstyle\pm.003}$ & $0.785{\scriptstyle\pm.004}$ & $\mathbf{0.988}{\scriptstyle\pm.001}$ \\
Deep kNN & $0.994{\scriptstyle\pm.000}$ & $0.012{\scriptstyle\pm.002}$ & $0.588{\scriptstyle\pm.000}$ & $0.625{\scriptstyle\pm.001}$ & $0.973{\scriptstyle\pm.002}$ \\
Energy score & $0.850{\scriptstyle\pm.003}$ & $0.432{\scriptstyle\pm.017}$ & $0.636{\scriptstyle\pm.003}$ & $0.603{\scriptstyle\pm.001}$ & $0.873{\scriptstyle\pm.003}$ \\
\midrule
Softmax & $0.883{\scriptstyle\pm.004}$ & $0.413{\scriptstyle\pm.012}$ & $0.631{\scriptstyle\pm.000}$ & $0.558{\scriptstyle\pm.001}$ & $0.805{\scriptstyle\pm.005}$ \\
MC Dropout & $0.936{\scriptstyle\pm.002}$ & $0.313{\scriptstyle\pm.009}$ & $0.621{\scriptstyle\pm.001}$ & $0.545{\scriptstyle\pm.002}$ & $0.834{\scriptstyle\pm.004}$ \\
EDL & $0.883{\scriptstyle\pm.004}$ & $0.658{\scriptstyle\pm.032}$ & $0.547{\scriptstyle\pm.000}$ & $0.510{\scriptstyle\pm.000}$ & $0.767{\scriptstyle\pm.005}$ \\
\bottomrule
\end{tabular}}
\caption{Detection of distribution shifts (mean $\pm$ std, three evaluation seeds; higher AUROC / lower FPR@95 is better): OOD in both directions between MNIST and Fashion-MNIST, and clean-vs-corrupted separation in the mixed batch. The ablation row replaces conformity-based input opinions with the constant fully-trusted opinion; it is a single-seed run evaluated on the forward (MNIST) tasks only, hence the dashes and the absence of $\pm$. Conformity owns support violations, feature-space detectors own the subset direction, and the composed score degrades gracefully where its input factor fails.}
\label{tab:comparative}
\end{table}

On in-distribution data the two signals coincide: PaTAS matches the softmax filter on correct-vs-incorrect discrimination on clean test data ($0.979 \pm 0.003$ for both, MNIST), with equal calibration error after identical post-hoc calibration. Under corruption they separate: accuracy barely drops, so the softmax retains a higher correct-vs-incorrect AUROC ($0.974$ vs.\ $0.895$) precisely because it ignores the distribution violation that PaTAS flags; confidence ranks correctness \emph{within} the training distribution, while trust exposes inputs that \emph{leave} it. Directionality makes the division of labor explicit: in the reverse task (Fashion-MNIST in-distribution, MNIST as OOD), MNIST digits occupy a quiet subset of Fashion-MNIST's marginals rather than violating them, the conformity score collapses below chance ($0.481$), and the feature-space detectors win; conformity-based trust detects support \emph{violations}, not support \emph{subsets}. The composition fails safely there, the confidence factor keeping the composed score at the output-baseline level ($0.781$ vs.\ $0.805$ for softmax) instead of inheriting the collapse. Operationally, the composed score must not be used as an OOD filter in the subset direction, where on the CNN it even becomes anti-correlated ($0.201$, \appref{app:conv}); deployment should gate on the diagnostic of \xcref{tab:diagnostics}, and SL offers the natural remedy as future work: fusing in a Mahalanobis- or kNN-derived opinion as an additional input factor, since those detectors own exactly this direction.

\paragraph*{When propagation improves on simpler signals} For detecting inputs that violate the training support, conformity alone is the better detector (\cref{tab:comparative}) and should be used on its own. Propagation improves a decision in three situations: (i)~when the input is clean but the model is not, as for the poisoned-class gap of \cref{tab:mnistpoisipta} and the training-quality audit of \cref{tab:audit}, which no input-space or output-confidence signal expresses; (ii)~when the input factor fails, as in the subset direction, where the composed score stays at the output-baseline level ($0.781$) while conformity alone falls below chance ($0.481$); and (iii)~when violations actually cause errors, as on CIFAR-10, where the composed score is the strongest corruption detector ($0.732$ vs.\ $0.723$ for conformity alone, \xcref{tab:conv}).

\xcrefin{tab:diagnostics}{app:conformity} explains when this input-level signal exists. Fitting the conformity model on each dataset's training features yields a self-diagnostic: spatially registered domains (MNIST, Fashion-MNIST) concentrate evidence in a minority of tight-marginal ``reliable witness'' features, whose fraction predicts the ordering of the detection margins ($0.40 \to 0.972$, $0.18 \to 0.835$, $0.00 \to$ chance), while unregistered imagery (GTSRB, CIFAR-10) has uniformly wide marginals, so marginal conformity is predicted, and for GTSRB confirmed ($0.545$), to be uninformative. Applicability can thus be established per dataset before deployment; richer input-trust models for unregistered domains are future work.

The conformity opinions also remove the oracle assumption left open in \cref{exp:3}: feeding them, unchanged, into the poisoned model's IPTA flags patched inputs with no knowledge of the trigger. The per-sample input trust alone separates clean from patched inputs at AUROC $1.000$ for both classes, since the patch violates the near-constant border features the conformity model weights most; the propagated opinions acquire a distrust mass that clean inputs lack; and the parameter-side gap between the clean and the poisoned class persists under these estimated opinions ($b = 0.780$ vs.\ $0.499$; oracle assignment: $0.871$ vs.\ $0.559$), so input- and parameter-level trust compose end to end without externally supplied opinions. The remaining assumption, a clean feature source for the conformity fit, was also tested: fitting on the poisoned training set itself degrades trigger detection gracefully ($0.73/0.67$ per class, still far above chance), while a naive robust fit (median and MAD) does not recover ($0.65$), since MNIST pixel marginals are zero-inflated; contamination-aware conformity fits are explicit future work.

Finally, the model-side trust provides a training-quality audit that requires \emph{no test data at all}. Training the MNIST network under increasing label-flip rates $p$ and reading the PaTAS feedforward trust under a fully trusted input (\cref{tab:audit}) shows a tripwire behavior: trust drops by $31.7\%$ relative to the clean-trained model at \emph{any} non-zero corruption rate, including $p{=}0.1$ where test accuracy moves by only $3$ points. The flat level is designed saturation: at $p{=}0.1$ essentially every 128-sample batch already contains corrupted labels, so every batch delivers the large-gradient corrections \textsc{NodeTrust} counts as negative evidence, and the converged trust registers \emph{that} corruption is present rather than how much. The saturation is a property of the label opinion's informativeness, not of the calculus: when \cref{alg:trustupdate} consumes rate-informed label opinions $T_y = (1{-}p, 0, p)$ instead of vacuous ones, the same machinery yields a monotonically graded signal ($0.951 \to 0.805$, \cref{tab:audit}); tripwire versus graded meter is a choice of input opinion, not an architectural limit. Against data-centric severity estimators the comparison is honest in both directions: the fraction of training labels disagreeing with the model's confident predictions, in the spirit of confident learning~\cite{northcutt2021confidentlearning}, recovers $p$ almost exactly, and mean training loss grades likewise, but both require retaining the training data at audit time and neither expresses parameter-level trust, and mean confidence measures severity but needs clean held-out data; PaTAS instead detects \emph{that} training was corrupted from the trained trust state alone.

\begin{table}[ht]
\centering\footnotesize
\setlength{\tabcolsep}{5pt}
\begin{tabular}{ccccccc}
\toprule
$p$ & Acc.\ (\%) & Conf. & Trust & Trust$_{\mathrm{inf}}$ & Disagree & Train loss \\
\midrule
0   & 97.9 & 0.981 & 0.998 & 0.998 & 0.001 & 0.014 \\
0.1 & 94.7 & 0.841 & 0.682 & 0.951 & 0.099 & 0.419 \\
0.3 & 93.7 & 0.684 & 0.682 & 0.872 & 0.288 & 1.013 \\
0.5 & 82.8 & 0.476 & 0.681 & 0.805 & 0.461 & 1.472 \\
\bottomrule
\end{tabular}
\caption{Training-quality audit (MNIST, 784-512-10) under label-flip rate $p$. PaTAS trust (feedforward under a fully trusted input; no test data needed) is a corruption tripwire; Trust$_{\mathrm{inf}}$ uses rate-informed label opinions $(1{-}p, 0, p)$ and grades monotonically; Disagree (labels contradicting confident predictions) and train loss need the training data at audit time; confidence needs clean held-out data.}
\label{tab:audit}
\end{table}

\paragraph*{Convolutional architectures}
The battery transfers to CNNs through activity-weighted path conditioning (\cref{sec:convipta}); scoring never failed across all runs. On the $99.3\%$ MNIST CNN the composed score matches the softmax filter on clean correct-vs-incorrect ranking ($0.986$ for both) while dominating detection (corruption $0.988$ vs.\ $0.662$; OOD $0.991$ vs.\ $0.967$); on CIFAR-10, where corruption genuinely reduces accuracy, it is the strongest corruption detector outright ($0.732$); and the model-side audit transfers, with the label-noise-trained CIFAR model carrying path trust $0.521$ against $0.697$ clean-trained. Full results, per-method numbers, and caveats, including the subset-direction anti-correlation discussed above, are in \xcrefin{tab:conv}{app:conv}.

\vspace{-0.5em}
\section{Discussion}\label{sec:discussion}
\vspace{-0.2em}
\subsection{Discussion of PaTAS Findings}

Beyond per-output assessment, PaTAS can estimate a static trust opinion of the network as \(T(\text{NN}) = PaTAS_{FF}((1,0,0))\), on the principle that a trustworthy model preserves trust; under this definition the IPTA results of Experiment~\ref{exp:3} show clear degradation for adversarial samples (\cref{tab:mnistpoisipta}). Although trust mass and accuracy are often correlated, our results reveal meaningful divergences: in the breast cancer task (\cref{tab:cancer_results}) the mixed profile \((0.25,0,0.75)\) yields a higher final trust mass than the uncertain-features/trusted-label case (\(0.35\) vs.\ \(0.32\)) yet worse test accuracy (94\% vs.\ 96\%), and in poisoned MNIST (\cref{tab:mnistpoisipta}) clean digits \(3\) and \(6\) reach comparable accuracy but IPTA trust \(0.884\) vs.\ \(0.503\), robustly across ten trainings. Higher trust mass therefore does not imply superior predictive performance, and equal accuracy does not imply equal trustworthiness: PaTAS evaluates the reliability of the inference path, not predictive accuracy alone, and digit \(6\) was a poisoned class during training, so the parameters predicting it are less trusted than those for \(3\) even on clean test samples.

The finding that \emph{distrust is more damaging than uncertainty} is partly a property of the chosen operators, not of the data alone. Disbelief composes as \(d_{x \odot y} = d_x + d_y - d_x d_y\), so recurring distrust is absorbing across thousands of batch updates (Theorem~\xref{thm:geo}), and the conservative combination \(\oslash\) propagates the strongest warning by construction (\(d = \max\)); this also explains the rapid collapse to zero in the distrusted rows of \cref{tab:cancer_results}. We regard this as intended fail-safe semantics, but such comparisons should be read with this absorbing character in mind.

\paragraph*{Operational calibration} Absolute trust levels depend on architecture, depth, $\epsilon$, and the fusion variant, so deployment should establish a healthy reference level once by a proxy run under trusted assessments on the same architecture and act on relative drops; the tripwire of \cref{tab:audit} operates at a $31.7\%$ drop, far outside run-to-run variation ($\pm 0.001$ over ten trainings).

A key factor is the threshold \(\epsilon\) separating positive from negative gradient evidence: too small an \(\epsilon\) makes PaTAS read gradients as large even when the model performs well (\cref{tab:cancer_results}), resembling the \(g \to +\infty\) case in \appref{app:asymptotic}. A sweep over three orders of magnitude (\appref{app:ablations}) shows a broad plateau on MNIST ($\epsilon \in [0.05, 1]$) while the breast-cancer grid moves substantially between $0.01$ and $0.1$: $\epsilon$ is a scale parameter to be set relative to the task's prevailing gradient magnitudes, e.g., as a fixed multiple of the early-training per-layer gradient median. The revision-fusion variant is immaterial by comparison (differences $\leq 0.002$ across average, cumulative, and weighted fusion, \appref{app:ablations}); averaging is retained because Theorem~\xref{thm:arithm} guarantees non-degenerate long-run behavior, whereas cumulative fusion drives uncertainty toward dogmatic saturation.

\paragraph*{Obtaining input and label trust}\label{sec:acquisition}
Three routes provide the input- and label-level trust opinions consumed by PaTAS, in decreasing order of information. (i) \emph{Provenance-based assessment}: when data come from sources of known reliability, the quantification models of \cref{sec:back} map supporting and contradicting evidence directly into a binomial opinion, as~\cite{10.1007/978-3-032-19096-3_17} does for dataset trustworthiness and for regions known to be unreliable, such as an adversarial patch. (ii) \emph{Estimation from data}: label agreement across annotators or models and input-quality signals convert to opinions through the same evidence mapping; the conformity construction of \cref{exp:4} instantiates this route. (iii) \emph{Vacuous fallback}: absent any information, the opinion is initialized to \((0,0,1)\); since a vacuous input yields a vacuous output (\cref{thm:infvac}), the fallback is conservative, and PaTAS still contributes through parameter and path trust.

\emph{Without a reliable external assessment}, the consequences of this dependency are bounded. Under route~(iii) the input side contributes nothing, but the parameter side still functions: the audit of \cref{tab:audit} detects label corruption from the trained trust state under vacuous label opinions. Under route~(ii) the conformity opinions are informative in the support-violation regime that the diagnostic of \xcref{tab:diagnostics} identifies before deployment, and are predicted, and for GTSRB observed, to be uninformative outside it. Oracle opinions (route~(i)) are used only in Experiments~1 and~3 to isolate the propagation mechanism, and for Experiment~3 the estimated opinions of \cref{exp:4} preserve the parameter-side gap ($0.780$ vs.\ $0.499$). Imperfect estimates degrade the result gradually rather than silently: a conformity model fitted on the poisoned training set itself still detects the trigger well above chance ($0.73/0.67$).

\vspace{-0.5em}
\subsection{Limitations}\label{sec:limitations}
\paragraph*{Parameter trust} Parameter trust is validated at the level at which PaTAS consumes it, the class and the inference path: the poisoned-class gap of \cref{tab:mnistpoisipta} is stable over ten trainings and persists under estimated input opinions, and the audit of \cref{tab:audit} responds to any non-zero label corruption, far outside run-to-run variation. We do not claim that the opinion on an individual parameter measures that parameter's reliability in isolation: gradient magnitude is a proxy, confounded by learning-rate schedules, mini-batch noise, and saddle regions (\cref{sec:ptasfunc}). Per-parameter validation and a more direct evidence model remain future work.

\paragraph*{Input and label trust} The input-side value of PaTAS is bounded by the opinions it consumes (\cref{sec:acquisition}); conformity-based input trust detects violations of the training support, not inputs occupying a subset of it, and is uninformative on unregistered imagery, so the composed score must not serve as an OOD filter in the subset direction (\cref{exp:4}). Beyond the contaminated conformity fit, sensitivity to noisy input and label opinions remains to be studied.

\paragraph*{Magnitude blindness, scale, and cost}
The Trust Function combines opinions independently of weight and activation magnitudes: trust deliberately tracks the provenance and reliability of the quantities entering a computation rather than their functional influence, so an input the network effectively ignores can still contribute to the output opinion. Partial influence-aware steps exist (the softmax-weighted IPTA aggregation of \cref{exp:3}, the activity-weighted conv IPTA); fully magnitude-aware trust functions remain future work. On cost: one binomial opinion per parameter means a roughly threefold memory overhead; on an A100 (784-512-10), inference-time propagation costs $0.43$\,ms per batch, amortizing from $14.8\times$ overhead at batch size 1 to $3.7\times$ at 512, near the analytical $3\times$ estimate. During training, wall-clock is dominated by the synchronous gradient stream and revision (${\approx}7$ vs.\ ${\approx}2200$ steps/s unattached), an engineering bottleneck orthogonal to the method; revision can run as an offline replay after training, as done for all cached experiments.

\vspace{-0.5em}
\subsection{Threat Model and Security Scope}\label{sec:threat}
\vspace{-0.5em}

The asset PaTAS protects is the \emph{reliability of individual model predictions at inference time}. PaTAS is a \emph{detection and monitoring} mechanism, not a hardening one: it exposes predictions with weak supporting evidence without altering the model. We consider a \emph{training-time data-poisoning adversary}~\cite{chen2017targetedbackdoorattacksdeep} who can inject or relabel a bounded fraction of training samples, including fixed-pixel-patch backdoor triggers, to induce targeted misclassifications hidden behind high accuracy and confidence; the adversary controls neither the architecture, the training procedure, nor the trust assessments of known-provenance data (\cref{exp:3}). The defender has white-box access (parameters, activations, gradients) and trust opinions on inputs and labels, initialized to the vacuous opinion \((0,0,1)\) when unavailable, yielding conservative, uncertainty-dominated assessments rather than false assurance (\cref{sec:acquisition}). Evasion, model-stealing, and membership-inference attacks are out of scope; we prove consistency properties (\cref{sec:ptastheory}) rather than formal robustness guarantees.

\vspace{-0.5em}
\section{Conclusion}\label{sec:conclusion}
This paper presented \emph{PaTAS}, a Subjective Logic framework propagating trust in neural networks through a parallel structure of Trust Nodes and Trust Functions whose Parameter Trust Update and Inference-Path Trust Assessment capture how input quality, learned parameters, and activation paths jointly shape a prediction's trustworthiness.

On real-world and adversarial datasets, PaTAS detects trust degradation from data poisoning and adversarial patching while remaining interpretable and stable. In comparisons with output-only uncertainty baselines and dedicated input- and feature-space detectors, violation detection is carried by the conformity-derived input opinions, which alone outperform every dedicated detector there, while trust propagation contributes what no detector expresses: a test-data-free training audit, per-class trust gaps on poisoned models persisting without oracle patch knowledge, and graceful composition where a single signal fails; applicability is predictable a priori, and the mechanism carries to CNNs (\appref{app:conv}).

Several directions remain open: transformers and attention layers; input-trust constructions beyond marginal conformity, robust to contaminated training features; sensitivity to noisy initial opinions; the gradient-to-evidence mapping and per-parameter validation of parameter trust; provable robustness under bounded poisoning; activation-function effects; and adaptive, magnitude-aware quantification.
 
\section*{Acknowledgment}
This work was supported by the European Union's Horizon Europe research and innovation programme under grant agreement No.~101069688 (CONNECT).
\vspace{-1em}
\bibliographystyle{IEEEtran}
\bibliography{bibtex/bib/references, bibtex/bib/calib, bibtex/bib/ecmlbias}

\ifextended
\clearpage
\appendices
% \newpage
\onecolumn

\section{Subjective Logic Operators: Definitions and Examples}
\label{app:operators}
\begin{definition}[Fusion~\cite{8009820}]
\label{def:fusion-app}
Let \(A\) be an agent forming an opinion about a proposition \(X=x\) based on two information sources, \(P\) and \(Q\). The fused opinion is defined as:
\begin{equation}
    \omega^A_{X=x} = \omega^{P \& Q}_{X=x} = \omega^P_{X=x} \odot \omega^Q_{X=x}.
\end{equation}
The specific fusion operator depends on the relationship between the sources. SL defines several variants such as consensus, averaging, weighting, and cumulative fusion, each representing a distinct way of aggregating evidence. 

\textbf{Example.} Suppose two temperature sensors estimate whether the room temperature exceeds \(25^\circ\)C. If both sensors are of the same type and installed in the same place, their evidence is correlated and averaging or weighted fusion is appropriate. If they have different measurement strategies (e.g., infrared and contact-based), cumulative fusion is more suitable. The information from each independent sensor is treated as an additional, non-redundant contribution so that the combined evidence grows with each agreeing source.
\end{definition}

\begin{definition}[Trust Discounting~\cite{10706345}]
\label{def:discount}
Let \(A\) have a referral trust \(\omega^A_B\) in another agent \(B\), who holds an opinion \(\omega^B_X\) on variable \(X\). The trust-discounted opinion of \(A\) derived from \(B\)'s opinion is:
\(
    \omega^{[A;B]}_{X=x} = \omega^A_B \otimes \omega^B_{X=x} \notag
\). Referral trust is domain-specific and expresses how much \(A\) relies on \(B\) regarding \(X\).

\textbf{Example.} Consider a monitoring system where agent $A$ receives readings from a temperature sensor $B$. 
The sensor usually works well but is known to drift at times, so $A$ does not fully trust it. 
When $B$ reports that the temperature is above a safety threshold, $A$ discounts this opinion by reducing its strength and increasing uncertainty. 
This illustrates the principle of trust discounting: evidence from a partially reliable source is treated cautiously.
\end{definition}

\begin{definition}[Inferential Operators~\cite{josang2006trust}]
\label{def:inferential_ops}
Inferential operators generalize Bayesian reasoning by enabling opinion propagation through conditional relationships. 
Let \( A \) be an agent reasoning about a variable \( X \) and its potential implications for another variable \( Y \). Suppose \( A \) holds opinions \(\omega^A_{Y\mid X}\) on the relationship \(X \Rightarrow Y\).
The main operators are:
\begin{itemize}
    \item \textbf{Deduction:} derive $\omega^A_Y$ from $\omega^A_X$ using $\omega^A_{Y\mid X}$.
    \item \textbf{Abduction:} derive $\omega^A_X$ from $\omega^A_Y$ using $\omega^A_{Y\mid X}$.
\end{itemize}

\textbf{Example.} For the relation “if it rains ($X$), then Bob carries an umbrella ($Y$)”, agent $A$ holds a conditional opinion $\omega^A_{Y\mid X} = (b,d,u)$. Given an opinion on \(X\), deductive inference yields an opinion about \(Y\); conversely, abduction infers the likelihood \(X\) from \(Y\) observations.
\end{definition}

\section{Subjective Logic Operators and Notation for Parameter-Trust Update}\label{app:operator}
\begin{table}[H]
\centering
\caption{Summary of Subjective Logic Operators Used in This Work}
\label{tab:operators}
\renewcommand{\arraystretch}{1.3}
\begin{tabular}{|c|l|c|}
\hline
\textbf{Symbol} & \textbf{Name} & \textbf{Definition / Equation} \\ \hline

$\odot$ & Binomial multiplication &
\cite{josang2016subjective} \\ \hline

$\otimes$ & Trust discounting &
$\begin{aligned}
\omega^{[A;B]}_{X} &= \omega^A_B \otimes \omega^B_X \\
(b,d,u) \otimes (b',d',u') &= \left(Pb',\; Pd',\; 1-P\left(b'+d'\right)\right)\\
P &= b+au 
\end{aligned}$ \\ \hline

$\oplus$ & Fusion (averaging variant used throughout this work) &
\cite{8009820}\\ \hline

$\circleddash$ & Fusion-based revision &
$\begin{aligned}
T_{\theta} &\leftarrow T_{\theta} \circleddash T_{n \parallel Y}
\end{aligned}$ 
(as defined in Sec.~\ref{sec:ptasfunc}) \\ \hline

$\oslash$ & Conservative combination &
$\begin{aligned}
(b,d,u) &= (b_1,d_1,u_1) \oslash (b_2,d_2,u_2) \\
b &= \min(b_1,b_2), \\
d &= \max(d_1,d_2), \\
u &= 1-(b+d)
\end{aligned}$ \\ \hline

$\circledcirc $ & Deduction &
\cite{7528033,josang2016subjective}  \\ \hline

\end{tabular}
\end{table}

The conservative combination $\oslash$ ($b = \min(b_1, b_2)$, $d = \max(d_1, d_2)$, $u = 1 - b - d$) is not a standard SL operator; its algebraic behavior is elementary: it is commutative, associative, and idempotent (componentwise $\min$ and $\max$ are), and monotone in both arguments (increasing either belief cannot decrease the combined belief, increasing either disbelief cannot decrease the combined disbelief). It is used only to combine the feature and label opinions into the most conservative joint reading before revision.
\begin{table}[H]
\centering
\caption{Symbols Used in the Parameter-Trust Update Subsection and Algorithm}
\label{tab:params-algorithm-append}
\renewcommand{\arraystretch}{1.2}
\begin{tabular}{ll}
\hline
\textbf{Symbol} & \textbf{Meaning / Description} \\
\hline
\multicolumn{2}{l}{\textbf{Inputs to the Algorithm}} \\
$g$ & Collection of gradients for all parameters in the batch \\
$T_y$ & Trust opinion on label $y$ \\
$\epsilon$ & Gradient sensitivity threshold in \textsc{NodeTrust} \\[4pt]

\multicolumn{2}{l}{\textbf{Batch and Layer Quantities}} \\
$T_{y_{\text{batch}}}$ & Aggregated trust over labels in the current batch \\
$l$ & Layer index \\
$n_i^{(l)}$ & Neuron $i$ in layer $l$ \\
$\mathcal{N}(i)$ & Set of incoming edges to neuron $i$ \\[4pt]

\multicolumn{2}{l}{\textbf{Gradient Evidence}} \\
$g^{(l)}_{i,j}$ & Gradient of loss w.r.t.\ $\theta^{(l)}_{i,j}$ \\
$g^{(l)}_{i}$ & Gradient vector for neuron $n_i^{(l)}$ \\
$r$ & Count of weak gradients: $|g^{(l)}_{i,j}| < \epsilon$ \\
$s$ & Count of strong gradients: $|g^{(l)}_{i,j}| \ge \epsilon$ \\[4pt]

\multicolumn{2}{l}{\textbf{Trust Values Computed in the Algorithm}} \\
$T_{n_i \mid y_{\text{batch}}}$ & Trust in neuron $n_i$ conditioned on batch labels \\
$T_{n_i \mid \overline{y_{\text{batch}}}}$ & Trust in neuron under incorrect labels (vacuous) \\
$T_{n_i \parallel Y_{\text{batch}}}$ & Deduced trust in neuron $n_i$ \\[4pt]

\multicolumn{2}{l}{\textbf{Parameter Trust}} \\
$T_{\theta^{(l)}_{i,j}}$ & Trust opinion on parameter $\theta^{(l)}_{i,j}$ \\
% $T_{lr}$ & Trust in the learning rate \\
$T_{x^{(l-1)}_j}$ & Trust in the input feature to parameter $\theta^{(l)}_{i,j}$ \\[4pt]

\multicolumn{2}{l}{\textbf{Operators Used}} \\
$\bigwedge$ & Batch-wise fusion of trust opinions \\
$\circledcirc$ & Inferential deduction operator \\
$\circleddash$ & Trust-revision operator (fusion-based revision)\\
$\odot$ & SL binomial multiplication \\
$\oslash$ & Conservative combination operator \\
\hline
\end{tabular}
\end{table}

\section{Symbols and Parameters Used in PaTAS}\label{app:symbols}
\begin{table}[H]
\centering
\caption{List of Symbols and Parameters Used in this work}
\label{tab:params-appendix}
\label{tab:params-full}
\renewcommand{\arraystretch}{1.2}
\begin{tabular}{ll}
\hline
\textbf{Symbol} & \textbf{Meaning / Description} \\
\hline
\multicolumn{2}{l}{\textbf{Neural Network Quantities}} \\
$x$ & Input feature vector \\
$y$ & Ground-truth label \\
$y' = f_\Theta(x)$ & Output of the neural network \\
$\Theta = (W,b,f)$ & Neural-network parameters (weights, biases, activations) \\
$W^{(l)}$ & Weight matrix at layer $l$ \\
$b^{(l)}$ & Bias vector at layer $l$ \\
$\phi^{(l)}(\cdot)$ & Activation function at layer $l$ \\
$z^{(l)}$ & Pre-activation vector of layer $l$ \\
$x^{(l)}$ & Activation vector of layer $l$ \\
$n_i^{(l)}$ & Neuron $i$ in layer $l$ \\
$\theta^{(l)}_{i,j}$ & Parameter from neuron $j$ in layer $l\!-\!1$ to neuron $i$ in layer $l$ \\
$\delta^{(l)}_i$ & Backpropagated error signal of neuron $i$ in layer $l$ \\
$x^{(l-1)}_j$ & Activation of neuron $j$ in the previous layer \\
$\mathcal{N}(i)$ & Set of incoming neighbors of neuron $i$ \\[4pt]

\multicolumn{2}{l}{\textbf{Gradients and Learning Dynamics}} \\
$g$ & Collection of gradients for all network parameters \\
$g^{(l)}_{i,j}$ & Gradient w.r.t.\ weight $\theta^{(l)}_{i,j}$ \\
$g^{(l)}_i$ & Vector of gradients for all incoming parameters of neuron $n_i^{(l)}$ \\
$l_r$ (or $\mathrm{lr}$) & Learning rate \\
$\mathcal{L}$ & Loss function \\[4pt]

\multicolumn{2}{l}{\textbf{Trust Assessments and Trust Nodes}} \\
$T(x)$ & Trust assessment of input features \\
$T_y$ & Trust opinion on label $y$ \\
$T_{y_{\text{batch}}}$ & Aggregated trust opinion over all labels in a batch \\
$T_{\theta^{(l)}_{i,j}}$ & Trust opinion on parameter $\theta^{(l)}_{i,j}$ \\
$T_{x^{(l)}_i}$ & Trust opinion on activation $x^{(l)}_i$ \\
$T_{x^{(l)}_j}$ & Trust opinion of feature contributing to edge $(j \to i)$ \\
% $T_{lr}$ & Trust opinion on the learning rate \\
$T_{n_i \mid y_{\text{batch}}}$ & Trust in neuron $n_i$ conditioned on batch labels \\
$T_{n_i \mid \overline{y_{\text{batch}}}}$ & Trust in neuron given incorrect labels (vacuous opinion) \\
$T_{n_i \parallel Y_{\text{batch}}}$ & Deduced trust in neuron $n_i$ after combining conditional evidence \\[4pt]

\multicolumn{2}{l}{\textbf{Subjective Logic Opinions and Evidence}} \\
$\omega = (b,d,u,a)$ & SL binomial opinion: belief, disbelief, uncertainty, base rate \\
$r$ & Positive evidence count (from weak gradients) \\
$s$ & Negative evidence count (from strong gradients) \\
$\epsilon$ & Gradient sensitivity threshold in \textsc{NodeTrust} \\[4pt]

\multicolumn{2}{l}{\textbf{Trust Operators (SL Operators + PaTAS-specific)}} \\
$\oplus$ & SL fusion operator (cumulative or averaging fusion) \\
$\otimes$ & SL trust-discounting operator \\
$\odot$ & SL binomial multiplication \\
$\oslash$ & Conservative combination operator used in parameter updates \\
$\circledcirc$ & Inferential deduction operator in PaTAS \\
$\circleddash$ & Trust-revision operator (fusion-based revision) used for parameter updates \\
$\bigwedge$ & Batch-wise fusion of trust opinions \\[4pt]

\multicolumn{2}{l}{\textbf{PaTAS System Components}} \\
$\mathrm{Pf}_\Theta$ & Parallel Trust Function (PaTAS trust feedforward) \\
\text{TNN} & Trust Nodes Network (parallel trust architecture) \\
\text{GenIPTA} & Generator of the Inference-Path Trust Assessment \\
\text{IPTA} & Inference-Path Trust Assessment for a single inference \\
\hline
\end{tabular}
\end{table}

\section{Theorems and Proofs (\cref{sec:ptastheory})}\label{app:proof}
\subsection{Key theorems}
\begin{theorem}[Convergence of Averaging-Fusion Revision]\label{thm:arithm}
Let $\Omega = \{(b,d,u,a) \in [0,1]^4 : b+d+u = 1\}$ with the base rate $a$ held fixed, let $q = (b_q, d_q, u_q, a) \in \Omega$, and let the sequence $(\omega_n)$ be defined by the averaging-fusion revision used in our implementation,
\[
b_{n+1} = \frac{b_n u_q + b_q u_n}{u_n + u_q}, \qquad
d_{n+1} = \frac{d_n u_q + d_q u_n}{u_n + u_q}, \qquad
u_{n+1} = \frac{2\, u_n u_q}{u_n + u_q}.
\]
If $u_0 > 0$, then $\omega_n \to q$, and the convergence is geometric: with $m = \min(u_0, u_q)$,
\[
|b_{n+1} - b_q| = \lambda_n\, |b_n - b_q|, \qquad
\lambda_n = \frac{u_q}{u_n + u_q} \;\leq\; \frac{u_q}{m + u_q} < 1,
\]
and likewise for $d_n$; the uncertainty $u_n$ converges to $u_q$ monotonically, and $\lambda_n \to 1/2$. Two boundary cases complete the picture: if $u_q = 0$ and $u_0 > 0$, the sequence reaches $q$ in one step; if $u_0 = 0$, the sequence is constant, since dogmatic opinions are fixed points of averaging fusion. PaTAS avoids the latter case by initializing all parameter opinions as vacuous ($u_0 = 1$).
\end{theorem}

\begin{proof}
If $u_q = 0$ and $u_n > 0$, the update gives $b_{n+1} = b_q$, $d_{n+1} = d_q$, $u_{n+1} = 0$, so $\omega_1 = q$. If $u_0 = 0$, the update gives $b_{n+1} = b_n$, $d_{n+1} = d_n$, $u_{n+1} = 0$ for all $n$. Assume now $u_0 > 0$ and $u_q > 0$.

\emph{Uncertainty.} $u_{n+1}$ is the harmonic mean of $u_n$ and $u_q$, which lies strictly between $\min(u_n, u_q)$ and $\max(u_n, u_q)$ whenever $u_n \neq u_q$: indeed $u_{n+1} > u_q \iff 2 u_n > u_n + u_q \iff u_n > u_q$, and symmetrically for $u_n < u_q$. Hence $(u_n)$ is monotone toward $u_q$, bounded, and bounded below by $m = \min(u_0, u_q) > 0$. It converges to a limit $L \geq m$ satisfying $L = 2 L u_q/(L + u_q)$, i.e., $L (L - u_q) = 0$, hence $L = u_q$.

\emph{Belief and disbelief.} Subtracting $b_q$ from the belief update,
\[
b_{n+1} - b_q
= \frac{b_n u_q + b_q u_n - b_q (u_n + u_q)}{u_n + u_q}
= \frac{u_q}{u_n + u_q}\, (b_n - b_q),
\]
so the error contracts at every step by $\lambda_n = u_q/(u_n+u_q) \leq u_q/(m+u_q) < 1$, giving $|b_n - b_q| \leq \big(\tfrac{u_q}{m+u_q}\big)^n |b_0 - b_q| \to 0$. The same identity holds for $d_n$, the base rate is unchanged by fusion, and $u_n \to u_q$ was shown above. Hence $\omega_n \to q$, and since $u_n \to u_q$ the per-step rate satisfies $\lambda_n \to 1/2$.
\end{proof}

Averaging fusion is therefore affected by a vacuous revision opinion: for $q = (0,0,1,a)$ the theorem yields $\omega_n \to q$, i.e., repeated revision with fully uncertain evidence progressively erodes committed mass, in contrast to cumulative fusion, for which the vacuous opinion is a neutral element. No group structure on $(\Omega, \oplus)$ is required or assumed.

\begin{theorem}[Convergence of Subjective Logic Geometric Sequence]\label{thm:geo}
Let $\Omega = \{(b,d,u,a) \in [0,1]^4 : b+d+u = 1\}$ be the opinion space of Theorem~\ref{thm:arithm}, here with the base rate evolving under the operator (no algebraic structure beyond closure of $\odot$ on $\Omega$ is assumed).
Let $\omega_n = (b_n, d_n, u_n, a_n) \in \Omega$ be a sequence defined recursively by the operator $\odot$ as follows:
\[
\omega_{n+1} = \omega_n \odot q,
\]
where \(q\in \Omega\) and the operator $\odot$ is the binomial multiplication defined by:
\[
\begin{cases}
b_{x \odot y} = b_x b_y + \dfrac{(1 - a_x) a_y b_x u_y + a_x (1 - a_y) u_x b_y}{1 - a_x a_y}, \\
d_{x \odot y} = d_x + d_y - d_x d_y, \\
u_{x \odot y} = u_x u_y + \dfrac{(1 - a_y) b_x u_y + (1 - a_x) u_x b_y}{1 - a_x a_y}, \\
a_{x \odot y} = a_x a_y.
\end{cases}
\]
Then the sequence $(\omega_n)$ converges in $[0,1]^4$ if $a_q < 1$. In particular:
\begin{itemize}
    \item $a_n = a_0 a_q^n \to 0$ as $n \to \infty$,
    \item $d_n \to 1$ if $d_q > 0$, and $d_n = d_0$ if $d_q = 0$,
    \item $b_n \to 0$ if $p_q = b_q + a_q u_q < 1$; if $p_q = 1$, $(b_n)$ converges to a limit within $a_0/(1-a_q)$ of $b_0$.
\end{itemize}
\end{theorem}

Theorem~\ref{thm:geo} analyzes the operator with Jøsang's base-rate dynamics $a_{x \odot y} = a_x a_y$. The implementation instead holds every base rate fixed at $0.5$ through all operators (the convention stated with the experiments), which only strengthens the conclusion: with $a_n = a$ constant, the remainder bound in the proof holds with $a_n$ replaced by the constant, and the same limits follow.

\begin{proof}
We analyze each component of $\omega_n$ separately.

\emph{1. Convergence of $a_n$:}  
By definition, $a_{n+1} = a_n a_q$. Since $a_q \in [0,1[$, this is a geometric sequence:
\[
a_n = a_0 a_q^n \to 0 \quad \text{as } n \to \infty.
\]

\emph{2. Convergence of $d_n$:}  
The recurrence relation is:
\[
d_{n+1} = d_n + d_q - d_n d_q = d_n (1 - d_q) + d_q.
\]
This is a first-order linear recurrence. If $d_q > 0$, the sequence is increasing and bounded above by 1. Therefore:
\[
\lim_{n \to \infty} d_n = 1\, (\text{the fixed point})
\]
If $d_q = 0$, then $d_{n+1} = d_n = d_0$ for all $n$.

\emph{3. Convergence of $b_n$:}
Separating the terms proportional to $a_n$ in the exact update gives
\[
b_{n+1} = p_q\, b_n + \delta_n, \qquad
\delta_n = \frac{a_n (1 - a_q)}{1 - a_n a_q}\,\big(u_n b_q - a_q b_n u_q\big),
\qquad p_q = b_q + a_q u_q,
\]
and since $1 - a_n a_q \geq 1 - a_q$ and $|u_n b_q - a_q b_n u_q| \leq 1$, the remainder is bounded by $|\delta_n| \leq a_n = a_0 a_q^n$, which is summable. If $p_q < 1$, unrolling the recursion yields $0 \leq b_n \leq p_q^n b_0 + \sum_{k<n} p_q^{\,n-1-k} a_0 a_q^k \to 0$, a sum of two vanishing geometric terms, so $b_n \to 0$. If $p_q = 1$, then $b_{n+1} = b_n + \delta_n$ with $\sum_k |\delta_k| \leq a_0/(1-a_q) < \infty$, so $(b_n)$ is Cauchy and converges, with limit within $a_0/(1-a_q)$ of $b_0$.

Finally, $u_n = 1 - (b_n + d_n)$ converges as the difference of convergent sequences.
\end{proof}

\subsection{Proof for Theorem~\ref{thm:convergence}}
\begin{proof}
The proof operates under the idealization stated in the theorem: the network has been trained to convergence and the batch gradients \( g \) have stabilized, so that the revision opinion applied to each parameter is fixed across iterations. Mini-batch SGD does not provide exact gradient stabilization; the empirical counterpart of this idealization, bounded oscillation of the trust trajectories after accuracy convergence, is documented in Appendix~\ref{app:expres}.

Moreover, the stability of \( T_x \), and \( T_y \)implies that trust inputs to the update mechanism are stable. Consequently, each new trust update for \( T_{\theta} \) is computed using consistent and bounded evidence, which over time results in the stabilization of the subjective opinions assigned to \( T_{\theta} \).

In detail:
\begin{itemize}
    \item \(T_x\) converging implies (\(T_{y\prime}\) converging assuming the same \(T_\theta\)) implies convergence of \(T_{y_{batch}}\).

    \item \(g\) converges so \(T_{n_i \mid y_\text{batch}}\) will remain the same. 
    
    \item since \(T_{n_i\mid \mid {Y_\text{batch}}}\) is a deterministic calculation from the convergent terms  \(T_{n_i \mid y_\text{batch}}\) and \(T_{y_{batch}}\), it follows that \(T_{n_i\mid \mid {Y_\text{batch}}}\) also converges 

    \item Finally \(T_{\theta} \longleftarrow T_{\theta} \circleddash T_{n_i\mid \mid {Y_\text{batch}}}\) converges if \(\circleddash\) is set to any fusion operator or the binomial multiplication operator (see Theorem~\ref{thm:arithm} and \ref{thm:geo} in Appendix~\ref{app:proof}). 

    \item The function \(f_{upd}\) in our implementation is based on \(T_{\theta}\), \(T_{y}\) and \({T_{x}}\) which all converge.
\end{itemize}

Therefore, assuming convergence of $T_x$, $T_y$, and $g$, the trust values for all PaTAS parameters stabilize as training progresses, proving convergence of PaTAS creation. 
\end{proof}

\emph{Scope.} The convergence statement covers the revision dynamics: Theorem~\ref{thm:arithm} applies to the $\circleddash$ step with a stabilized revision opinion. Each batch additionally applies the multiplicative adjustment of \cref{eq:layerprop}-style operators (Eq.~(8)); the fixed point and contraction of the composed revise-then-multiply map are not established here and are left open. Empirically the composed iteration is stable (Appendix~\ref{app:expres}); a fixed-point analysis of the composed map is future theoretical work.

\subsection{Proof for \cref{thm:infvac}, Part 1 (Vacuity)}

Let \( \omega^\emptyset = (0, 0, 1, a) \) denote a vacuous binomial opinion over any variable, with arbitrary base rate \( a \in [0, 1] \). Then the following two properties hold:

\begin{enumerate}
    \item \emph{Discounting a Vacuous Opinion Yields a Vacuous Opinion.} \\
    For any trust value binomial opinion \(\omega^A_B \), the discounted opinion
    \[
    \omega^{[A;B]}_X = \omega^A_B \otimes (0,0,1,a)= (0,0,1,a)
    \]
    \textit{Proof.} Using the trust discounting operator from Subjective Logic, we set $P$ to the projected probability of \(\omega^A_B\):
    \[\omega^{[A;B]}_X = 
    \begin{cases}
    b^{[A;B]}_X = P\cdot 0 = 0, \\
    d^{[A;B]}_X = P\cdot 0 = 0, \\
    u^{[A;B]}_X = 1 - b^{[A;B]}_X-d^{[A;B]}_X = 1, \\
    a^{[A;B]}_X = a.
    \end{cases}
    \]
    Hence, \( \omega^{[A;B]} = (0,0,1,a) \).

    \item \emph{PaTAS Feedforward on Vacuous Input Yields Vacuous Output.} \\
    Let \( T_x = (0,0,1,a)\) be the trust assessment of an input to PaTAS. Then for any parameter trust configuration \( T_\theta \), the output trust assessment satisfies:
    \[
    T_y = \text{IPTA}(T_x) = (0,0,1,a).
    \]
    \textit{Proof.} The PaTAS feedforward mechanism computes for each neuron (\cref{eq:layerprop}):
    \[
    T_z = \bigoplus_i \left( T_{\theta_i} \otimes T_{x_i} \right).
    \]
    Since each \( T_{x_i} = (0,0,1,a) \) sits in the discounted (second) position, the result from part (1) applies with \(\omega^A_B = T_{\theta_i}\):
    \[
    T_{\theta_i} \otimes T_{x_i} = (0,0,1,a), \quad \forall i.
    \]
    Then, by fusion of vacuous opinions:
    \[
    T_z = \bigoplus_i (0,0,1,a) = (0,0,1,a).
    \]
    This holds recursively through all layers of PaTAS, including the output layer, hence:
    \[
    T_y = (0,0,1,a).
    \]
\end{enumerate}

\subsection{Proof for \cref{thm:infvac}, Part 2 (Symmetry)}
\begin{proof}
We prove the theorem in two steps.

\emph{(1) Symmetry of the Discount Operator:} \\
Let $\omega_\theta = (b_\theta, d_\theta, u_\theta)$ be a binomial opinion representing the trust weight associated with a connection in the PaTAS. Let $P$ denote the projected probability of $\omega_\theta$, defined as:
\[
P = b_\theta + a u_\theta,
\]
where $a$ is the base rate (typically $a = 0.5$ for binary domains).

Let $x = (b, d, u)$ be any binomial opinion and $\bar{x} = (d, b, u)$ its symmetric counterpart. Then the trust discounting operation yields:
\[
\omega_\theta \otimes x = (P \cdot b, P \cdot d, 1 - P \cdot (b + d)),
\]
\[
\omega_\theta \otimes \bar{x} = (P \cdot d, P \cdot b, 1 - P \cdot (b + d)).
\]
Since $b + d = 1 - u$, both discounted opinions have identical uncertainty and symmetric belief/disbelief masses. Hence, $\omega_\theta \otimes x$ and $\omega_\theta \otimes \bar{x}$ are symmetric.

\emph{(2) Symmetry Preservation under Fusion:} \\
Let $\mathcal{X} = \{x_1, \dots, x_n\}$ be a set of discounted opinions resulting from symmetric inputs, and let $\bar{\mathcal{X}} = \{\bar{x}_1, \dots, \bar{x}_n\}$ be their symmetric counterparts. Consider any symmetric fusion operator $\bigoplus$ (such as averaging, cumulative fusion, or consensus fusion in Subjective Logic) applied to $\mathcal{X}$ and $\bar{\mathcal{X}}$.

Since each pair $(x_i, \bar{x}_i)$ is symmetric and the operator treats belief and disbelief symmetrically, we have:
\[
\bigoplus_{i=1}^n x_i = (b', d', u') \quad \Rightarrow \quad \bigoplus_{i=1}^n \bar{x}_i = (d', b', u').
\]
Thus, the output of the PaTAS feedforward inference remains symmetric when symmetric inputs are provided.

\emph{(3) Invariance under Fusion:} \\
A fundamental property of Subjective Logic fusion operators is that if all input opinions assign the same value to a specific mass (e.g., belief or disbelief), the result will preserve that value. In particular, if all input opinions have belief mass equal to zero, the fused result will also have belief mass zero. The same applies to disbelief mass.

Therefore, if the discounting step yields discounted opinions with zero belief (or zero disbelief) across all components, then the fusion stage will preserve that zero mass in the final output.

\emph{(4) Fully Trusted and Distrusted Cases:} \\
For $x = (1, 0, 0)$ (fully trusted), we have:
\[
\omega_\theta \otimes x = (P, 0, 1 - P),
\]
so the discounted opinion assigns zero disbelief. As noted above, the fusion of such discounted opinions will also assign zero disbelief.

For $\bar{x} = (0, 1, 0)$ (fully distrusted), we have:
\[
\omega_\theta \otimes \bar{x} = (0, P, 1 - P),
\]
so the discounted opinion assigns zero belief. Therefore, the belief from a fully distrusted input is always zero in the PaTAS output.
\end{proof}

\section{Asymptotic Behavior of the Deduced Parameter Trust}
\label{app:asymptotic}

Table~\ref{tab:asymptotic} reports the asymptotic behavior 
of $T_{\theta_i \| Y_\text{batch}}$ as a function of the 
gradient magnitude $g$ and the label trust $T_{y_\text{batch}}$. 
When $g \to 0$ the neuron is assessed as Trusted regardless 
of label trust, reflecting that vanishing corrections indicate 
a well-settled parameter. When $g \to +\infty$ and labels are 
Trusted, the parameter is assessed as DisTrusted, since the 
trusted data must repeatedly and strongly correct it. Under 
Uncertain or DisTrusted labels the same large gradient produces 
a more conservative assessment, confirming that negative 
evidence is discounted in proportion to label unreliability.

\begin{table}[h]
\centering
\caption{Asymptotic behavior of $T_{\theta_i \| Y_\text{batch}}$}
\label{tab:asymptotic}
\begin{tabular}{ccc}
\toprule
$g \to$ & $T_{y_\text{batch}} \to$ & 
    $T_{\theta_i \| Y_\text{batch}} \to$ \\
\midrule
$0$ & Trusted      & Trusted      \\
    & Uncertain    & Uncertain    \\
    & DisTrusted   & Uncertain    \\
\midrule
$+\infty$ & Trusted    & DisTrusted       \\
          & Uncertain  & $(0, 0.5, 0.5)$  \\
          & DisTrusted & Uncertain        \\
\bottomrule
\end{tabular}
\end{table}

\section{Perceptron Case: Full Derivation (\cref{sec:perceptron})}\label{app:perceptron}
This appendix gives the derivation of \cref{lem:perceptron}. Throughout, we use a two-feature perceptron estimating the cost of renting an apartment,
\begin{equation}
    y\prime = 10 \times s + 100 \times n_r,
    \label{equ:smodel}
\end{equation}
where \(s\) is the size of the apartment and \(n_r\) the number of rooms, and we treat two sub-problems in turn: how trust in \(s\) and \(n_r\) propagates to \(y\prime\), and how trust in the parameters \(\theta_1 = 10\) and \(\theta_2 = 100\) refines that propagation.

\begin{figure}[H]
    \centering    \includegraphics[scale=0.3]{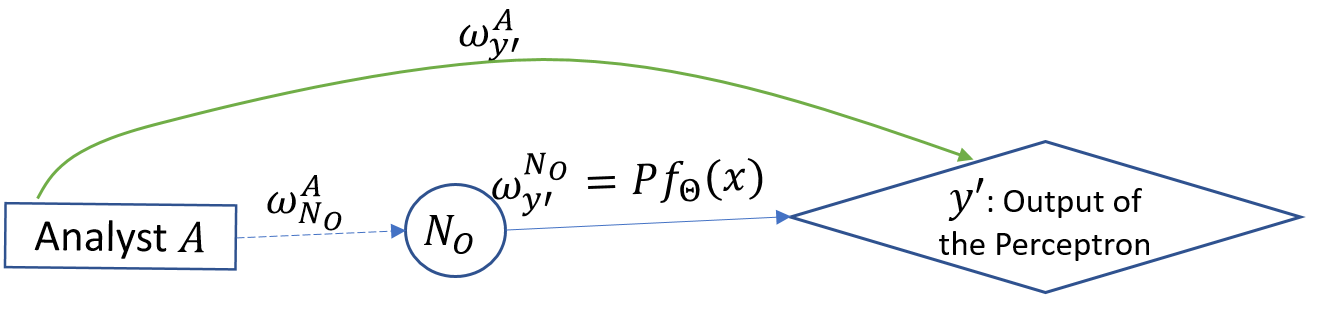}
    \caption{STN used to assess the output \(y\prime=f(x)\) of a perceptron \(\Theta\)}
    \label{fig:pbstate}
\end{figure}

% \subsection{Solution for the Perceptron}
\subsection{Trust Propagation from Input to Output}
% \todo[inline]{Describe Solution for Problem 1}
\begin{figure}
\centering
\subfloat[Perceptron for \cref{equ:smodel}]{
    \includegraphics[width=0.35\textwidth]{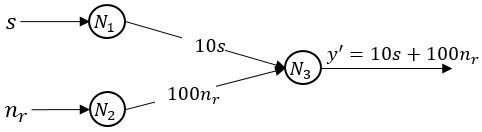}
     \label{fig:nn1}
    }
   
\hfill
\subfloat[Parallel Trust Function for \cref{equ:smodel}]{
    \includegraphics[width=0.45\textwidth]{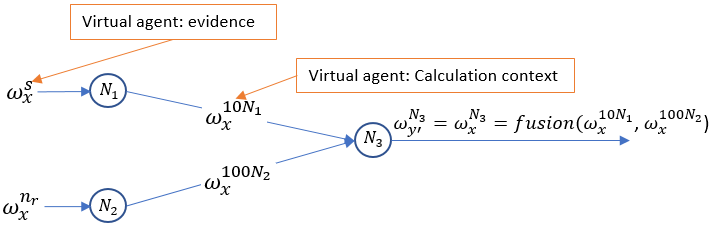}
   \label{fig:nn1s}}
\caption{Trust-propagation based solely on input-feature trust}
\label{fig:solution1}
\end{figure}
\noindent\paragraph*{Objective} Determine how trust in the individual input features (\(s\) and \(n_r\)) propagates through a simple perceptron to produce a trust assessment on the output variable \(y\prime\), representing the predicted apartment cost.
\paragraph*{Assumptions} For now, we assume that the model parameters \(\theta_1=10\) and  \(\theta_2=100\) are fully trusted and input feature trust opinions (\(T_s\) and \(T_{n_r}\)) are available.

The model specified in \cref{equ:smodel} is equivalent to the NN depicted in \cref{fig:nn1}. This network employs a linear transformation without an activation function. The input vector is:
$$ 
x = \begin{pmatrix}
    s \\
    n_r
\end{pmatrix}
$$ 
Let $\omega_x^s$ be the trust opinion on $x$ based on $s$ as evidence and  $\omega_x^{n_r}$ the trust opinion on $x$ based on $n_r$ as evidence. Since the output neuron $N_3$ computes \cref{equ:smodel}, we associate this computation with two agents: $10N_1$ from the context of computing \(10\cdot s\) and $100N_2$ from the context of computing \(100\cdot n_r\). The trust opinion of the output neuron \(N_3\) on \(x\) is then: 
\(\omega_x^{N_3} = \mathit{fusion}(\omega_x^{10N_1}, \omega_x^{100N_2})\). 

The choice of the fusion operator (that we will later denote by \(\oplus\) ) depends on the semantics of \(s\) and \(n_r\) (see Definition~\ref{def:fusion-app}). In this example, since $s$ and $n_r$ represent independent evidence, the fusion operator to use is cumulative fusion.

Assuming full trust in the parameters \(\theta_1 = 10\) and \(\theta_2 = 100\), we have:
\begin{align}
    \omega_x^{10N_1} = \omega_x^{N_1} = \omega_x^{s} = T_s \, \text{ (Trust in the feature \(s\) of \(x\))} \notag \\
    \omega_x^{100N_2} = \omega_x^{N_2} = \omega_x^{n_r} = T_{n_r} \, \text{ (Trust in the feature \(n_r\) of \(x\))} \notag 
\end{align}

Since the output value is deterministically related to the input via
\[
y' = f(x),
\]
the trust opinion of \(N_3\) on \(y'\) is the same as the trust opinion already computed for \(x\). Clearly, the transformation performed by \(f\) is encoded in the agent algebra, not in the variable algebra. Thus,
\begin{align}
    \omega_{y\prime}^{N_3} &= \omega_x^{N_3} = \omega_x^{10N_1} \oplus \omega_x^{100N_2} \notag \notag \\
&= \omega_x^{s} \oplus \omega_x^{n_r} = T_s \oplus T_{n_r}
\end{align}
In summary, if we define \(T_x = \begin{pmatrix}
    T_s&T_{n_r}
\end{pmatrix} \), then:
\[\pf_\Theta(T_x) = \pf_\Theta(\begin{pmatrix}
    T_s&T_{n_r}
\end{pmatrix}) =  T_s \oplus T_{n_r} \]

\subsection{Impact of Parameter Trust}
\paragraph*{Objective} Analyze how trust in the model parameters \(\theta_1=10\) and \(\theta_2=100\) affects the resulting trust assessment on the output \(y\prime\), given trust in the inputs.
\paragraph*{Assumption} Trust Opinions \(T_{\theta_1}\) and \(T_{\theta_2}\) are assigned to the parameters. The approach used to calculate these parameter-trust values will be discussed in Section~\ref{sec:patas}.

In problem 2, the model is expressed as:
\begin{equation}
    y\prime = \theta_1\times s + \theta_2\times n_r
    \label{equ:smodel2}
\end{equation} 
and we assume trust parameters \(T_{\theta_1}\) and \(T_{\theta_2}\) respectively in \(\theta_1\) and \(\theta_2\) (we'll see in details in Section~\ref{sec:patas} how to calculate these trust parameters). Unlike the previous problem, where we fully trusted \(\theta_1\) and \(\theta_2\), here we do not fully trust these parameters, and we must account for their trust assessments. For the network to incorporate the trust in \(\theta_1\) and \(\theta_2\), we adjust the trust opinions on the features accordingly. The trust opinions are now refined.
\begin{figure}
    \centering
    \includegraphics[width=0.4\linewidth]{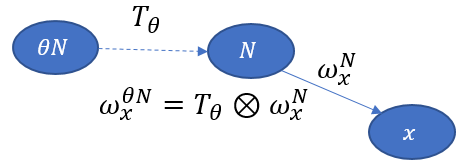}
    \caption{Trust-propagation Trust Network with parameter trust.}
    \label{fig:multprob2l}
\end{figure}
As depicted in \cref{fig:multprob2l}, we model this as a small subjective network. therefore:
\begin{align}
    \omega_x^{\theta_1 N_1} = T_{\theta_1} \otimes \omega_x^{N_1} \; \text{and} \;
    \omega_x^{\theta_1 N_2} =  T_{\theta_2} \otimes \omega_x^{N_2}
\end{align}
where \(\otimes\) is a trust discounting operator. This result is consistent with the solution for problem 1 as for fully trusted \(T_\theta=(1,0,0)\), we have \(\omega_x^{\theta N} = T_{\theta} \otimes \omega_x^{N} = \omega_x^{N}\) (for any neuron \(N\)).

Finally, the resulting output trust \(\omega_{y\prime}^{N_3}\) is calculated by the fusion of these adjusted trust opinions: 
\begin{equation}
\label{eq:sol222}
    \pf_\Theta(T_x) = \pf_\Theta(\begin{pmatrix}
    T_s&T_{n_r}
\end{pmatrix}) =  (T_{\theta_1}\otimes T_s)\oplus (T_{\theta_2}\otimes T_{n_r})
\end{equation}

Thus, we adjust the trust in the network output based on the trust in both the parameters and the features, ensuring that the trust propagation takes into account the trust in the parameters.

\section{Trust Node Structure}\label{app:trustnode}
\begin{figure}[H]
\centering
\begin{subfigure}{0.24\textwidth}
\includegraphics[width=\textwidth]{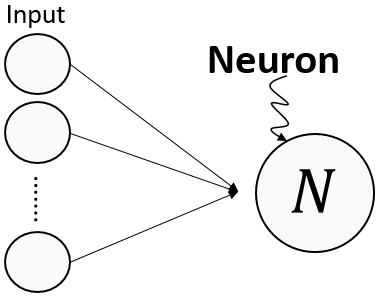}
  \caption{Structure of a Perceptron}
  \label{fig:neuron}
\end{subfigure}
\begin{subfigure}{0.24\textwidth}
  \includegraphics[width=\textwidth]{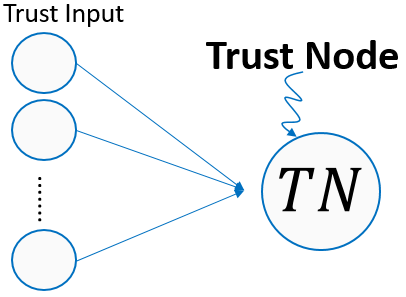}
  \caption{Mirrored Trust Structure}
\end{subfigure}

\begin{subfigure}{0.48\textwidth}
  \includegraphics[width=\textwidth]{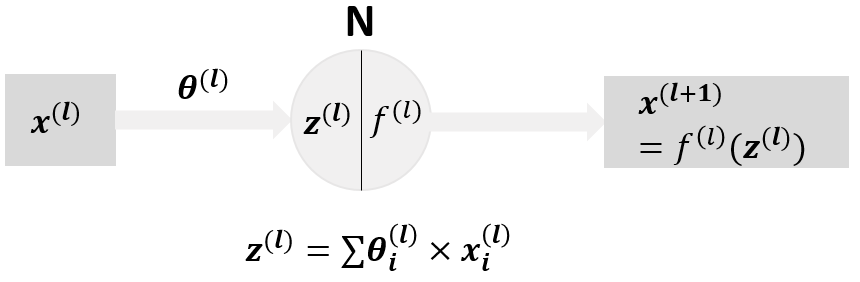}
  \caption{Perceptron Operation}
  \label{fig:neuronfunc}
\end{subfigure}
\begin{subfigure}{0.48\textwidth}
  \centering
  \resizebox{\textwidth}{!}{\tnopfigure}
  \caption{Trust Function Operation}
\end{subfigure}
\caption{Trust Node structure and computation, shown alongside the corresponding neuron operations.}
\label{fig:trustnodefull}
\end{figure}

\section{PaTAS Workflow and Functional Flow}\label{app:flow}
\begin{figure}[H]
    \centering
    \includegraphics[width=0.4\linewidth]{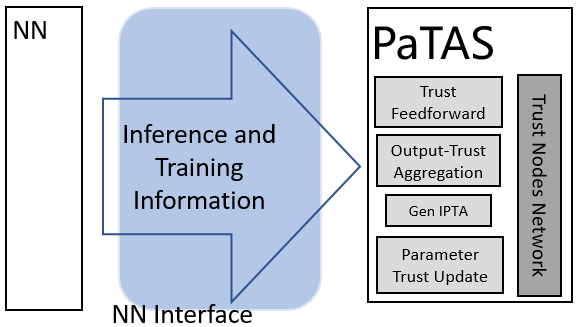}
    \caption{High-Level Overview of PaTAS Integration with a Neural Network}
    \label{fig:highptas}
\end{figure}

\begin{figure}[H]
    \centering
    \includegraphics[width=0.75\linewidth]{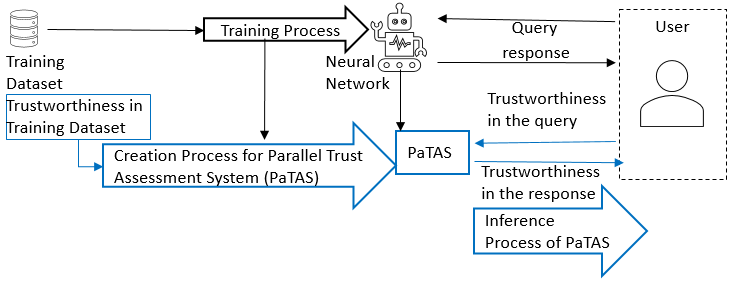}
    \caption{Overview of the PaTAS workflow.}
    \label{fig:ovvwptas}
\end{figure}

\begin{figure}[H]
    \centering
\includegraphics[width=0.85\linewidth]{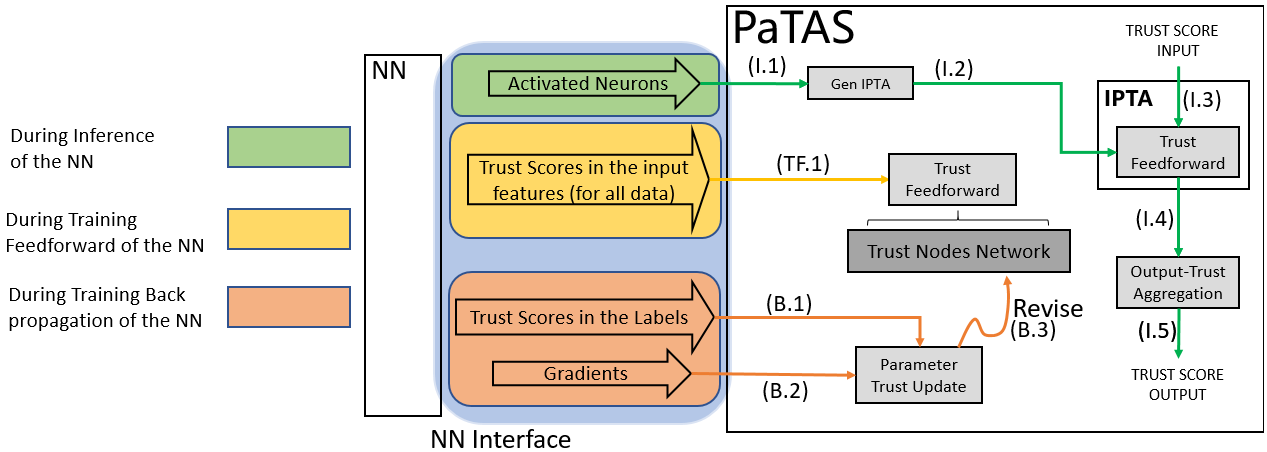}

\caption{
    Functional flow of the PaTAS framework integrated with a neural network, illustrating how trust is propagated and revised during feedforward, backpropagation, and inference. 
    \\
    \textbf{Feedforward (Training Phase):} 
    The NN interface extracts trust scores from input features for all data being processed. These trust scores (TF.1) are passed to the \textit{Trust Feedforward} module, which propagates them through the \textit{Trust Nodes Network} in parallel with the neural computations. During training, Trust Feedforward also stores the trust scores of all intermediate computations in the NN, to be reused in backpropagation. 
    \\
    \textbf{Backpropagation (Training Phase):} 
    Once the NN computes gradients, the trust scores of the labels (B.1) and the gradients are provided to the \textit{Parameter-Trust Update}. Gradients (B.2) indicate how dataset labels affect parameter updates, while label trust determines whether these updates should be considered reliable. The stored intermediate trust values are also incorporated in this process. Finally, the \textit{Parameter-Trust Update} module refines the trust values of the Trust Nodes Network parameters (B.3), aligning them with the evolving learning dynamics. 
    \\
    \textbf{Inference Phase:} 
    During operation, contextual information such as the set of activated neurons is retrieved. This information is passed to \textit{GenIPTA} (I.1), which constructs an \textit{Inference Path Trust Assessment (IPTA)} corresponding to the specific inference path (I.2). The IPTA takes input trust scores (I.3), propagates them along the path, and produces trust scores for each NN output. These can then be passed to the \textit{Output-Trust Aggregation} (I.4) which will combine them into a single consolidated trust score representing the reliability of the prediction (I.5).
    }
    \label{fig:ptasoperation}
\end{figure}

\section{Helper Functions for \cref{alg:trustupdate}}\label{app:algorithm}
\begin{algorithm}[H]
\caption{Helper functions used by the Parameter-Trust Update}
\begin{algorithmic}[1]
    \Function{NodeTrust}{$g^{(l)}_i, T_{y_\text{batch}}, \epsilon$}
        \State Count $r$: \#edges with $|g^{(l)}_{i,j}| < \epsilon$ 
        \State Count $s$: \#edges with $|g^{(l)}_{i,j}| \geq \epsilon$ 
        \State Map $(r,s)$ into a binomial opinion using Baseline-Prior Quantification
        \State \Return $T_{n_i \mid y_\text{batch}}$
    \EndFunction

    \Function{DeduceTrust}{$T_{n_i \mid y_\text{batch}}, T_{n_i \mid \overline{y_\text{batch}}}, T_{y_\text{batch}}$}
        \State $T_{n_i \mid \overline{y_\text{batch}}}  \longleftarrow  (0,0,1)$
        \State \Return $T_{n_i \parallel Y_\text{batch}} = T_{y_\text{batch}} \circledcirc (T_{n_i \mid y_\text{batch}}, T_{n_i \mid \overline{y_\text{batch}}})$
    \EndFunction

    \Function{Revise}{$T_{\theta_{i,j}^{(l)}}, T_{n_i \parallel Y_\text{batch}}$}
        \State \Return $T_{\theta_{i,j}^{(l)}} \circleddash T_{n_i \parallel Y_\text{batch}}$
    \EndFunction
    \Function{Update}{$T_{\theta^{(l)}_{i,j}},
    T_{x^{(l-1)}_j}, T_{y_\text{batch}}$}
    \State \Return $T_{\theta^{(l)}_{i,j}} \odot 
           (T_{x^{(l-1)}_j} \oslash T_{y_\text{batch}})$
    \EndFunction

\end{algorithmic}
\end{algorithm}

\section{Perturbation Functions Used in \cref{exp:1}}\label{app:perturb}
In practice, feature and label trust degradation may arise from poor-quality imaging, human annotation errors, or flaws in the preprocessing pipeline. 
In all experiments, we simulate degradations using controlled perturbation functions. 
For fully uncertain feature opinion, we introduce additive uniform noise 
\vspace{-1em}
\[
x' = x + \mu, \quad \mu \sim U(-\eta, \eta),
\]
where 
\(
\eta = 0.3 \times \max(\text{features}).
\)
Each feature is perturbed with probability $0.3$. 
After noise addition, if $x'$ lies outside the valid range 
$[\min(\text{features}), \max(\text{features})]$, it is clipped to the corresponding boundary:
\[
x' = 
\begin{cases}
\min(\text{features}), & x' < \min(\text{features}), \\
\max(\text{features}), & x' > \max(\text{features}), \\
x', & \text{otherwise}.
\end{cases}
\]
To model distrust, we generate corrupted inputs by sampling from a uniform distribution over the feature space:
\[
x' \sim U(\min(\text{features}), \max(\text{features})),
\]

Label degradation is modeled analogously: uncertainty is introduced via random label noise, while full distrust is represented by a complete replacement of the labels.

For labels, ``complete replacement'' draws a different class uniformly; with two classes this is deterministic flipping, which is why fully distrusted labels yield near-zero rather than chance-level test accuracy in \cref{tab:cancer_results}.

\section{Conformity-Based Input-Trust Construction (\cref{exp:4})}\label{app:conformity}
This appendix specifies the input-trust construction used in \cref{exp:4} precisely; all parameters are fixed once and shared by every dataset.

Let the clean training features be $x^{(1)},\dots,x^{(n)} \in \mathbb{R}^m$. For each feature $j$ we compute the marginal mean $\mu_j$ and standard deviation $\sigma_j$, and a robust span $S$ of the whole feature matrix as the difference between its $99.5$th and $0.5$th percentiles. The \emph{effective deviation} is floored at a fraction of the span,
\[
\sigma^{\mathrm{eff}}_j = \max(\sigma_j,\, 0.02\, S),
\]
so that near-constant features cannot produce unbounded standardized deviations from minimal perturbations. For an input $x$, the standardized deviation is $z_j = |x_j - \mu_j| / \sigma^{\mathrm{eff}}_j$ and the conformity is
\[
g_j = \exp\!\big(-\tfrac{1}{2}\max(z_j - 1,\, 0)^2\big) \in (0,1],
\]
so deviations within one effective standard deviation are fully conforming and only the excess erodes conformity. The \emph{reliability weight} of feature $j$ is the capped inverse-variance weight
\[
w_j = \min\!\big((\sigma_{\mathrm{ref}} / \sigma^{\mathrm{eff}}_j)^2,\, 64\big),
\qquad \sigma_{\mathrm{ref}} = \operatorname{median}_j \sigma^{\mathrm{eff}}_j,
\]
expressing that a feature with a tight training distribution is a reliable witness whose violation is unambiguous, while the cap bounds the influence any single feature can exert. Each feature contributes evidence $r_j = N w_j g_j$ and $s_j = N w_j (1-g_j)$ with base evidence mass $N = 50$, mapped to a binomial opinion by \cref{eq:q1} with $W = 2$; this is the same quantification used for parameter trust. The per-sample input trust $P_{\mathrm{input}}$ is the projected probability (base rate $0.5$) of the pooled opinion obtained by applying \cref{eq:q1} to the summed evidence $(\sum_j r_j, \sum_j s_j)$, which corresponds to cumulative fusion of independent evidence sources.

The columns of \cref{tab:diagnostics} are, in this notation: the median of $\sigma^{\mathrm{eff}}_j$, the maximum of $w_j$, the share of total weight mass held by the top decile of features ranked by $w_j$, and the fraction of features with $w_j \geq 4$. The last two are descriptive statistics of weight concentration used for reporting only (the threshold $w_j \geq 4$ corresponds to $\sigma^{\mathrm{eff}}_j$ at most half of $\sigma_{\mathrm{ref}}$); they are not parameters of the construction.

\begin{table}[ht]
\centering\footnotesize
\setlength{\tabcolsep}{4.5pt}
\begin{tabular}{lccccc}
\toprule
Dataset & Med.\ $\sigma_{\mathrm{eff}}$ & $w_{\max}$ & Top-10\% & $w{\geq}4$ & Corr.\ det. \\
\midrule
MNIST & 0.47 & 52.3 & 0.32 & 0.40 & 0.972 \\
Fashion-MNIST & 0.90 & 64.0 & 0.76 & 0.18 & 0.835 \\
GTSRB & 0.98 & 1.8 & 0.14 & 0.00 & 0.545 \\
CIFAR-10 (gray) & 0.23 & 1.1 & 0.11 & 0.00 & -- \\
\bottomrule
\end{tabular}
\caption{A-priori applicability diagnostics: conformity-model statistics on each dataset's clean training features \emph{before any model is trained}, against the corruption detection later observed. Registered domains concentrate evidence in reliable witnesses; unregistered imagery has none, so input-level conformity is predicted, and for GTSRB confirmed, to be uninformative. The CIFAR-10 row is computed on the raw grayscale features; the convolutional battery (Appendix~\ref{app:conv}) standardizes per channel, which restores partial conformity information (corruption detection $0.72$).}
\label{tab:diagnostics}
\end{table}

\vspace{-0.5em}
\section{Operator and Threshold Sensitivity}\label{app:ablations}
Both ablations use the MNIST 784-128-10 trust/trust scenario of \cref{exp:2} (20 epochs, 1000-sample evaluation subset, test-time feature noise $p = 0.3$ for the noisy columns).

\Cref{tab:abl-fusion} varies the trust-revision fusion operator. The converged trust mass and every downstream metric agree to within $0.002$ across averaging, cumulative, and weighted fusion, so the findings of this paper do not depend on the variant; averaging is used throughout because Theorem~\ref{thm:arithm} establishes non-degenerate long-run behavior under repeated revision, whereas cumulative fusion accumulates evidence monotonically toward dogmatic (zero-uncertainty) opinions.

\Cref{tab:abl-eps} sweeps the gradient-evidence threshold over three orders of magnitude. The trusted-input trust mass and the selective-prediction metrics are constant on the plateau $\epsilon \in [0.05, 1]$ and degrade only at degenerate-small values, where nearly every gradient is read as strong evidence; the equality of the trusted-input trust and distrusted-input distrust columns is the belief/disbelief symmetry of Theorem~\ref{thm:infvac} observed empirically.

\begin{table}[H]
\centering\footnotesize
\begin{tabular}{lccccc}
\toprule
Fusion & Trust $t_{\mathrm{tr}}$ & AUROC$_{\mathrm{cl}}$ & AUROC$_{\mathrm{no}}$ & ECE$_{\mathrm{cl}}$ & ECE$_{\mathrm{no}}$ \\
\midrule
Average (used) & 0.988 & 0.957 & 0.973 & 0.013 & 0.020 \\
Cumulative & 0.990 & 0.957 & 0.973 & 0.012 & 0.020 \\
Weighted & 0.989 & 0.957 & 0.973 & 0.013 & 0.020 \\
\bottomrule
\end{tabular}
\caption{Trust-revision fusion ablation: converged trusted-input trust mass and selective-prediction metrics on clean (cl) and feature-noised (no) test data.}
\label{tab:abl-fusion}
\end{table}

\begin{table}[H]
\centering\footnotesize
\begin{tabular}{lccccc}
\toprule
$\epsilon$ & $t_{\mathrm{tr}}$ & $d_{\mathrm{dis}}$ & AUROC$_{\mathrm{cl}}$ & AUROC$_{\mathrm{no}}$ & ECE$_{\mathrm{cl}}$ \\
\midrule
0.001 & 0.573 & 0.573 & 0.826 & 0.900 & 0.195 \\
0.01 & 0.976 & 0.976 & 0.940 & 0.964 & 0.018 \\
0.05$^{*}$ & 0.989 & 0.989 & 0.957 & 0.973 & 0.013 \\
0.1 & 0.989 & 0.989 & 0.957 & 0.973 & 0.013 \\
0.2 & 0.989 & 0.989 & 0.957 & 0.973 & 0.013 \\
0.5 & 0.989 & 0.989 & 0.957 & 0.973 & 0.013 \\
1.0 & 0.989 & 0.989 & 0.957 & 0.973 & 0.013 \\
\bottomrule
\end{tabular}
\caption{$\epsilon$ sweep ($^{*}$ = value used in Experiments 2--4): trusted-input trust $t_{\mathrm{tr}}$, distrusted-input distrust $d_{\mathrm{dis}}$ (equal by Theorem~\ref{thm:infvac}), and selective-prediction metrics. All quantities are constant on $\epsilon \in [0.05, 1]$.}
\label{tab:abl-eps}
\end{table}

\begin{table}[H]
\centering\footnotesize
\begin{tabular}{lccc}
\toprule
Hidden width & $\epsilon=0.02$ & $\epsilon=0.05$ & $\epsilon=0.1$ \\
\midrule
16 & 0.253 & 0.306 & 0.316 \\
32 & 0.296 & 0.325 & 0.328 \\
64 & 0.317 & 0.333 & 0.334 \\
128 & 0.325 & 0.337 & 0.337 \\
\bottomrule
\end{tabular}
\caption{Width control for \cref{exp:2} (vacuous assessments): converged trust mass across architectures at three gradient-evidence thresholds. The width ordering is identical at every $\epsilon$, ruling out the fixed-threshold confound; the magnitudes are $\epsilon$-dependent, consistent with \cref{tab:abl-eps}.}
\label{tab:abl-width}
\end{table}

\section{Poisoned-MNIST Patch-Size Sweep (\cref{exp:3})}\label{app:pois}
\Cref{tab:mnistpois} reports the per-class trust mass and accuracy across patch sizes for the poisoned-MNIST experiment.

\begin{table}[ht]
\centering\footnotesize
\setlength{\tabcolsep}{4pt}
\begin{tabular}{ccccccc}
\hline
Patch & \multicolumn{2}{c}{Trust} & \multicolumn{2}{c}{Clean acc.\ (\%)} & \multicolumn{2}{c}{Patched acc.\ (\%)} \\
size & 3 & 6 & 3 & 6 & 3 & 6 \\
\hline
\((1 \times 1)\) & 0.897 & 0.577 & 97.92 & 97.29 & 78.22 & 0.84 \\
\((4 \times 4)\) & 0.895 & 0.576 & 97.82 & 97.39 & 39.60 & 1.04 \\
\((10 \times 10)\) & 0.881 & 0.570 & 97.92 & 97.39 & 22.08 & 0.84 \\
\((27 \times 27)\) & 0.697 & 0.490 & 97.92 & 97.29 & 0 & 0 \\
\hline
\end{tabular}
\caption{Poisoned MNIST; overall train/test accuracy (96.4--99.6\%/97.7--97.8\% across patch sizes) is omitted.}
\label{tab:mnistpois}
\end{table}

\section{Convolutional Battery (\cref{exp:4})}\label{app:conv}
The convolutional networks comprise three conv layers plus a residual block (MNIST and Fashion-MNIST) and seven residual layers (CIFAR-10, per-channel standardized), all trained with momentum, weight decay, augmentation, and a stepped learning-rate schedule. \Cref{tab:conv} reports the detection tasks on these architectures. On the MNIST CNN, an architecture defensible for the domain ($99.3\%$), the composed score matches the softmax filter on clean correct-vs-incorrect ranking ($0.986$ for both) while dominating detection (corruption $0.988$ vs.\ $0.662$; OOD $0.991$ vs.\ $0.967$), so on this architecture the composition pays no in-distribution price at all. Evidential Deep Learning, weak on the MLPs, becomes the best in-distribution ranker on the CNN (clean AUROC $0.996$) yet stays blind to violations (corruption $0.527$): the complementarity of confidence-type and trust-type signals is architecture-independent. On CIFAR-10 the corruption genuinely reduces accuracy, and there the composed score is the strongest detector outright ($0.732$, above the conformity factor alone at $0.723$), with the mixed-batch correctness ranking a near-tie with softmax ($0.798$ vs.\ $0.807$): when violations do cause errors, path trust and confidence pull in the same direction. The model-side audit also transfers: the label-noise-trained CIFAR model carries path trust $0.521$ against $0.697$ for the clean-trained one, a $25\%$ relative drop read from the trained trust state through per-sample conv IPTA. Two honest notes: in the subset-direction OOD task (Fashion CNN, MNIST as OOD) the composed score becomes \emph{anti-correlated} ($0.20$), because sparse digit inputs activate few channels and activity weighting reads short paths as more trusted, so it must not be used as an OOD filter in this direction (see the operational guidance and SL-fusion remedy in \cref{exp:4}); and the feature-space detector baselines (Mahalanobis, kNN, energy) are implemented for the fully connected models only, so the convolutional comparison covers the output-only baselines and our own variants.

\begin{table}[ht]
\centering\footnotesize
\setlength{\tabcolsep}{4pt}
\begin{tabular}{lccccc}
\toprule
 & \multicolumn{2}{c}{MNIST CNN} & \multicolumn{2}{c}{Fashion CNN} & CIFAR-10 \\
Method & Corr.\ det. & OOD & Corr.\ det. & OOD F$\to$M & Corr.\ det. \\
\midrule
PaTAS (composed) & $0.988$ & $0.991$ & $0.859$ & $0.201$ & $\mathbf{0.732}$ \\
Conformity alone & $\mathbf{1.000}$ & $\mathbf{1.000}$ & $\mathbf{0.994}$ & $0.481$ & $0.723$ \\
Max softmax & $0.662$ & $0.967$ & $0.711$ & $0.281$ & $0.700$ \\
MC Dropout & $0.605$ & $0.978$ & $0.616$ & $\mathbf{0.712}$ & $0.622$ \\
EDL & $0.527$ & $0.820$ & $0.522$ & $0.608$ & $0.525$ \\
\bottomrule
\end{tabular}
\caption{Convolutional battery (mean over three evaluation seeds): corruption detection on the mixed batch and OOD detection where a counterpart is defined (MNIST CNN vs.\ Fashion-MNIST; Fashion CNN vs.\ MNIST, the subset direction). Base accuracies $99.3\%$, $88.7\%$, $84.5\%$; seed dispersion is at most $\pm 0.004$ for every entry.}
\label{tab:conv}
\end{table}

\section{Detailed results}\label{app:expres}

\begin{figure}[H]
\centering
\begin{subfigure}{0.72\linewidth}
    \centering
    \includegraphics[width=\linewidth]{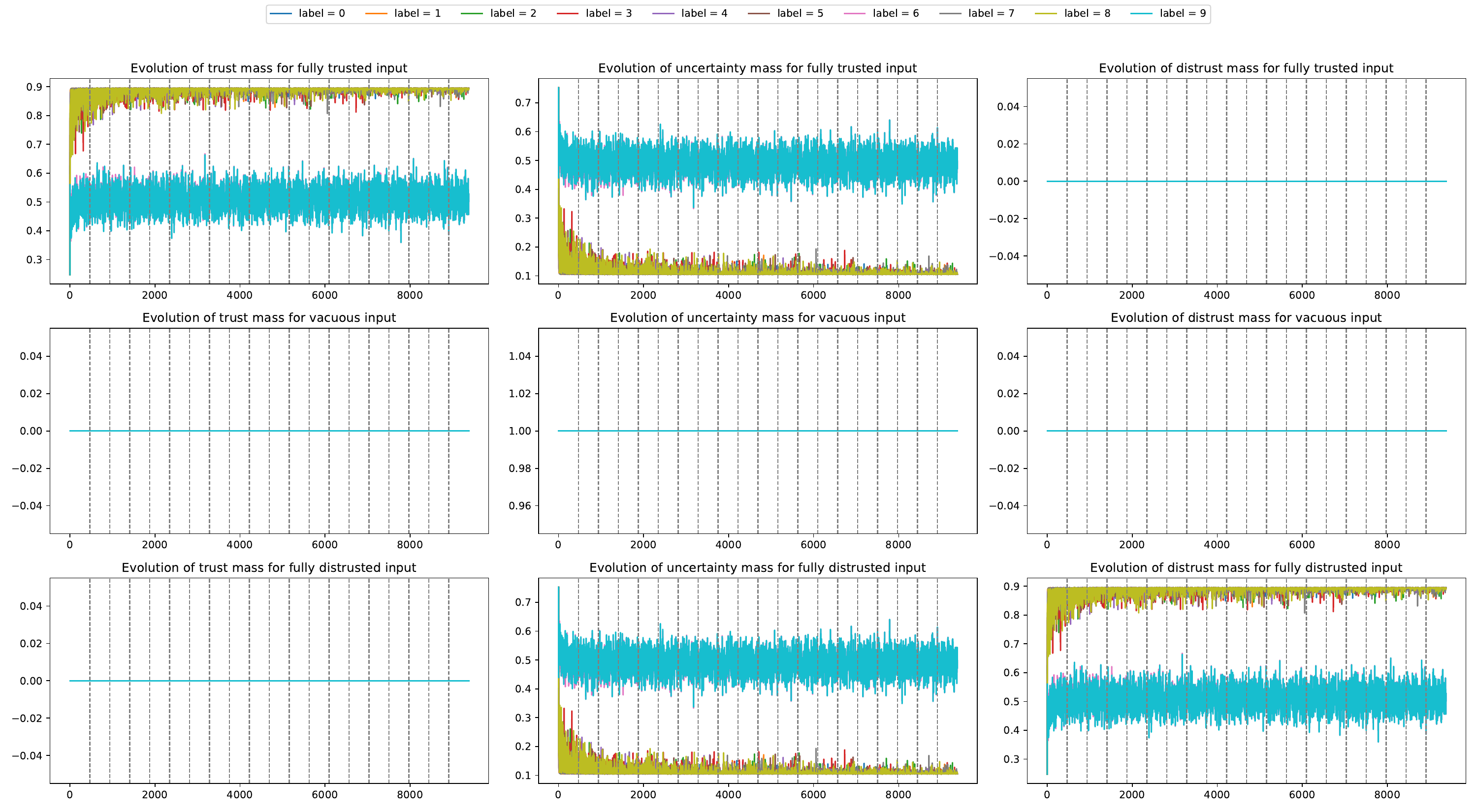}
    \caption{Trust mass}
    \label{fig:4x4}
\end{subfigure}
\begin{subfigure}{0.72\linewidth}
    \centering
    \includegraphics[width=\linewidth]{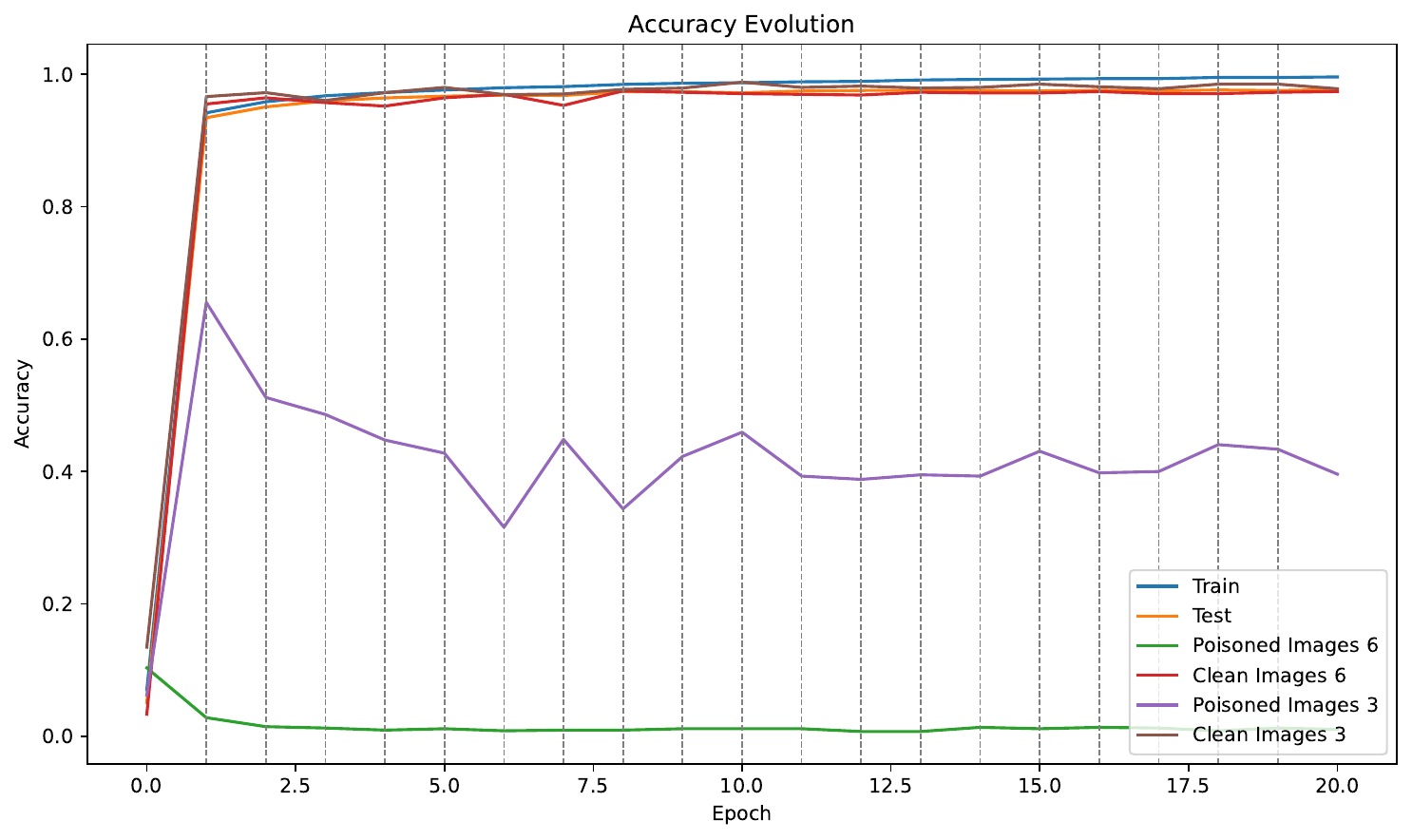}
    \caption{Accuracy}
    \label{fig:4x4acc}
\end{subfigure}
\caption{Evolution of trust mass and accuracy for a fully trusted input, MNIST poisoned experiment, patch size \(4\times4\).}
\label{fig:4x4both}
\end{figure}

All the results are depicted in Figures in \cref{sec:rescancer0.01,sec:rescancer0.1,sec:resmnistaccapp,sec:resmnist,sec:resmnistpois,sec:resrand}.
\begin{itemize}
\item \emph{Fully Trusted Input}: where the trust in the data is considered trusted (row 1 in each figure).
\item \emph{Fully Uncertain Input}: where the trust in the data is considered uncertain (row 2 in each figure).
\item \emph{Fully Distrusted Input}: where the data is assumed to be unreliable (row 3 in each figure).
\end{itemize}

\subsection{Plots for Cancer Model with \(\epsilon = 0.01\) (\cref{exp:1})}\label{sec:rescancer0.01}
\begin{figure}[H]
  \includegraphics[width=\textwidth]{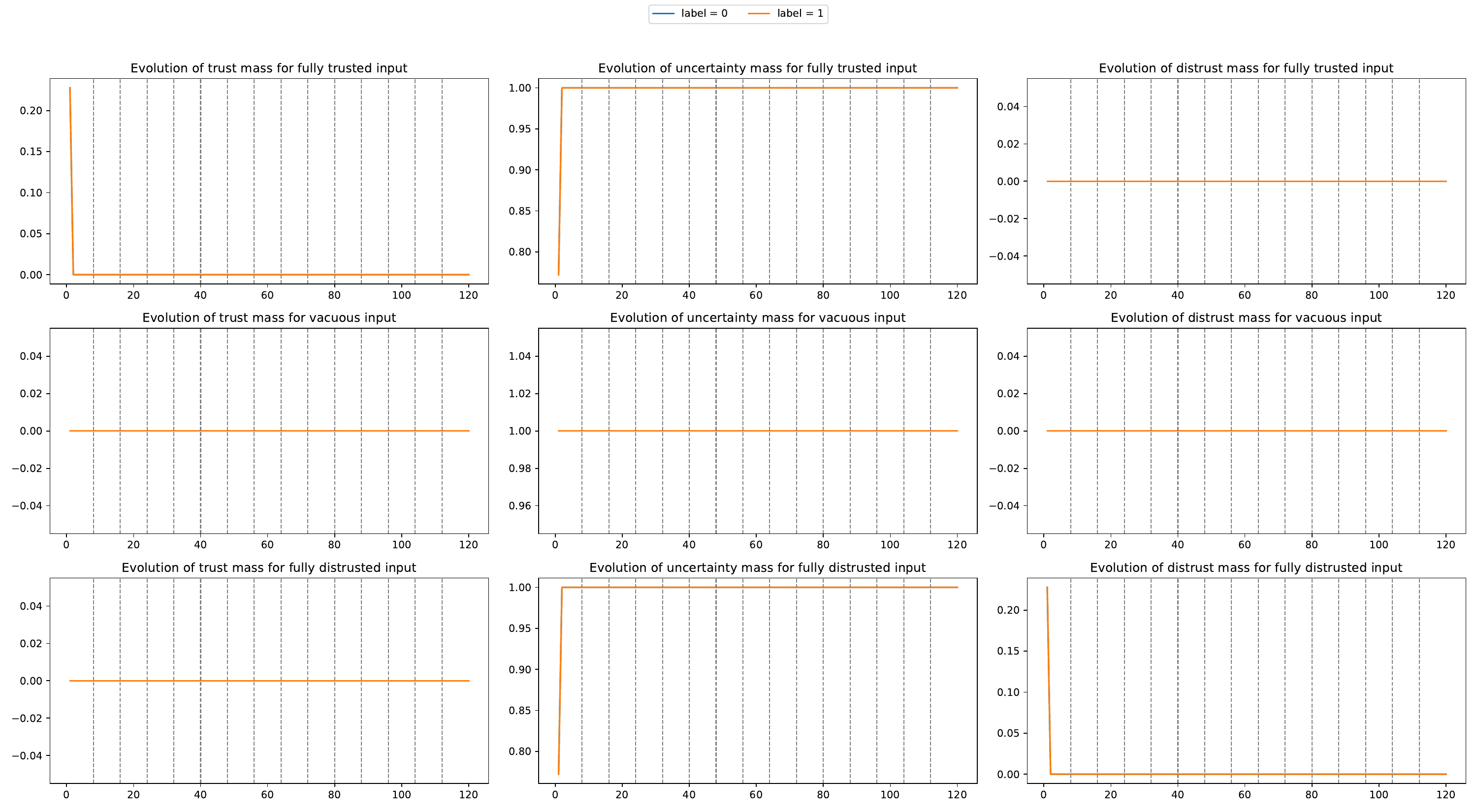}
  \caption{Features distrusted and labels distrusted}
\end{figure}
\begin{figure}[H]
  \includegraphics[width=\textwidth]{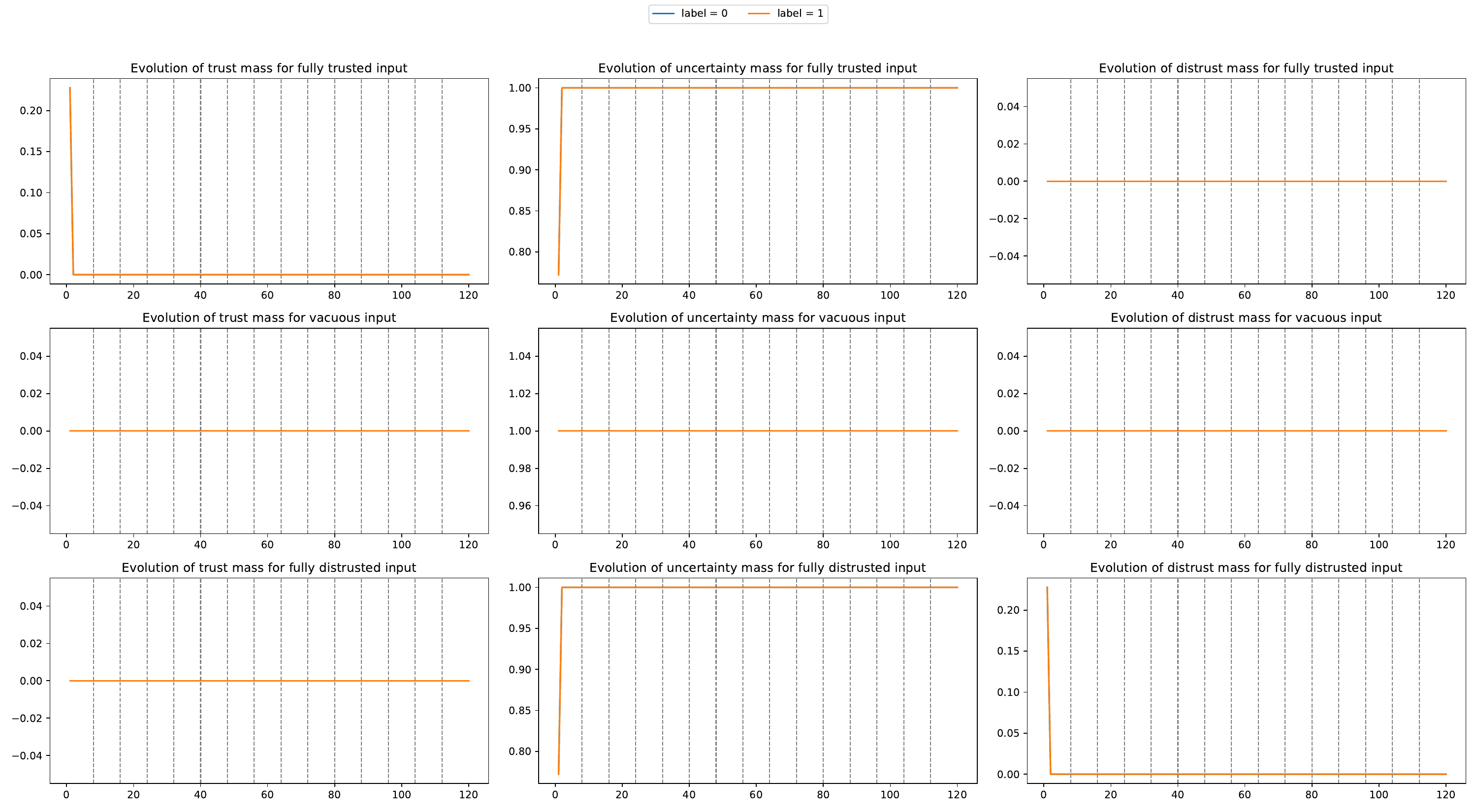}
  \caption{Features distrusted and labels vacuous}
\end{figure}
\begin{figure}[H]
  \includegraphics[width=\textwidth]{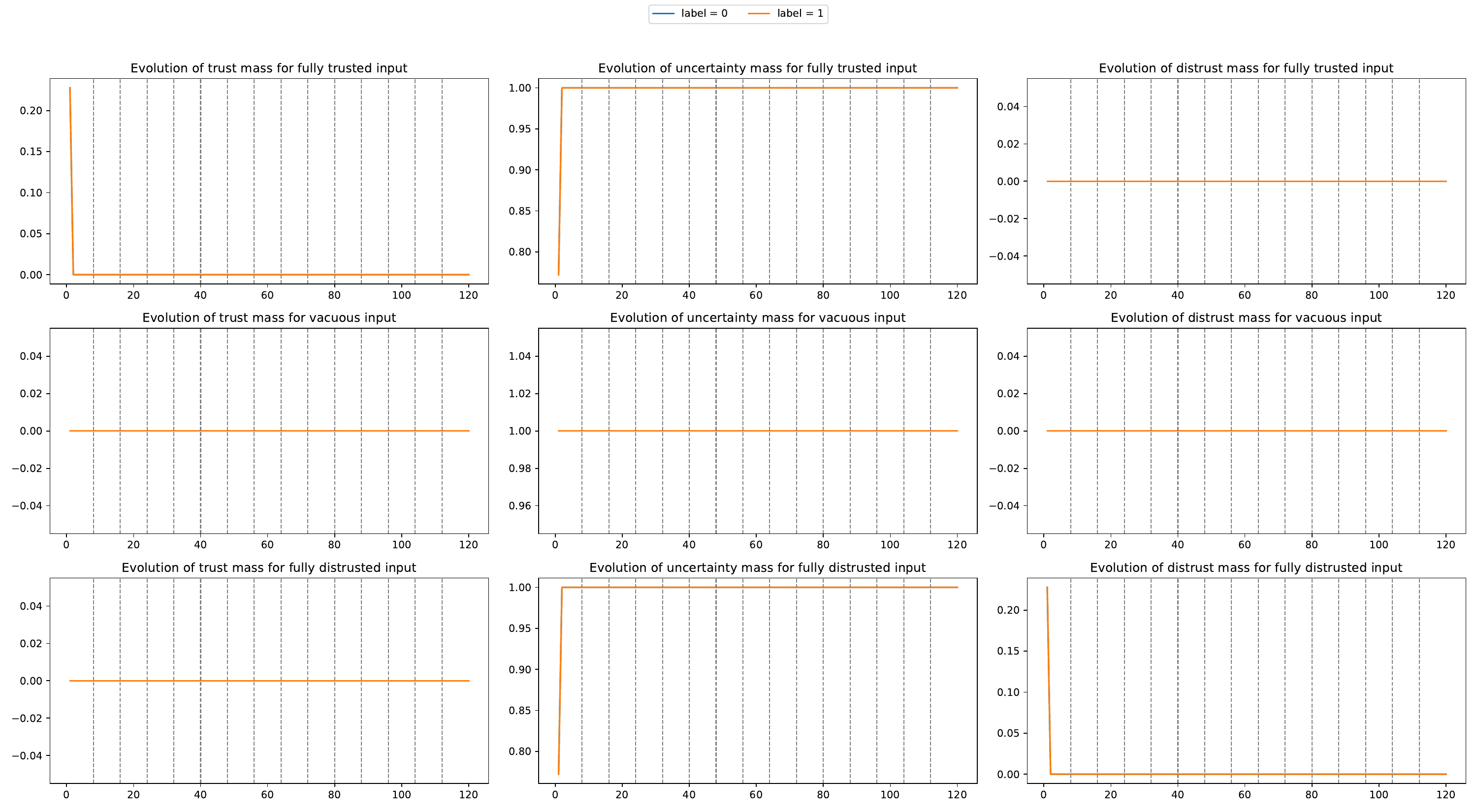}
  \caption{Features distrusted and labels trusted}
\end{figure}
\begin{figure}[H]
  \includegraphics[width=\textwidth]{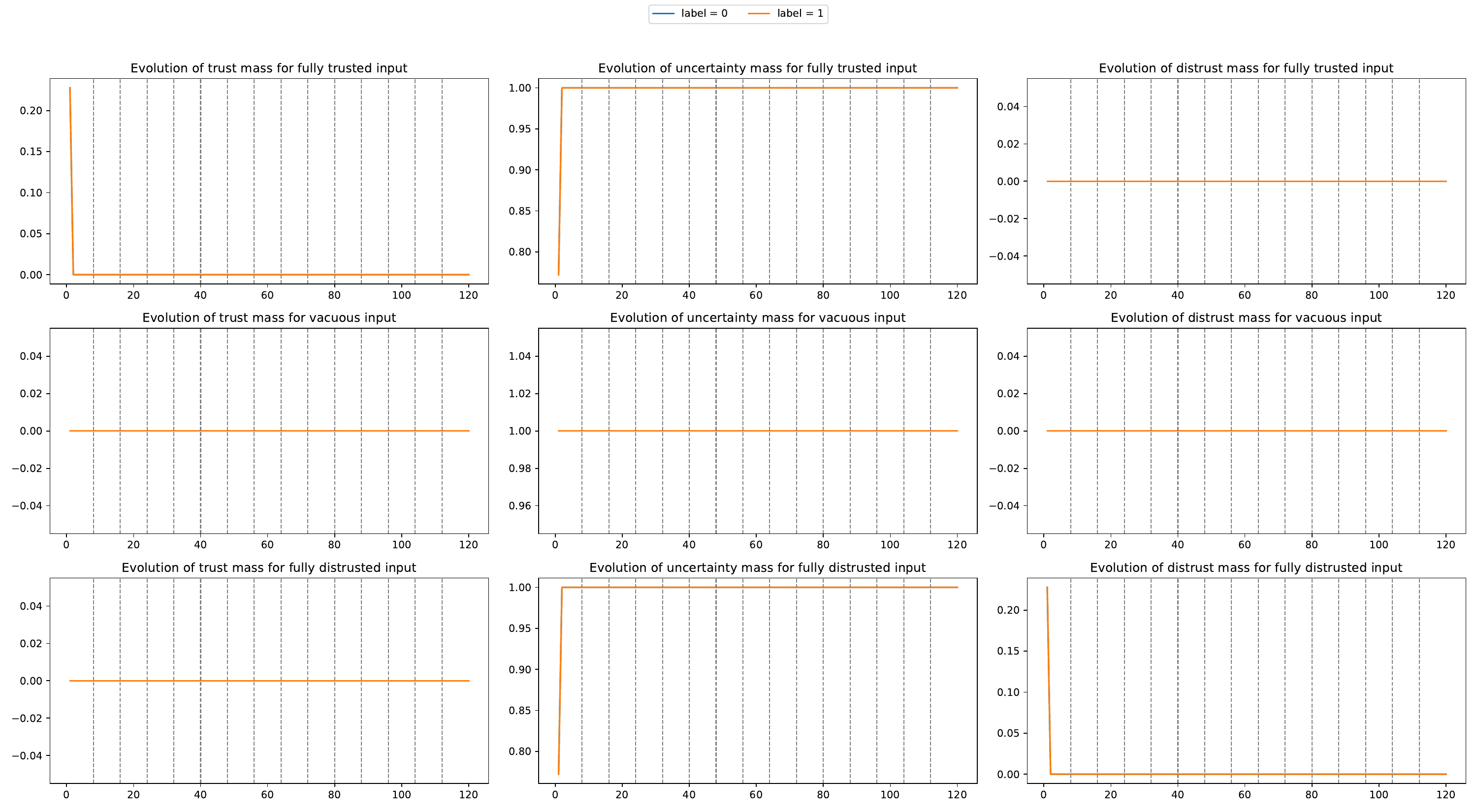}
  \caption{Features vacuous and labels distrusted}
\end{figure}
\begin{figure}[H]
  \includegraphics[width=\textwidth]{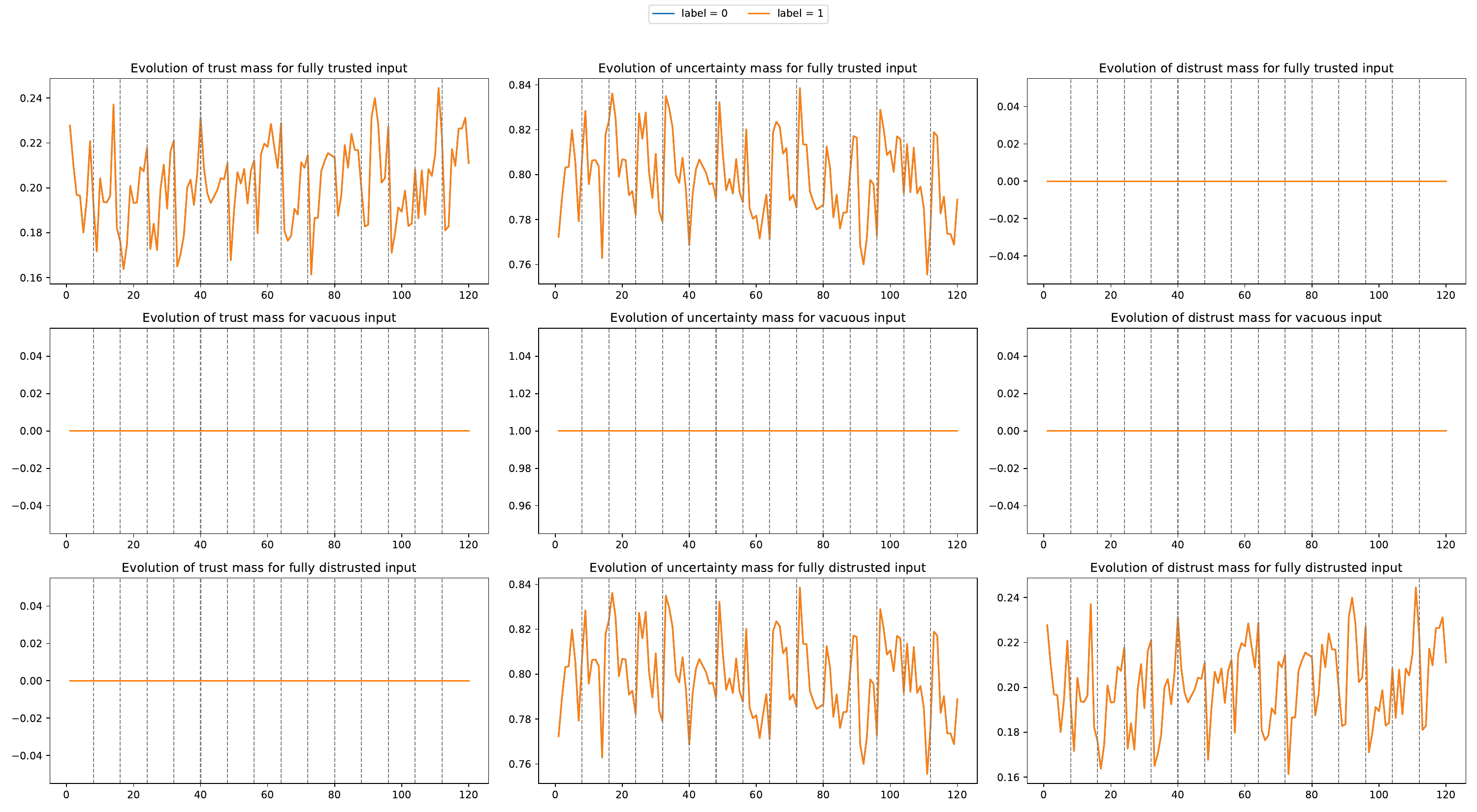}
  \caption{Features vacuous and labels vacuous}
\end{figure}
\begin{figure}[H]
  \includegraphics[width=\textwidth]{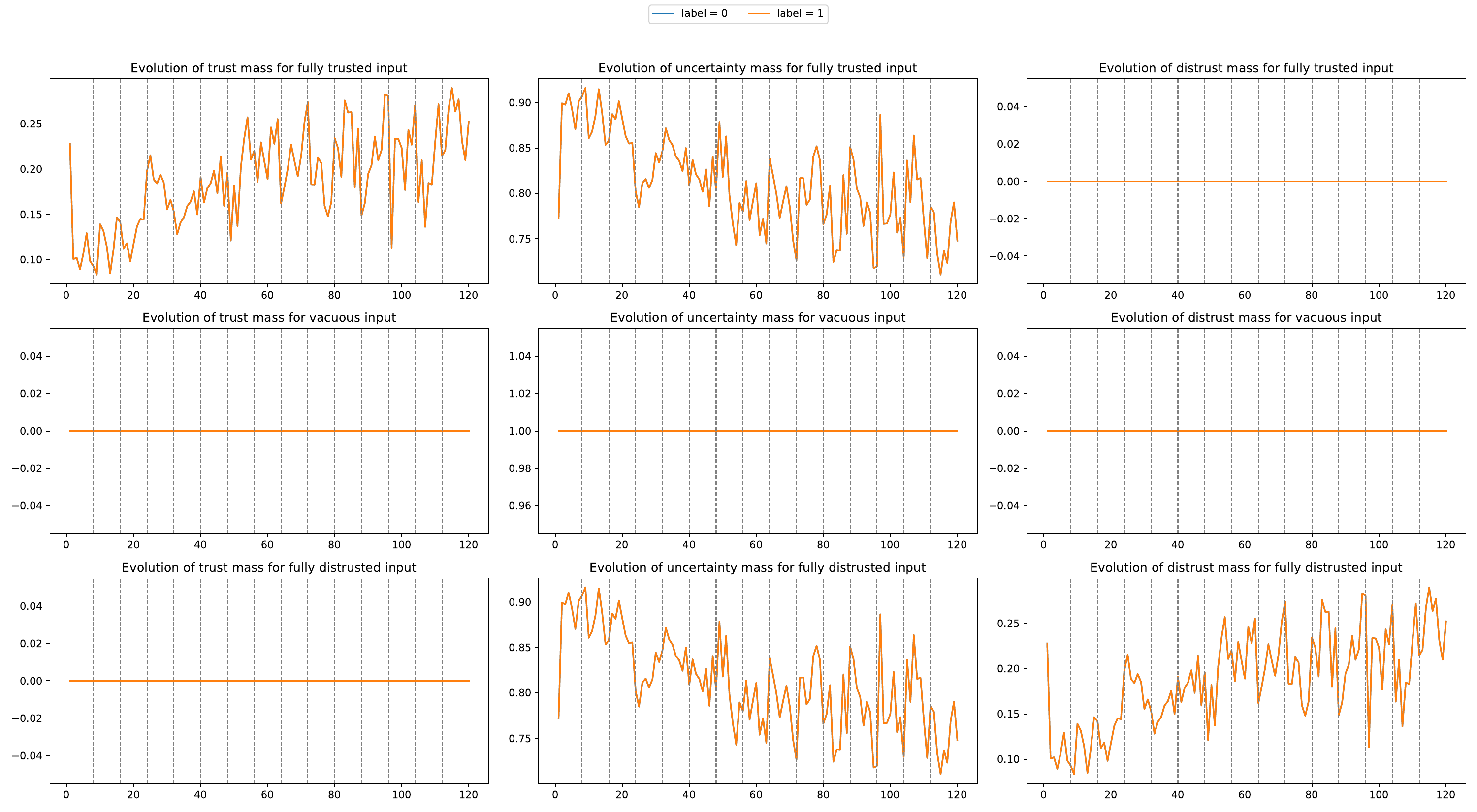}
  \caption{Features vacuous and labels trusted}
\end{figure}
\begin{figure}[H]
  \includegraphics[width=\textwidth]{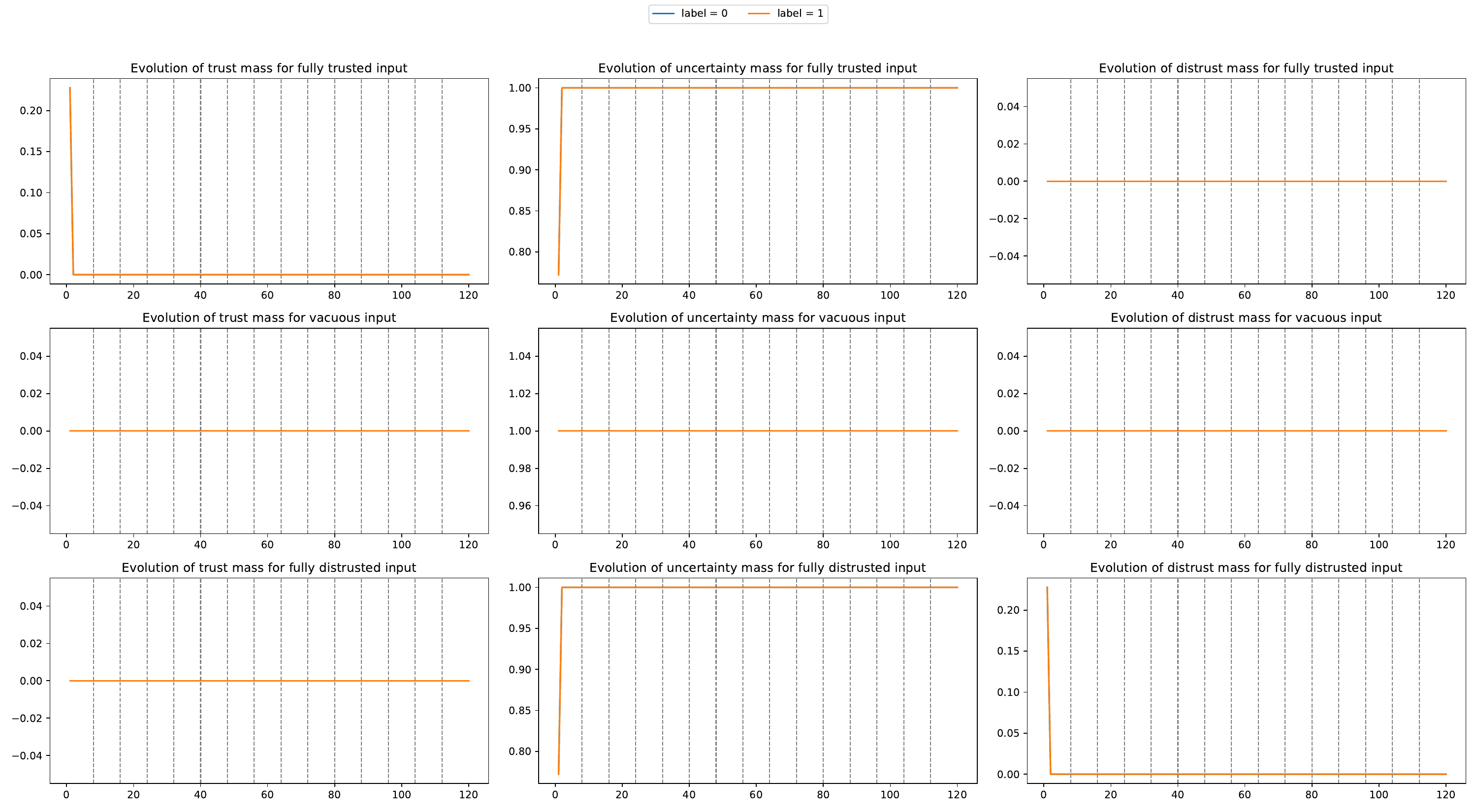}
  \caption{Features trusted and labels distrusted}
\end{figure}
\begin{figure}[H]
  \includegraphics[width=\textwidth]{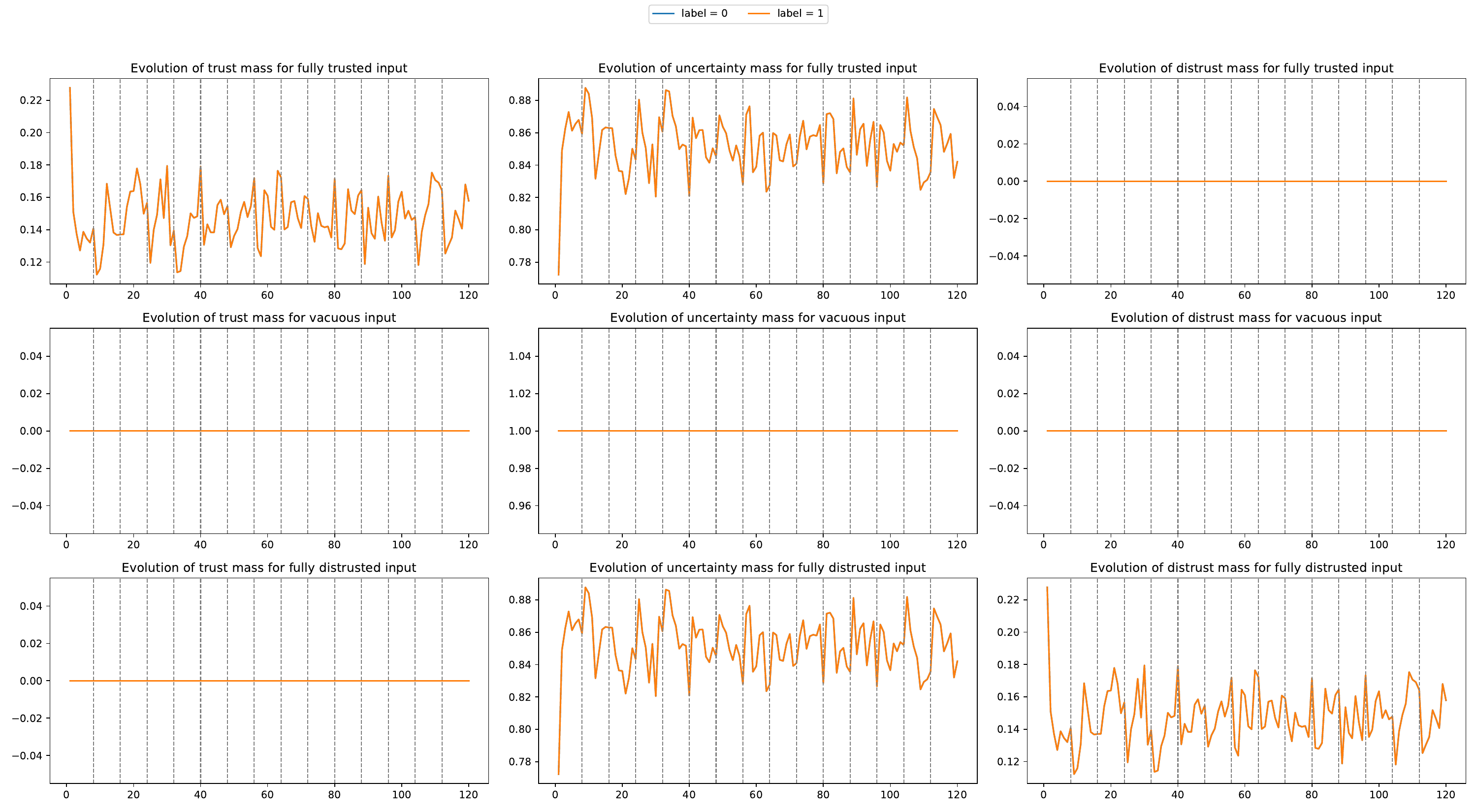}
  \caption{Features trusted and labels vacuous}
\end{figure}
\begin{figure}[H]
  \includegraphics[width=\textwidth]{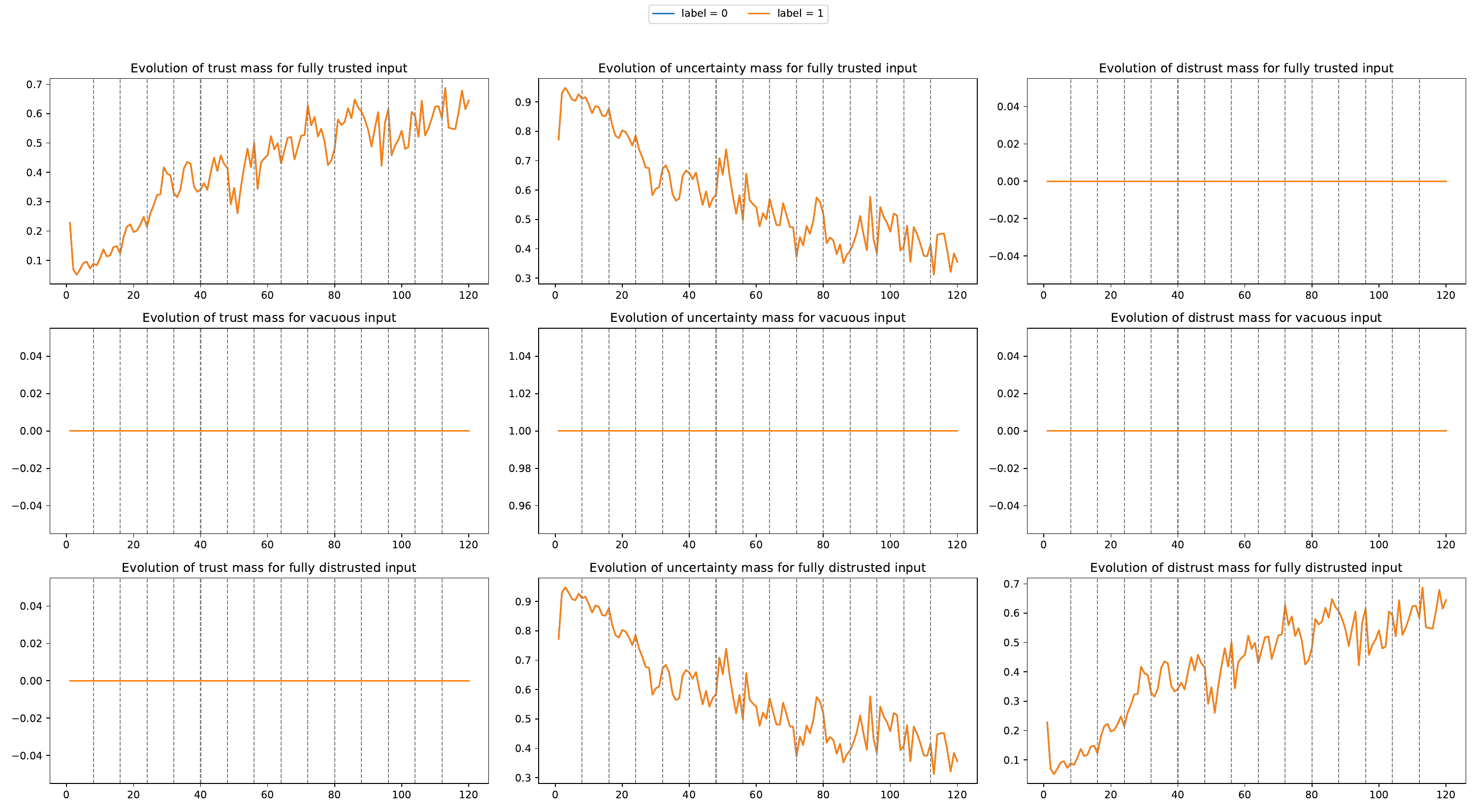}
  \caption{Features trusted and labels trusted}
\end{figure}

\subsection{Cancer Model with \( \epsilon = 0.1 \)(\cref{exp:1})}\label{sec:rescancer0.1}
\begin{figure}[H]
  \includegraphics[width=\textwidth]{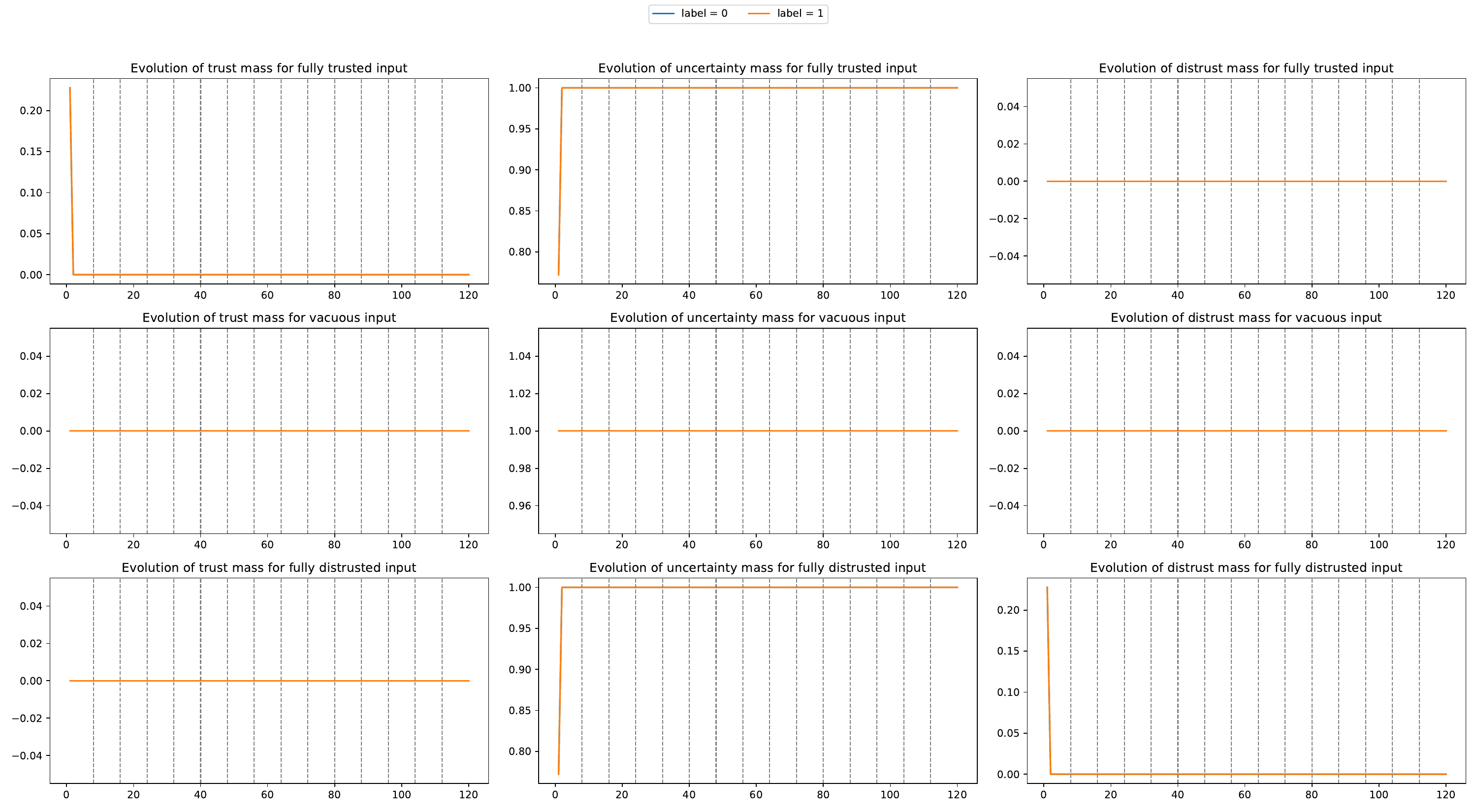}
  \caption{Features distrusted and labels distrusted}
\end{figure}
\begin{figure}[H]
  \includegraphics[width=\textwidth]{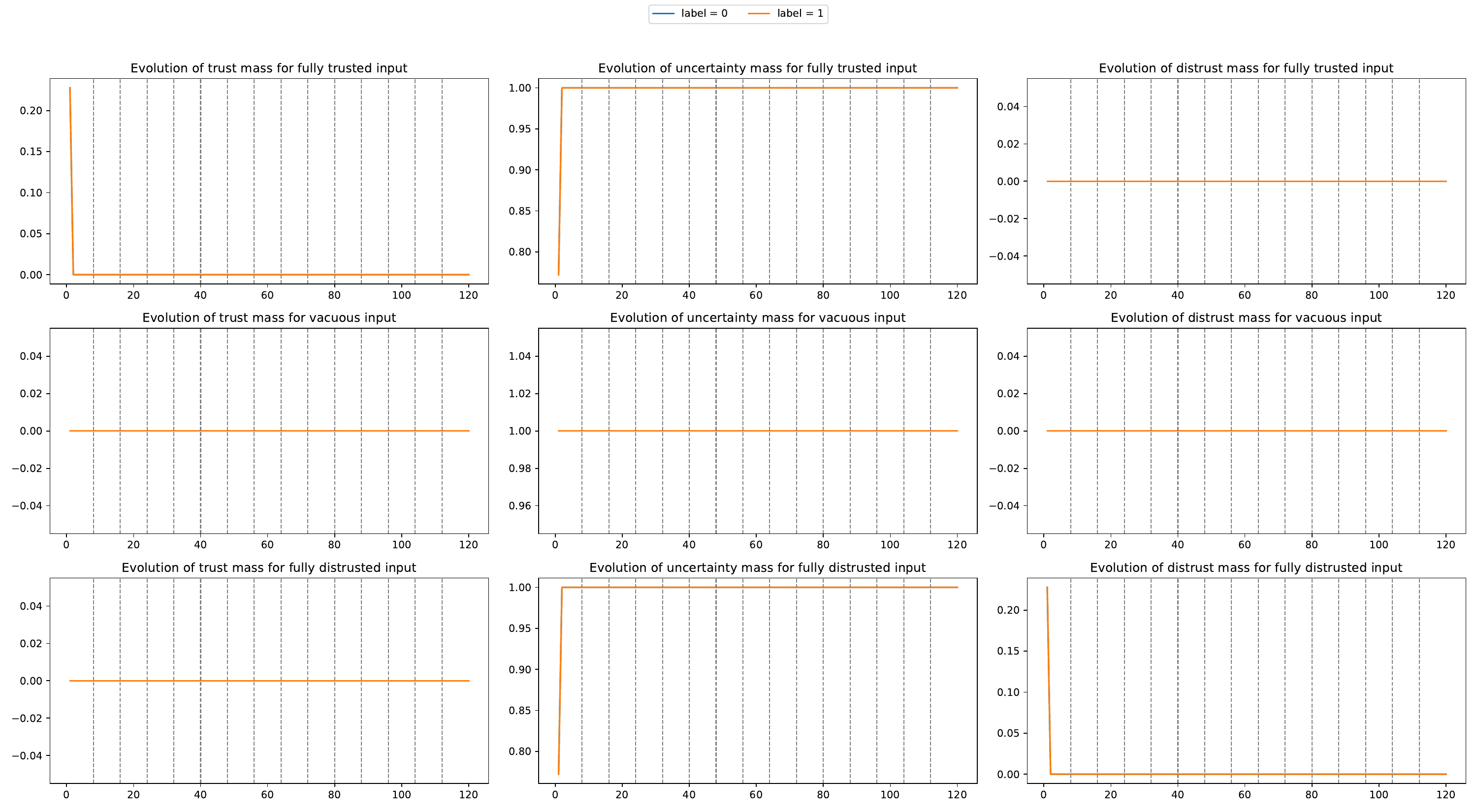}
  \caption{Features distrusted and labels vacuous}
\end{figure}
\begin{figure}[H]
  \includegraphics[width=\textwidth]{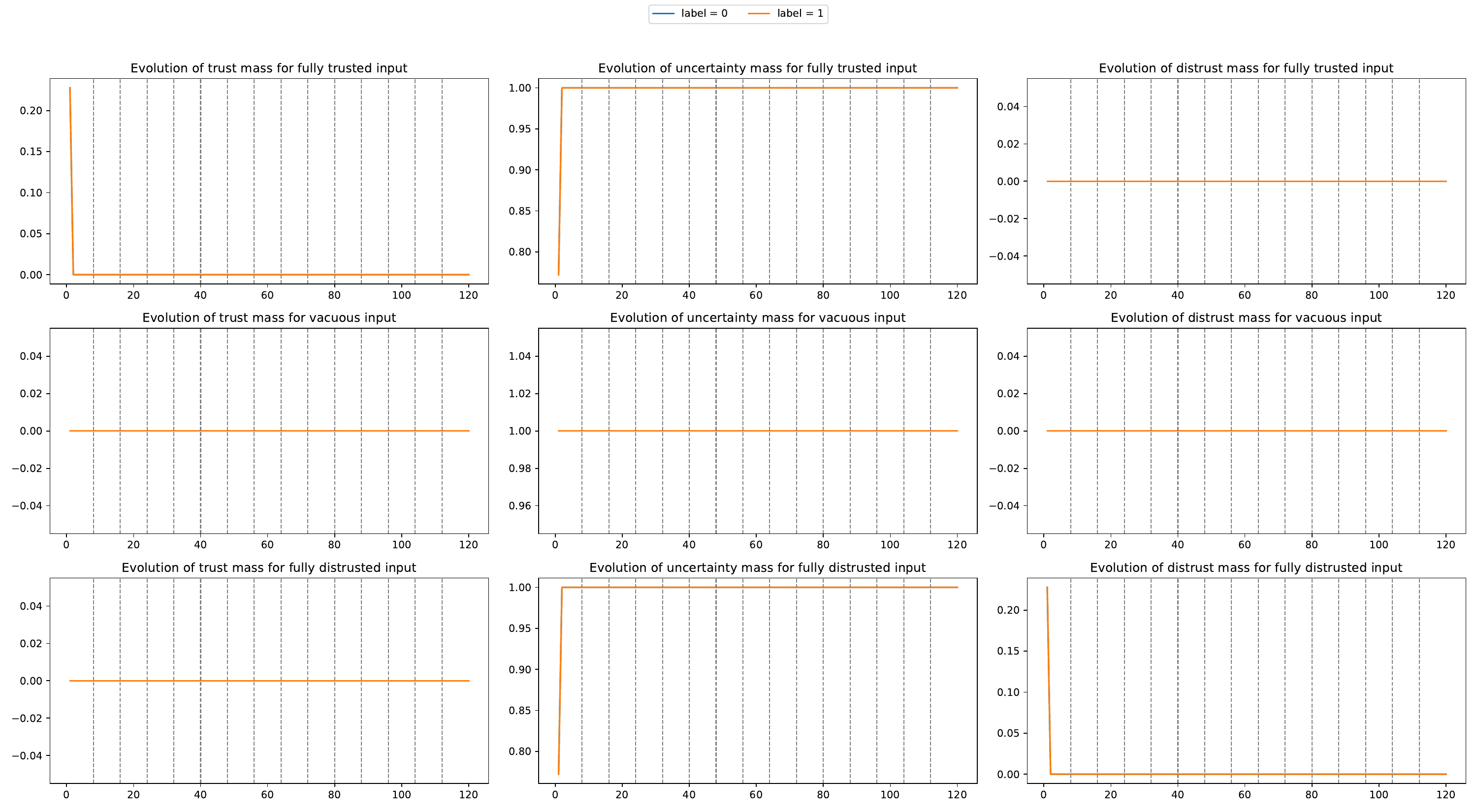}
  \caption{Features distrusted and labels trusted}
\end{figure}
\begin{figure}[H]
  \includegraphics[width=\textwidth]{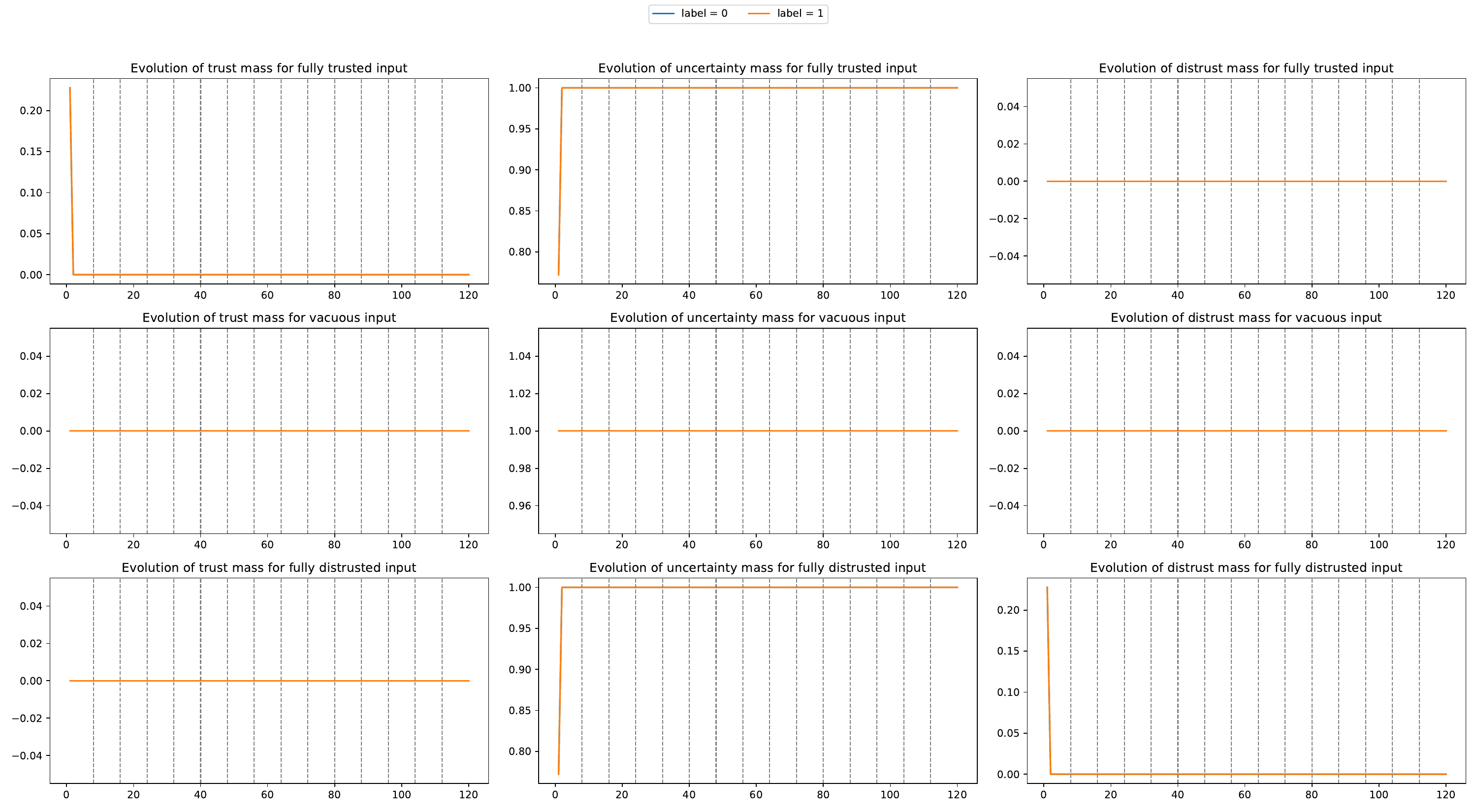}
  \caption{Features vacuous and labels distrusted}
\end{figure}
\begin{figure}[H]
  \includegraphics[width=\textwidth]{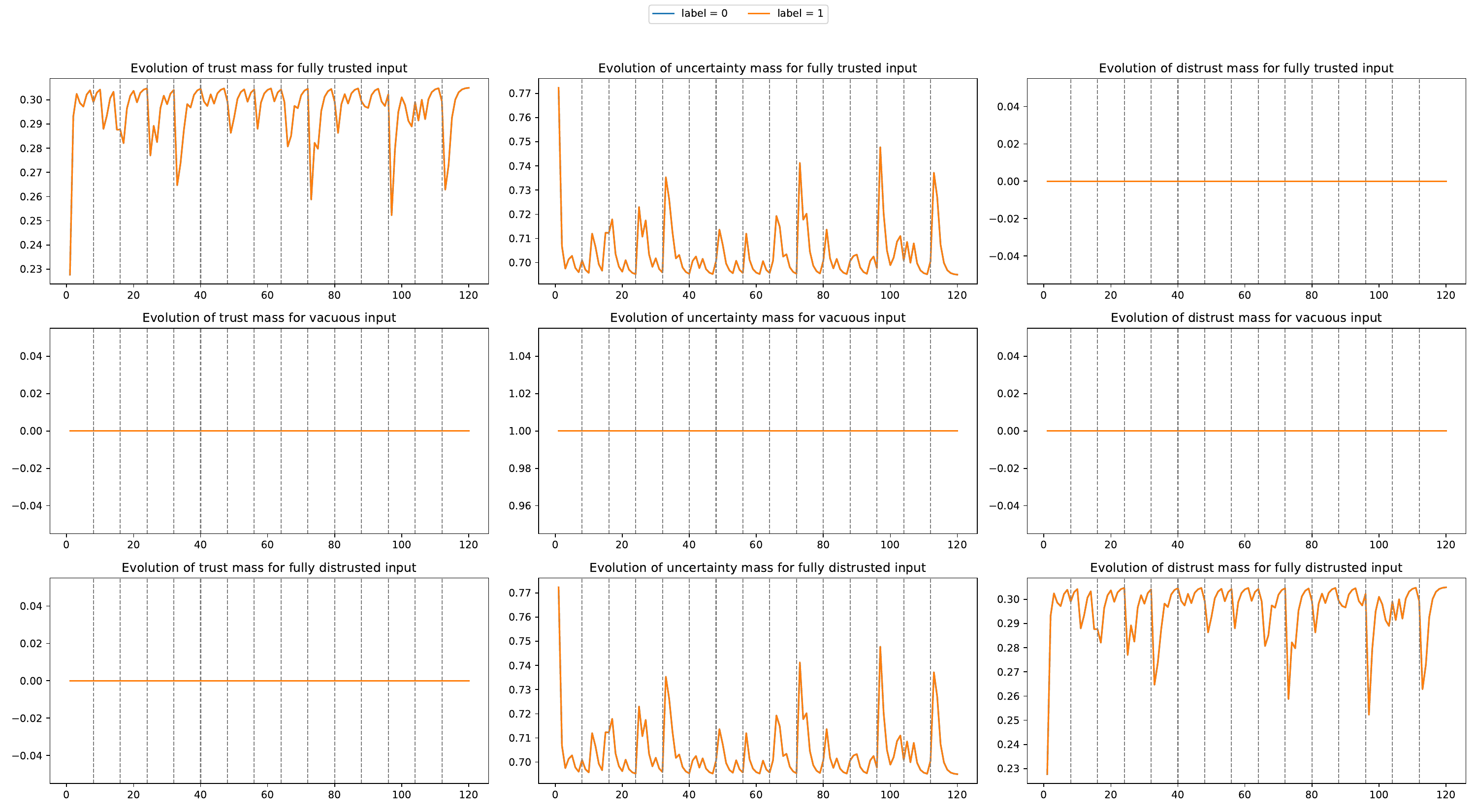}
  \caption{Features vacuous and labels vacuous}
\end{figure}
\begin{figure}[H]
  \includegraphics[width=\textwidth]{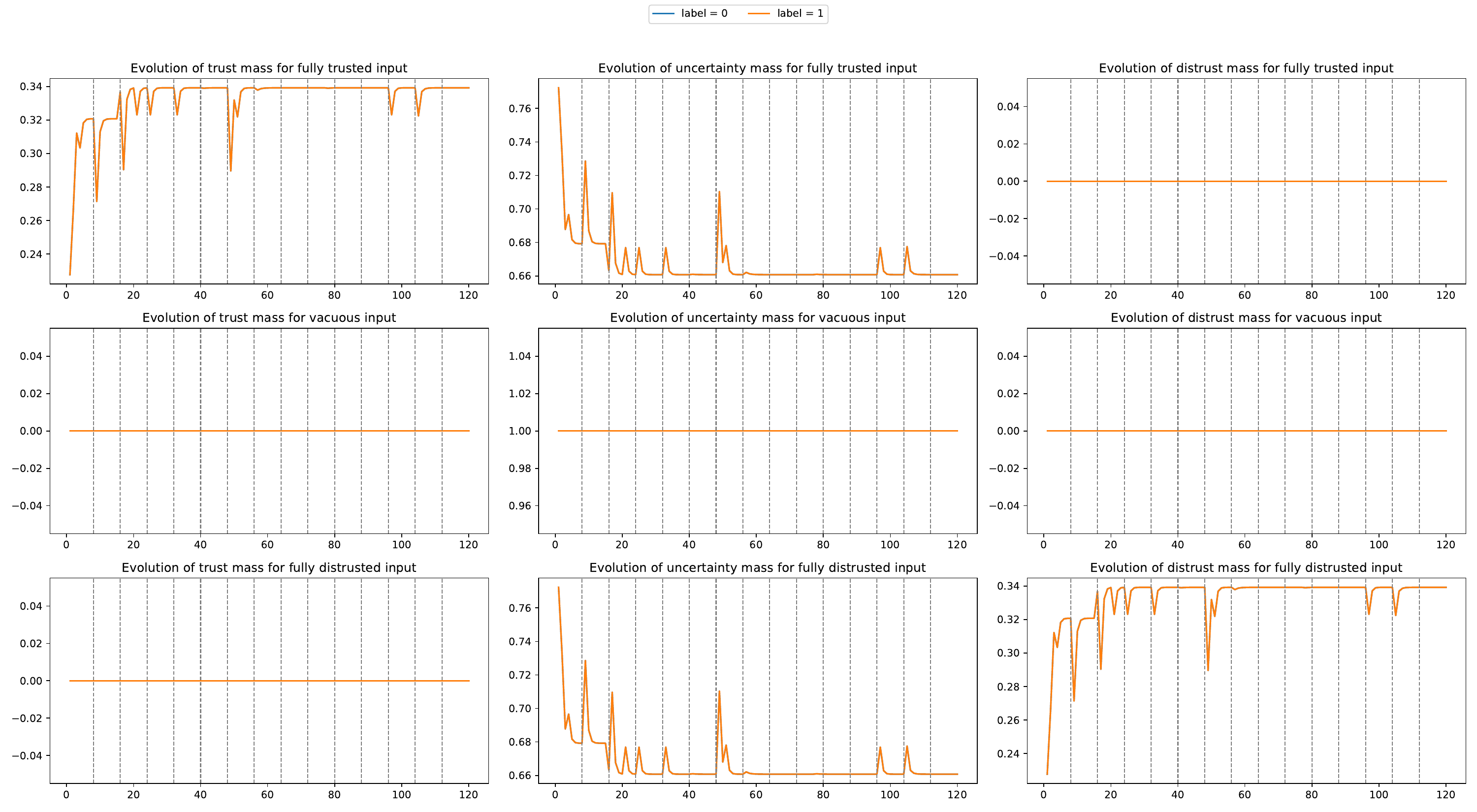}
  \caption{Features vacuous and labels trusted}
\end{figure}
\begin{figure}[H]
  \includegraphics[width=\textwidth]{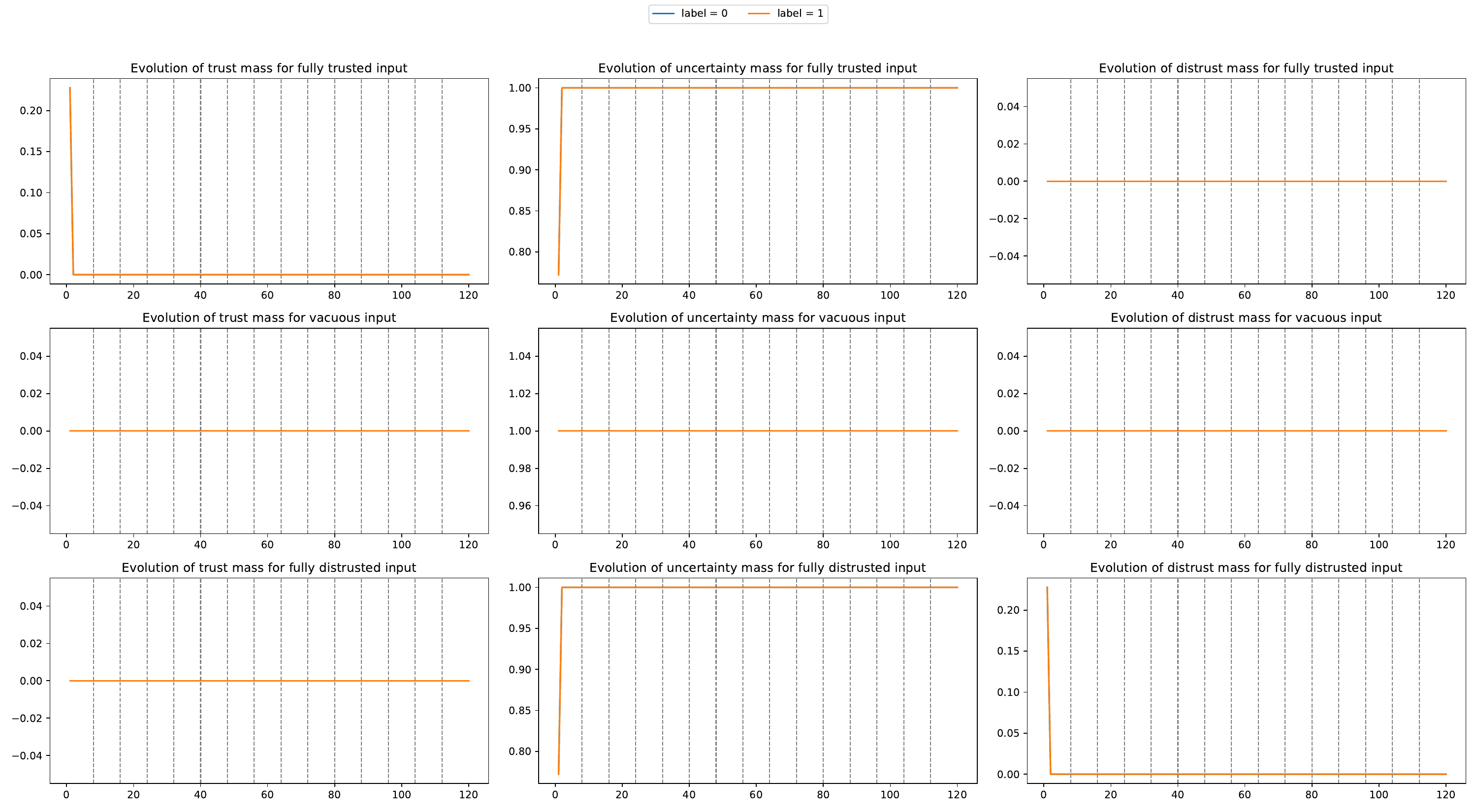}
  \caption{Features trusted and labels distrusted}
\end{figure}
\begin{figure}[H]
  \includegraphics[width=\textwidth]{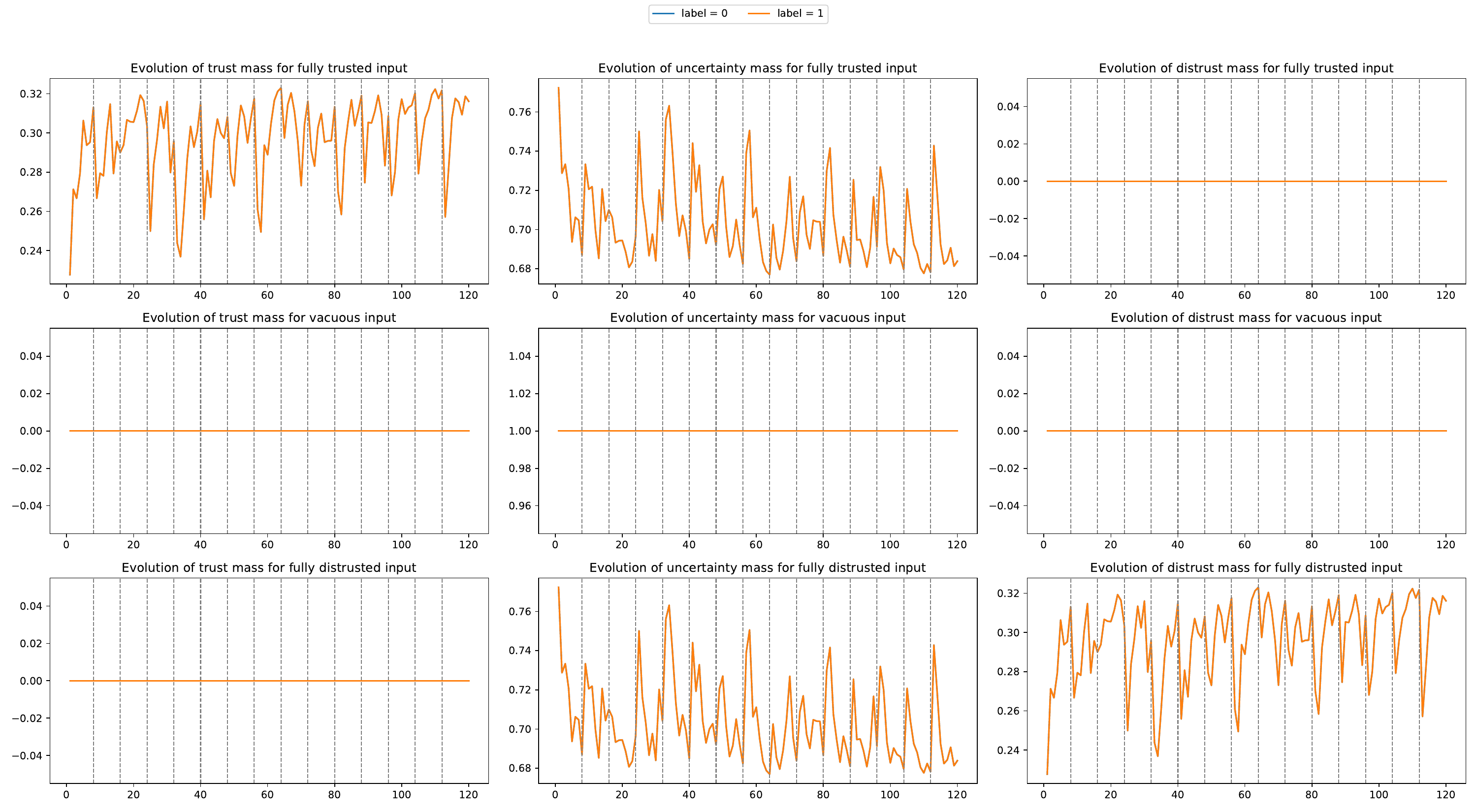}
  \caption{Features trusted and labels vacuous}
\end{figure}
\begin{figure}[H]
  \includegraphics[width=\textwidth]{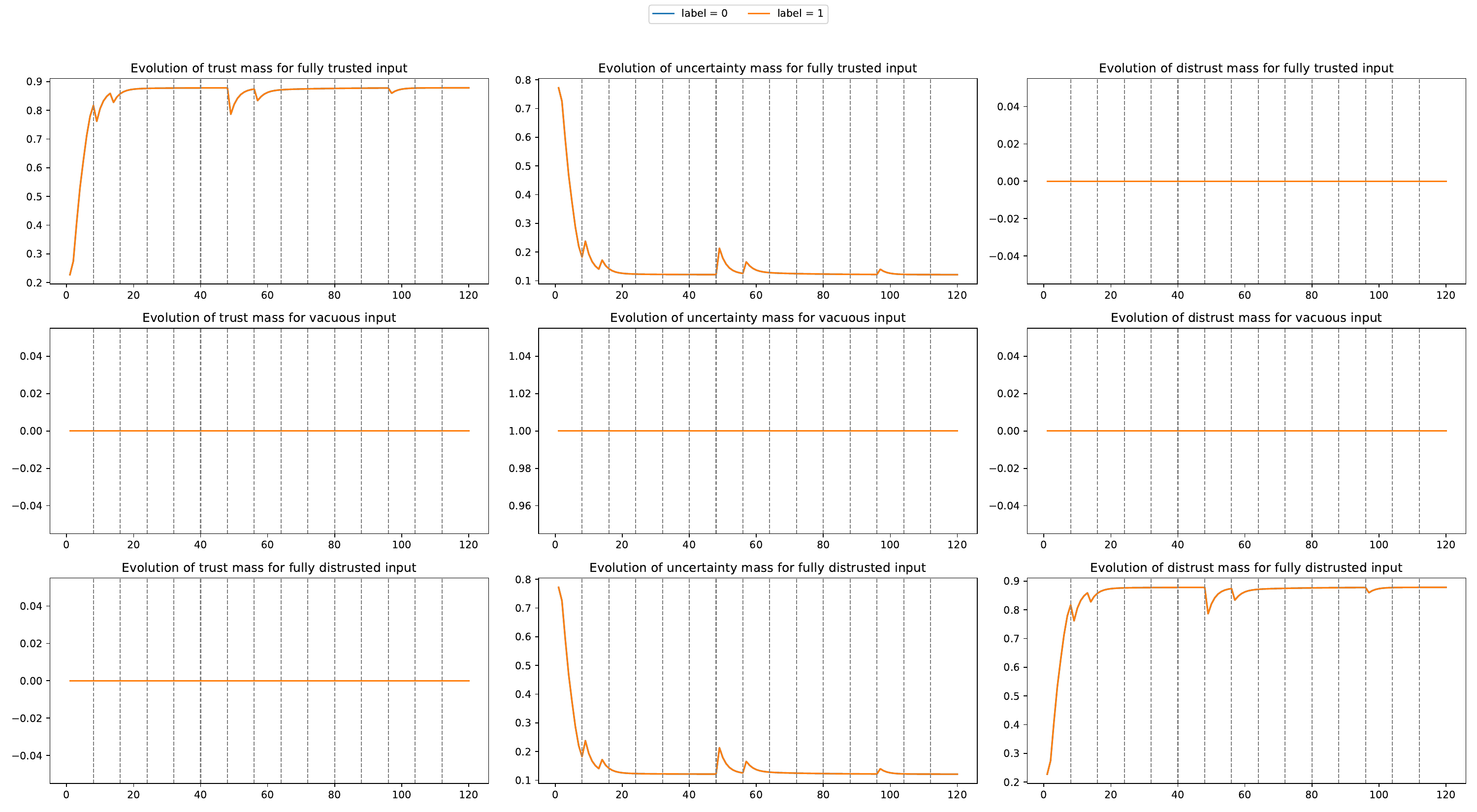}
  \caption{Features trusted and labels trusted}
\end{figure}

\subsection{Accuracy Evolution for the Cancer Model (\cref{exp:1})}\label{sec:resmnistaccapp}
\begin{figure}[H]
\centering
\begin{subfigure}[b]{0.32\textwidth}
  \includegraphics[width=\textwidth]{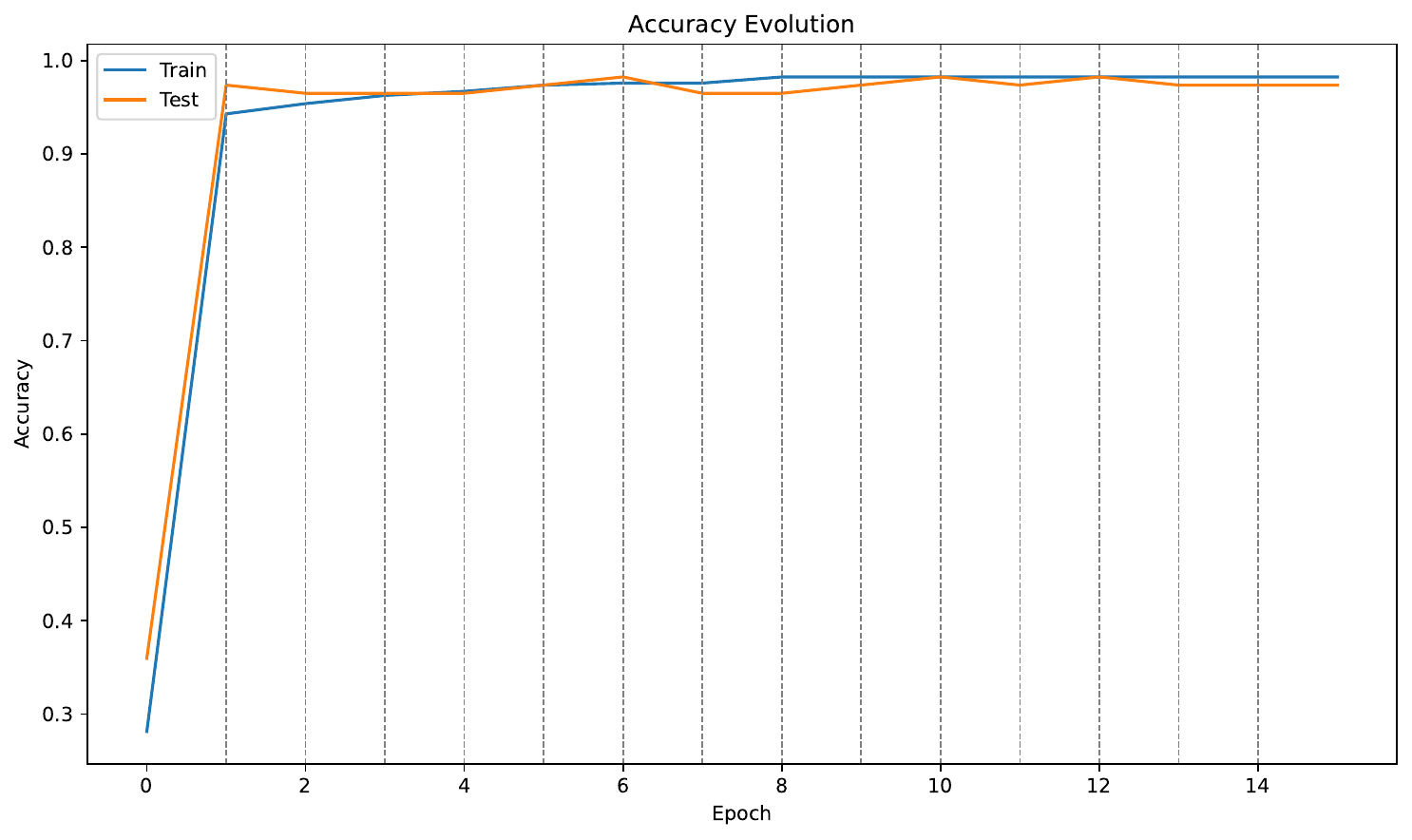}
  \caption{Clean Features and Labels}
\end{subfigure}
\begin{subfigure}[b]{0.32\textwidth}
  \includegraphics[width=\textwidth]{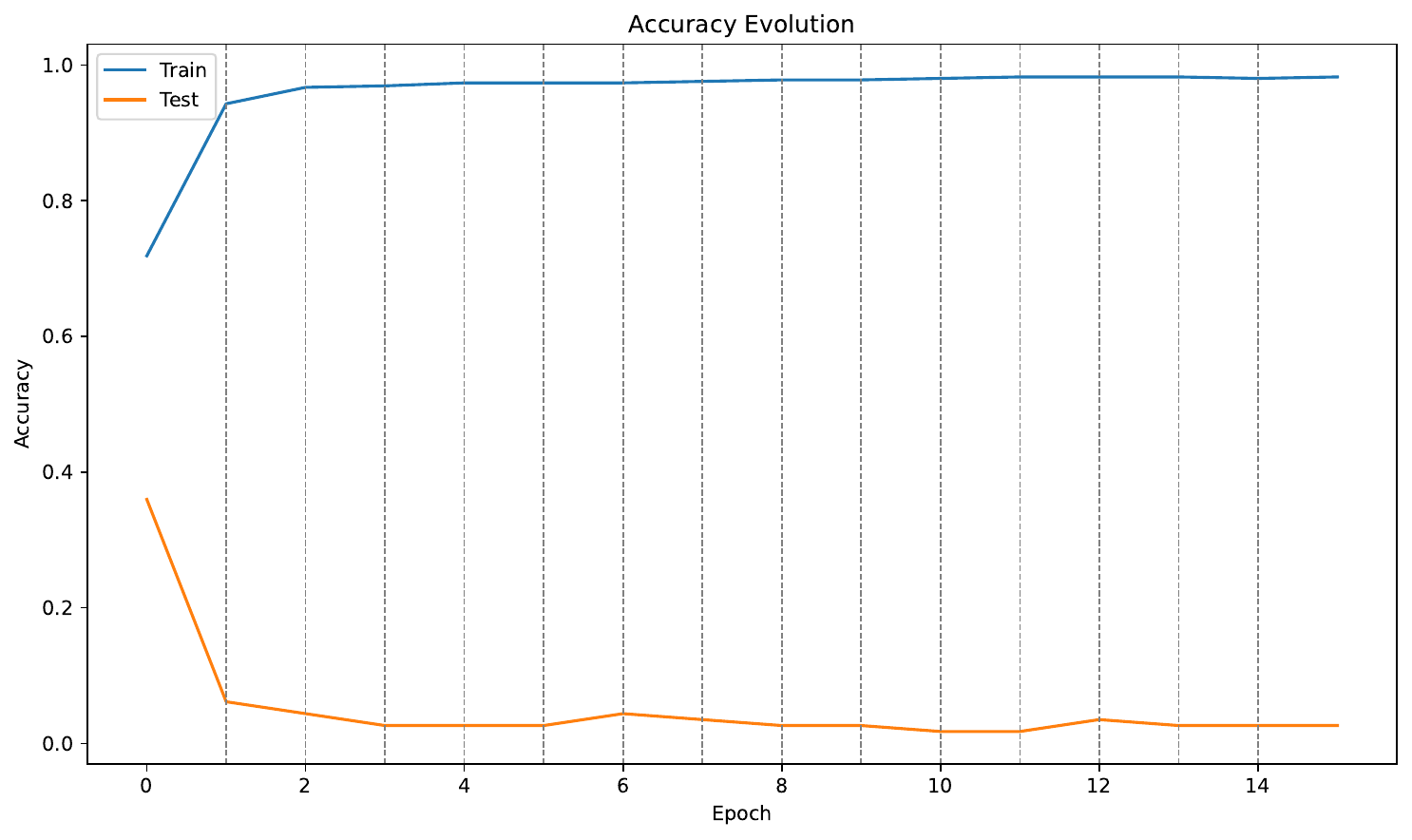}
  \caption{Clean Features and Corrupted Labels}
\end{subfigure}
\begin{subfigure}[b]{0.32\textwidth}
  \includegraphics[width=\textwidth]{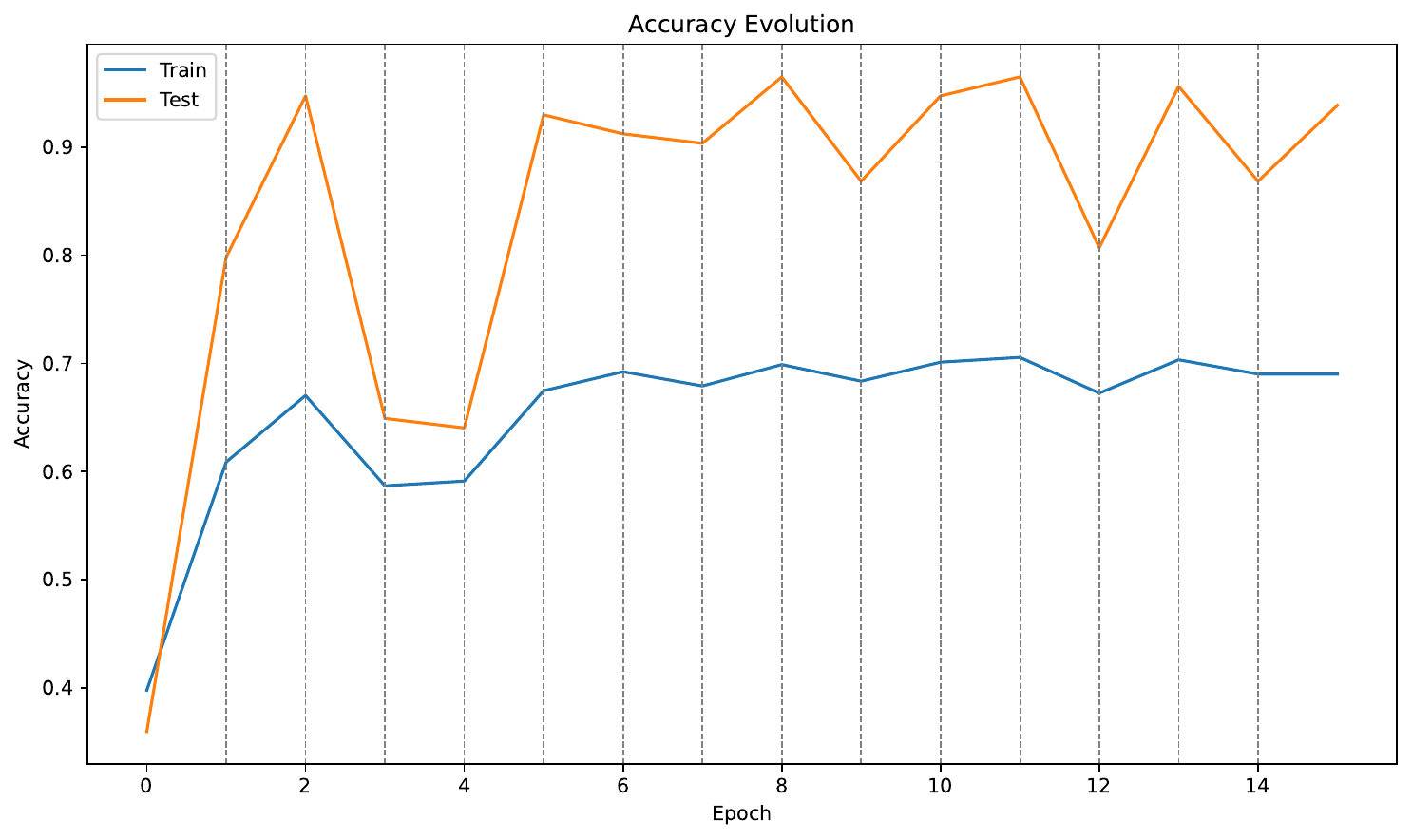}
  \caption{Clean Features and Noisy Labels}
\end{subfigure}
\begin{subfigure}[b]{0.32\textwidth}
  \includegraphics[width=\textwidth]{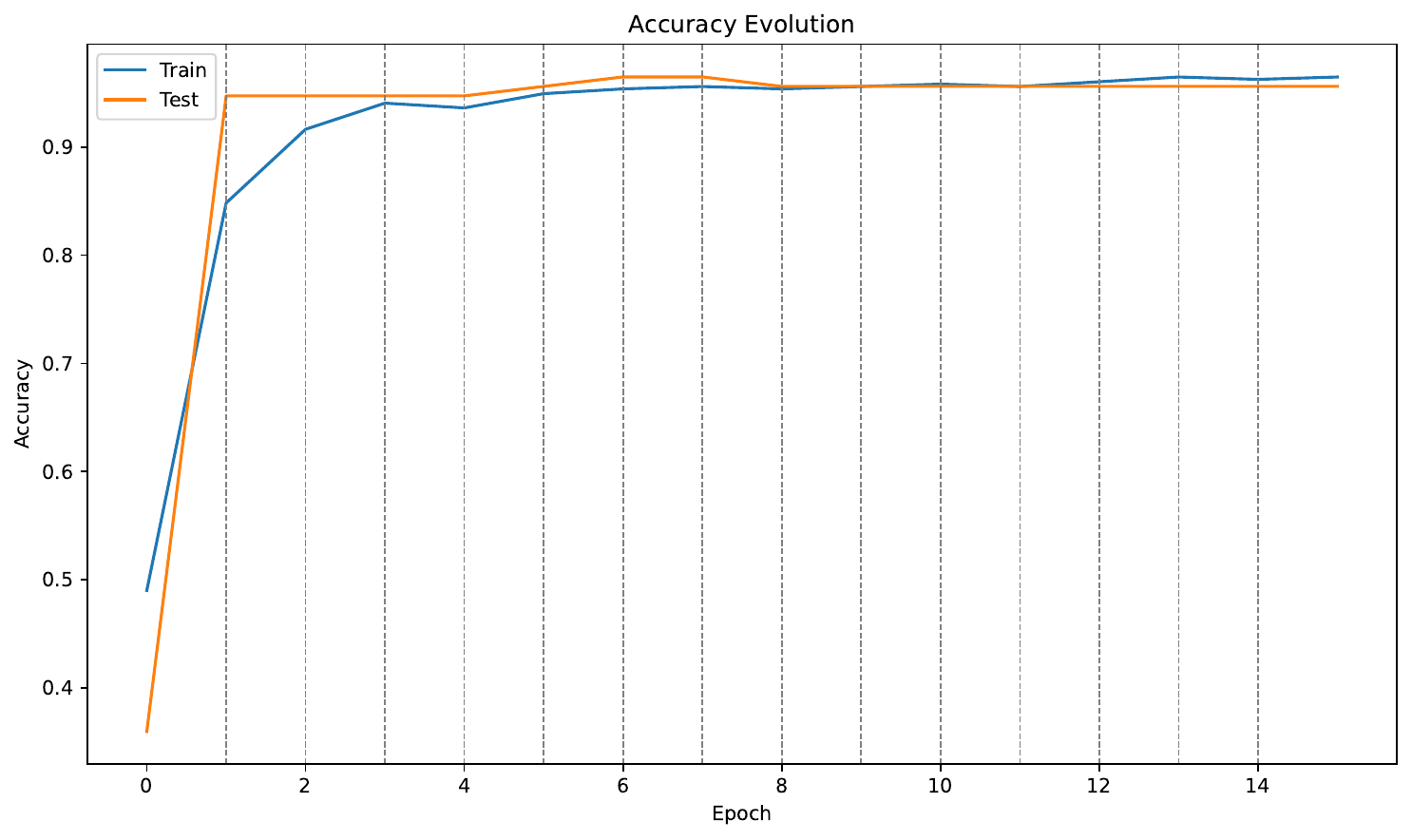}
  \caption{Noisy Features and Clean Labels}
\end{subfigure}
\begin{subfigure}[b]{0.32\textwidth}
  \includegraphics[width=\textwidth]{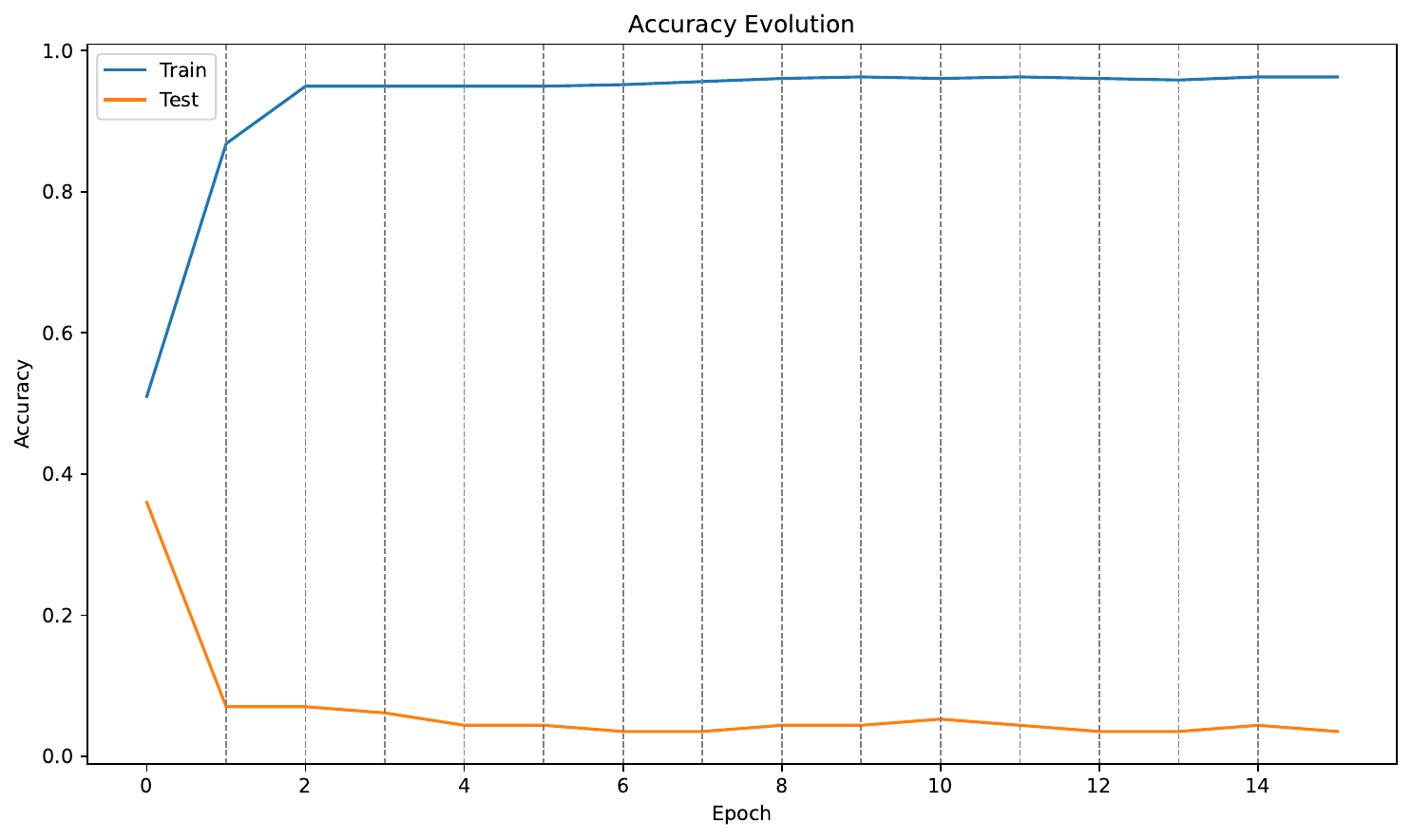}
  \caption{Noisy Features and Corrupted Labels}
\end{subfigure}
\begin{subfigure}[b]{0.32\textwidth}
  \includegraphics[width=\textwidth]{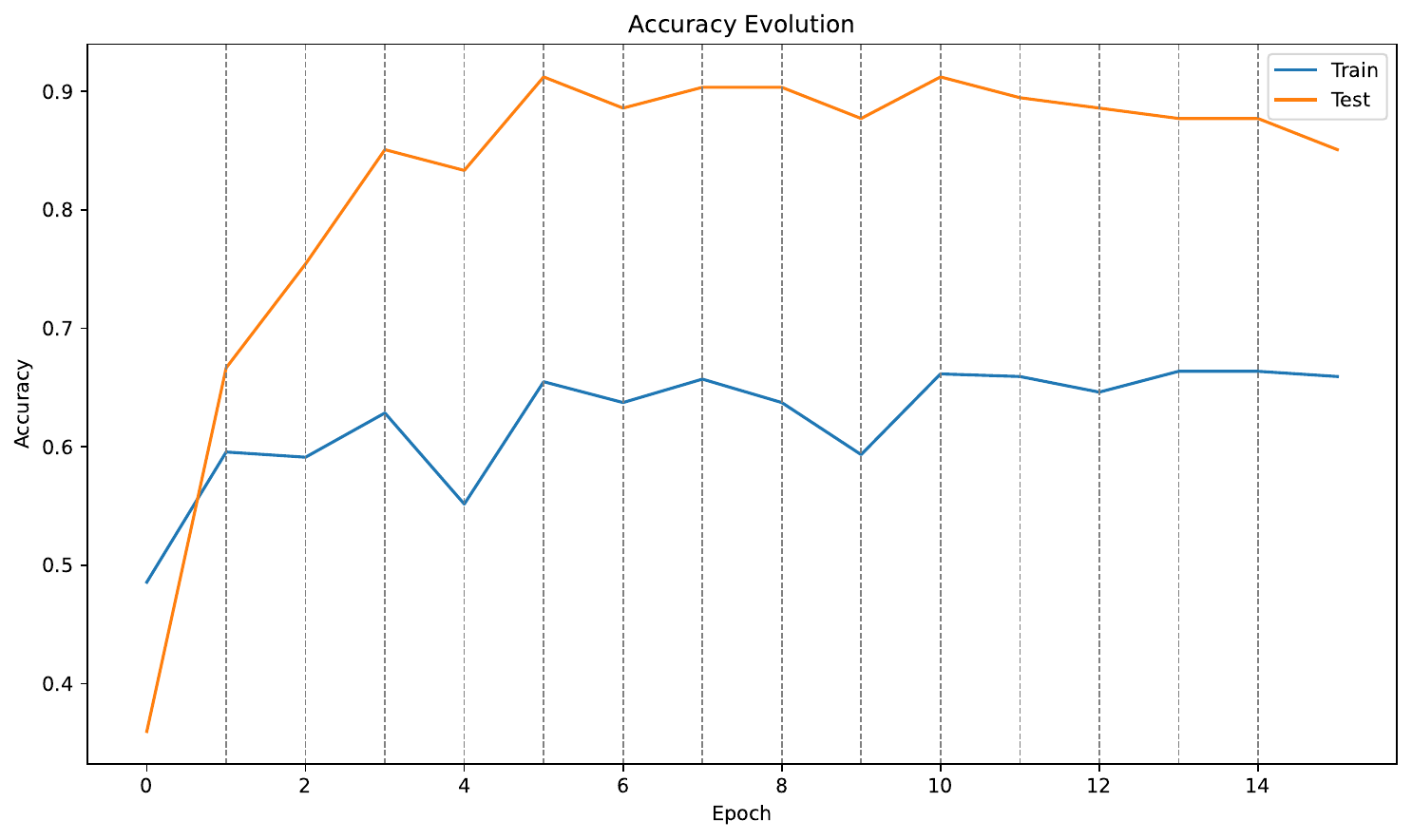}
  \caption{Noisy Features and Noisy Labels}
\end{subfigure}
\begin{subfigure}[b]{0.32\textwidth}
  \includegraphics[width=\textwidth]{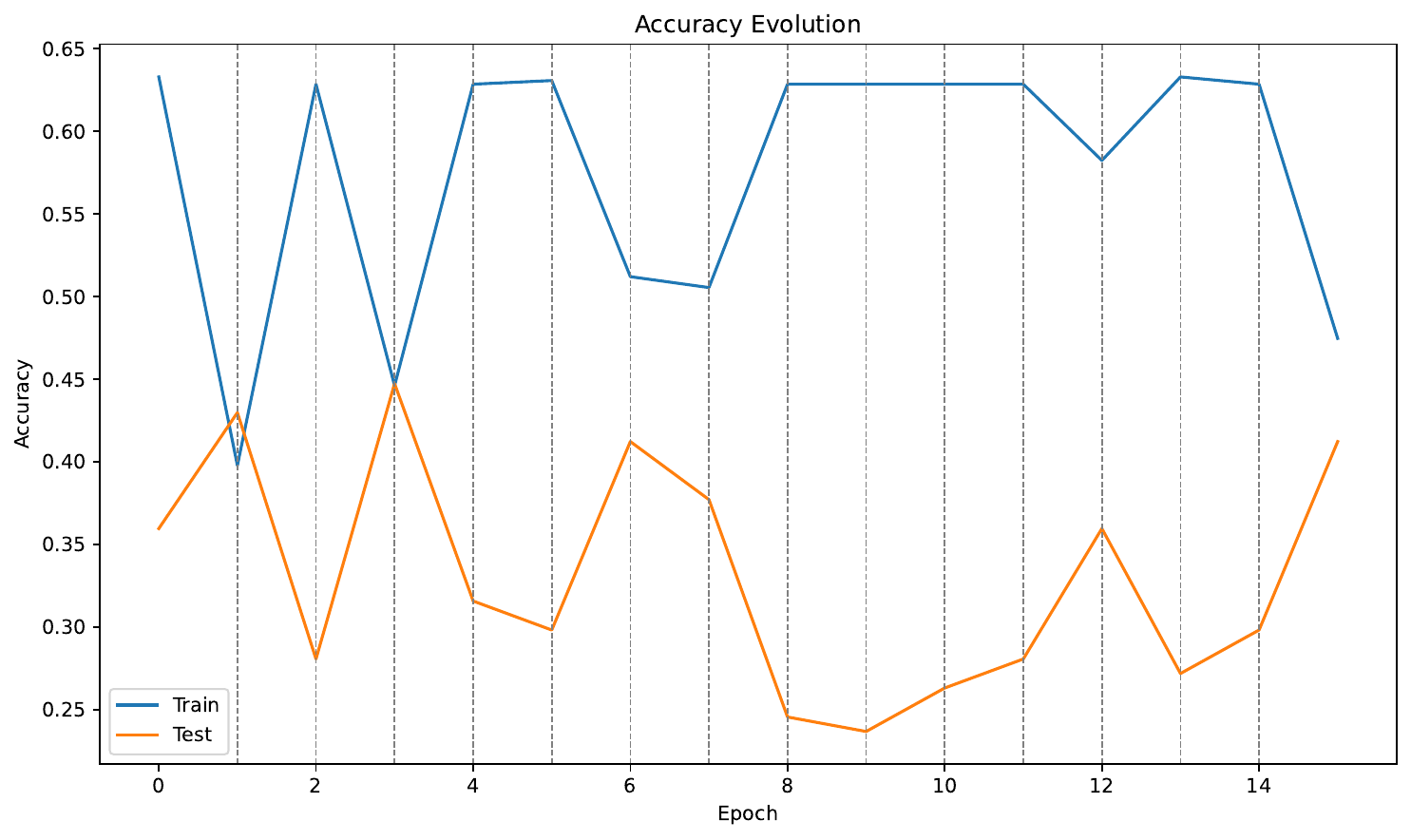}
  \caption{Corrupted Features and Clean Labels}
\end{subfigure}
\begin{subfigure}[b]{0.32\textwidth}
  \includegraphics[width=\textwidth]{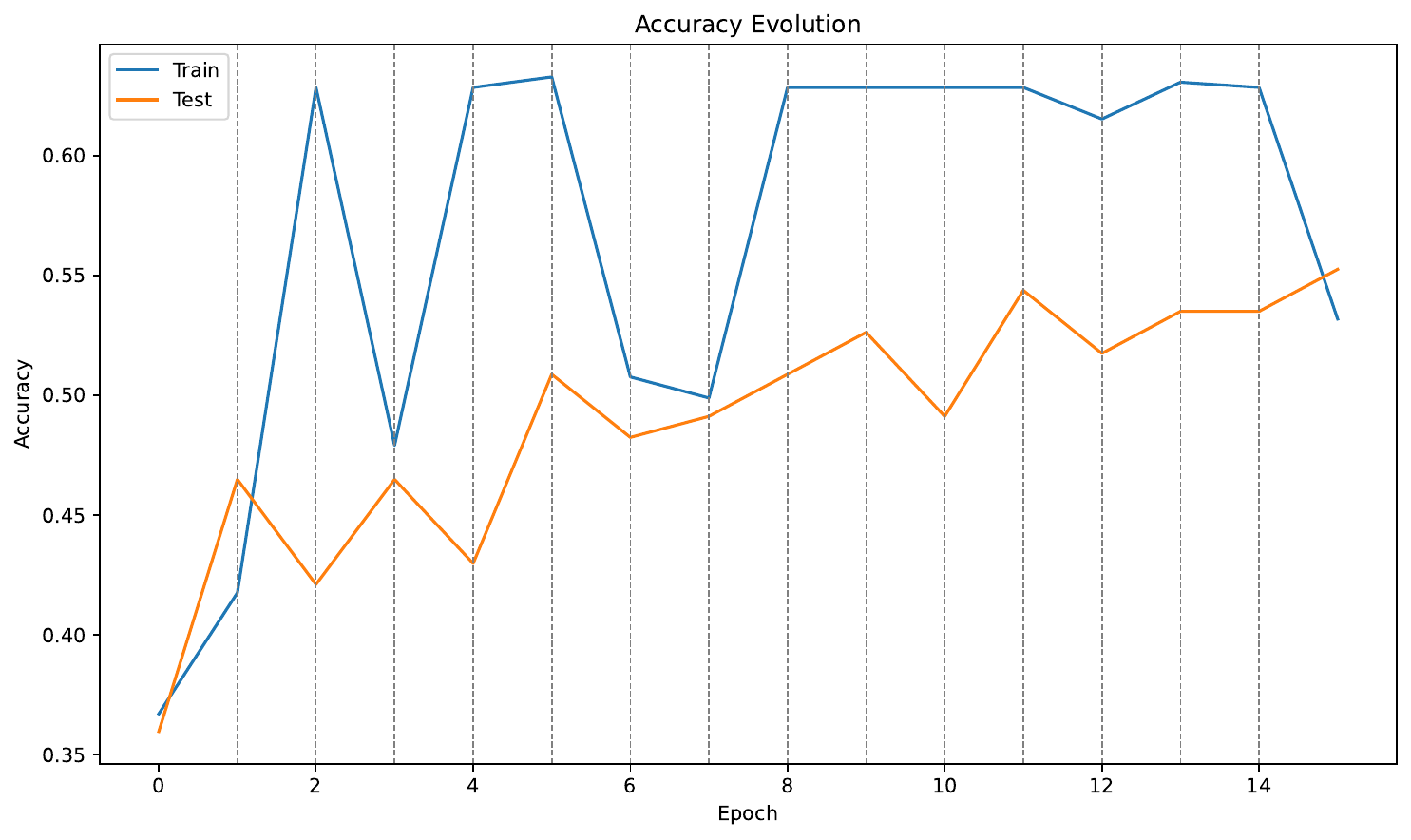}
  \caption{Corrupted Features and Corrupted Labels}
\end{subfigure}
\begin{subfigure}[b]{0.32\textwidth}
  \includegraphics[width=\textwidth]{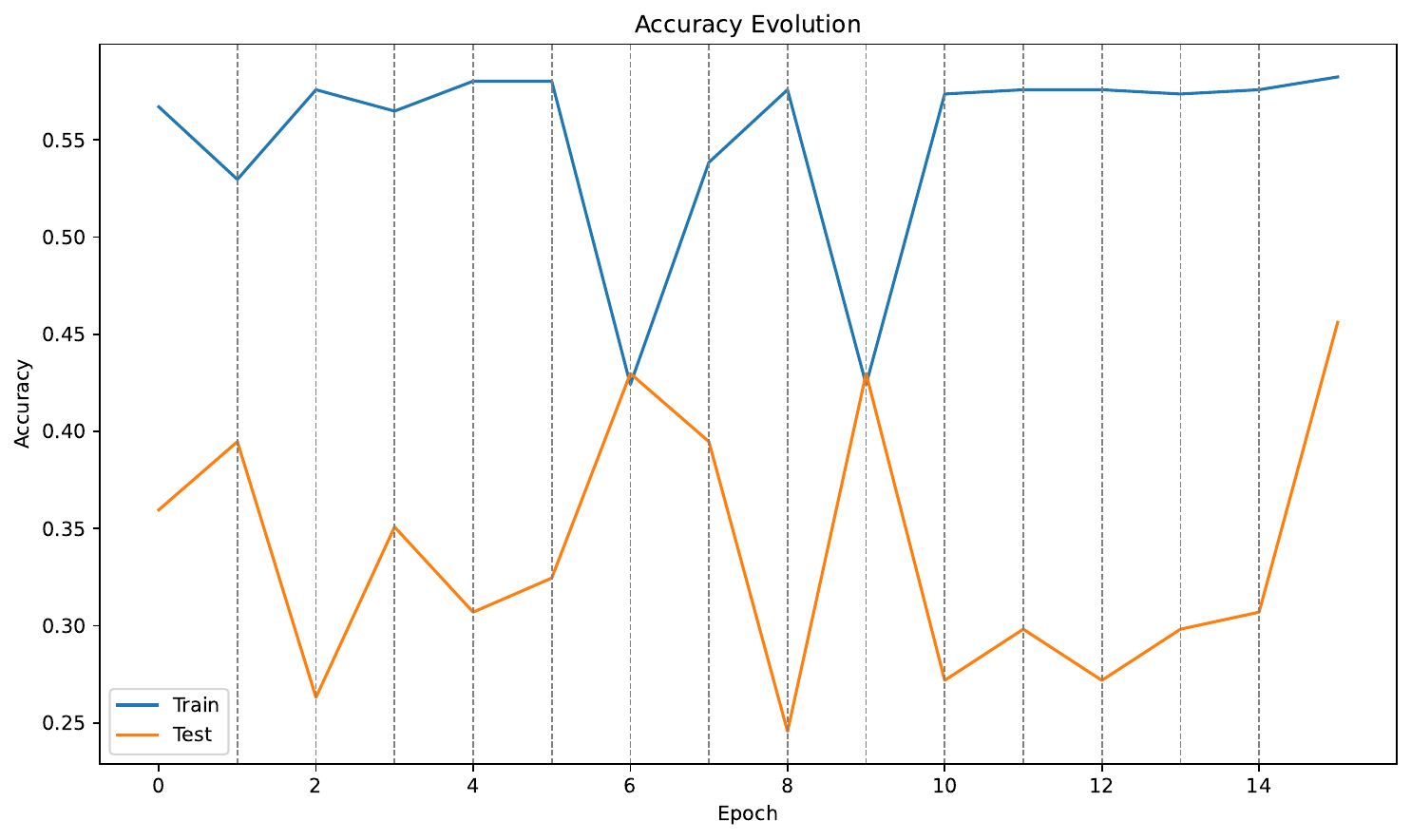}
  \caption{Corrupted Features and Noisy Labels}
\end{subfigure}
\caption{Accuracy evolution of the Cancer model under different combinations of clean, corrupted, and noisy features and labels.}
\label{fig:Acccancer}
\end{figure}
% \begin{figure}[H]
%     \centering
%     \includegraphics[width=0.8\linewidth]{good_plots/accuracy/Cancer/0.15-0.25/Cancer-noise-noise.pdf}
%     \caption{Accuracy evolution of the Cancer model when reducing the noise for noised features and labels.}
%     \label{fig:AcccancerInt}
% \end{figure}

\subsection{MNIST with Vacuous Trust Assessment (\cref{exp:2})}\label{sec:resmnist}
\begin{figure}[H]
  \centering
  \includegraphics[width=\textwidth]{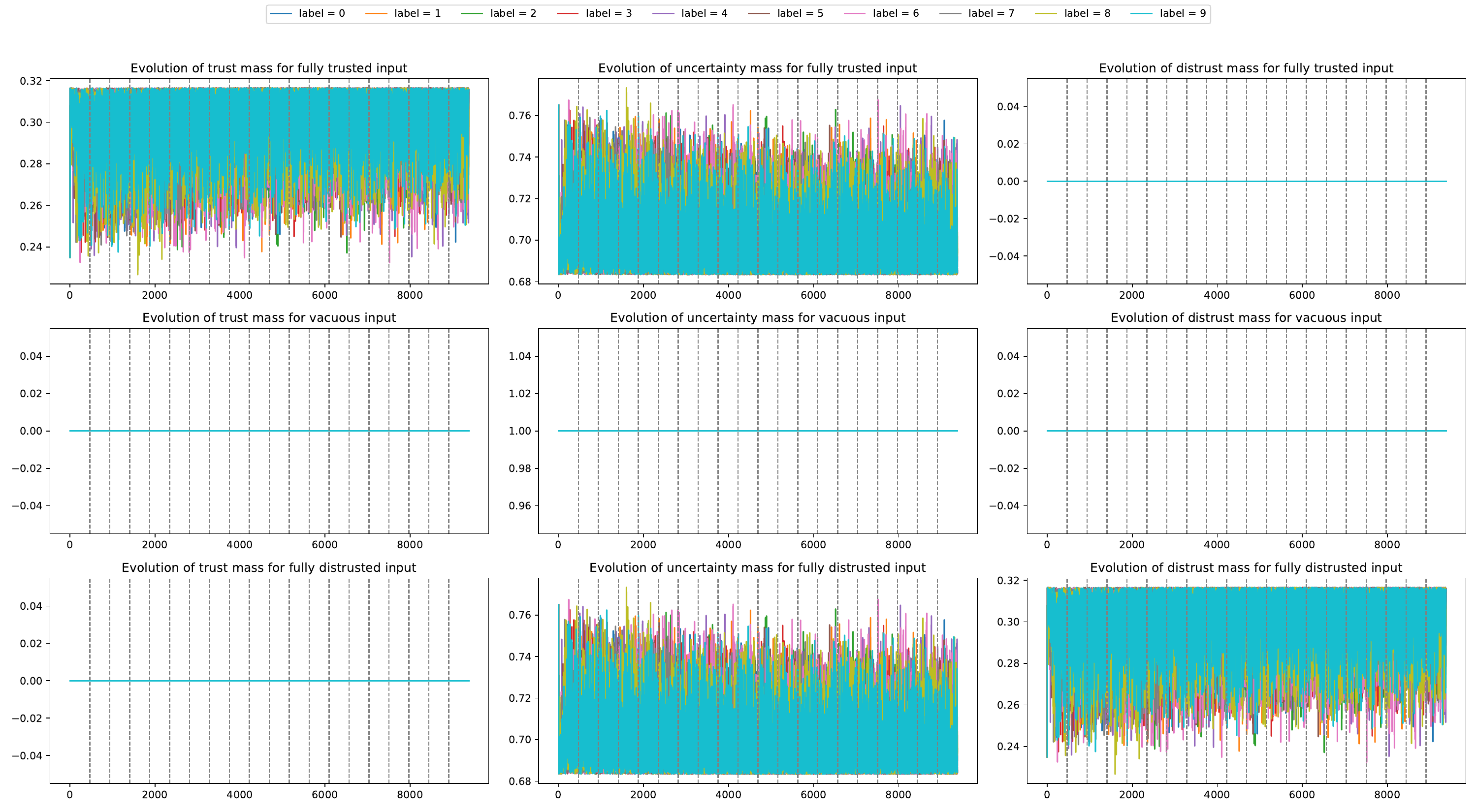}
  \caption{16 Hidden neurons}
\end{figure}
\begin{figure}[H]
  \centering
  \includegraphics[width=\textwidth]{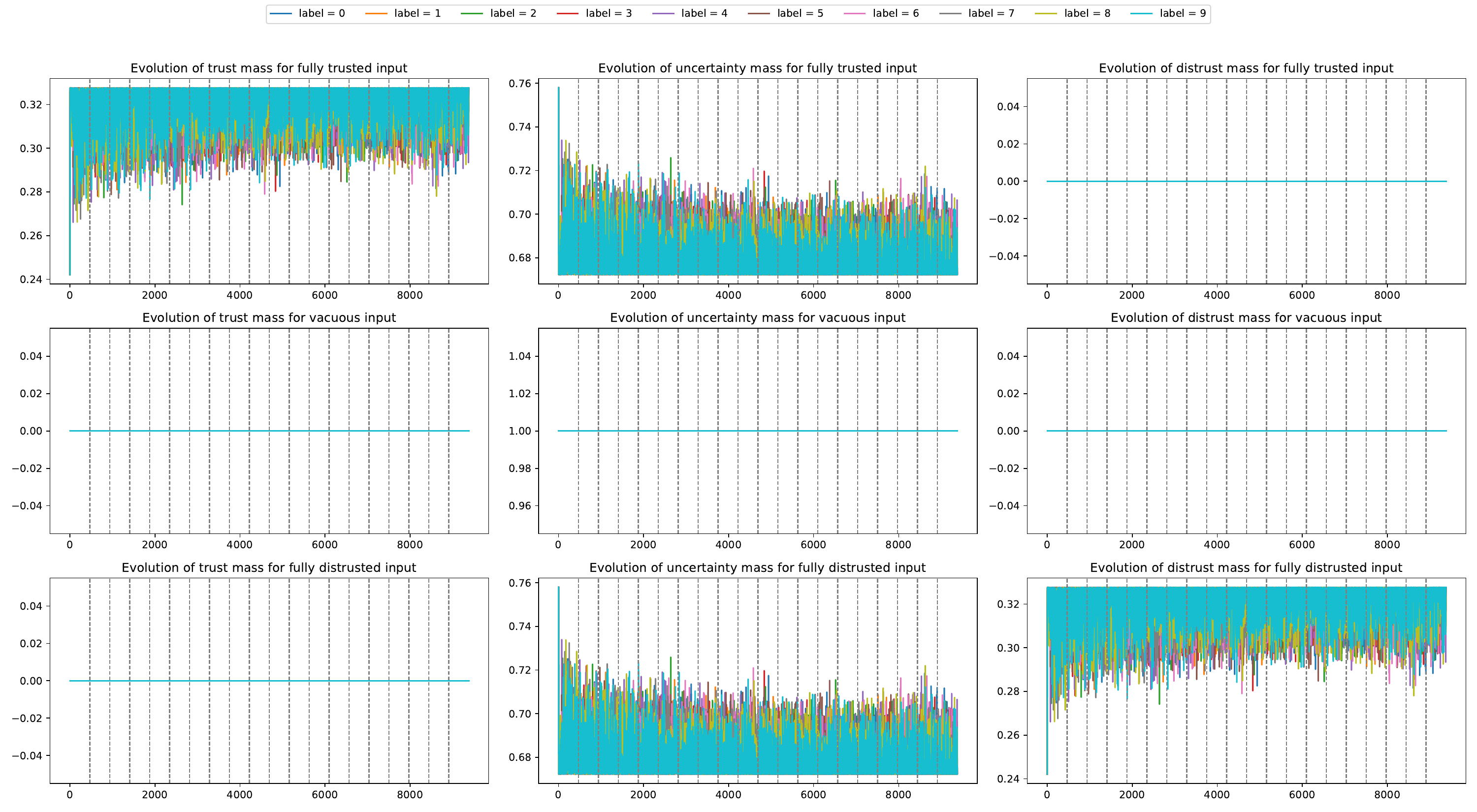}
  \caption{32 Hidden neurons}
\end{figure}
\begin{figure}[H]
  \centering
  \includegraphics[width=\textwidth]{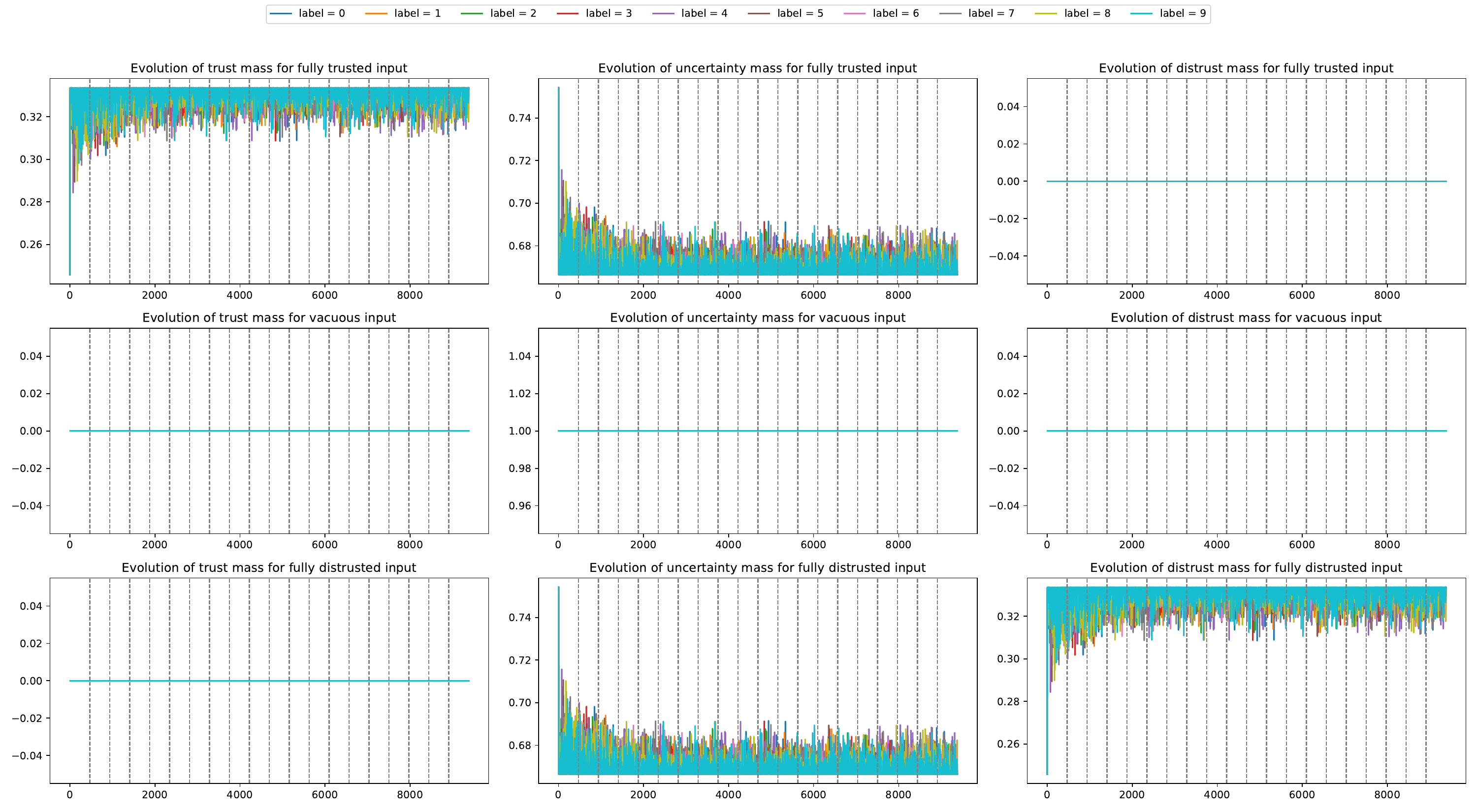}
  \caption{64 Hidden neurons}
\end{figure}
\begin{figure}[H]
  \centering
  \includegraphics[width=\textwidth]{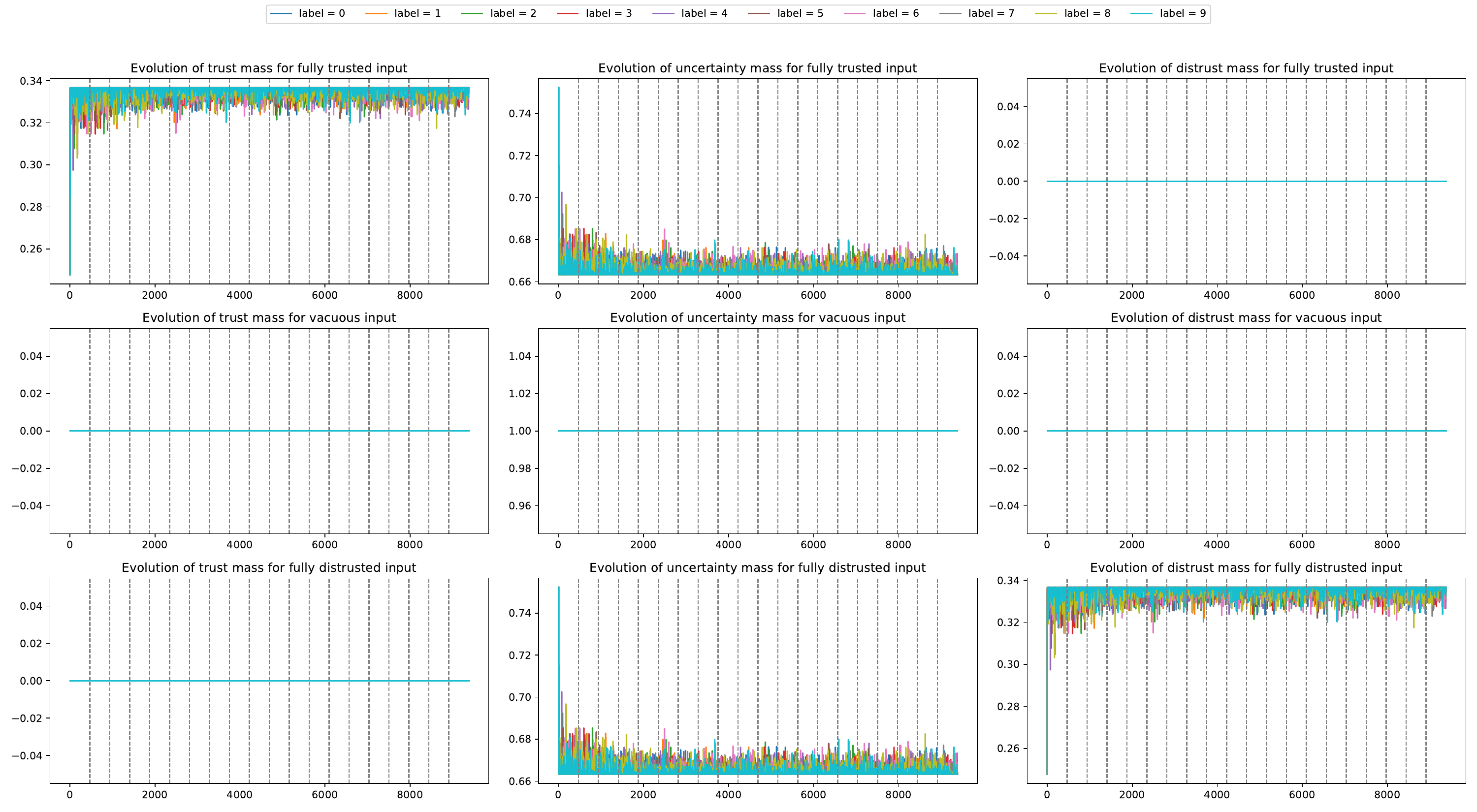}
  \caption{128 Hidden neurons}
\end{figure}
\begin{figure}[H]
  \centering
  \includegraphics[width=\textwidth]{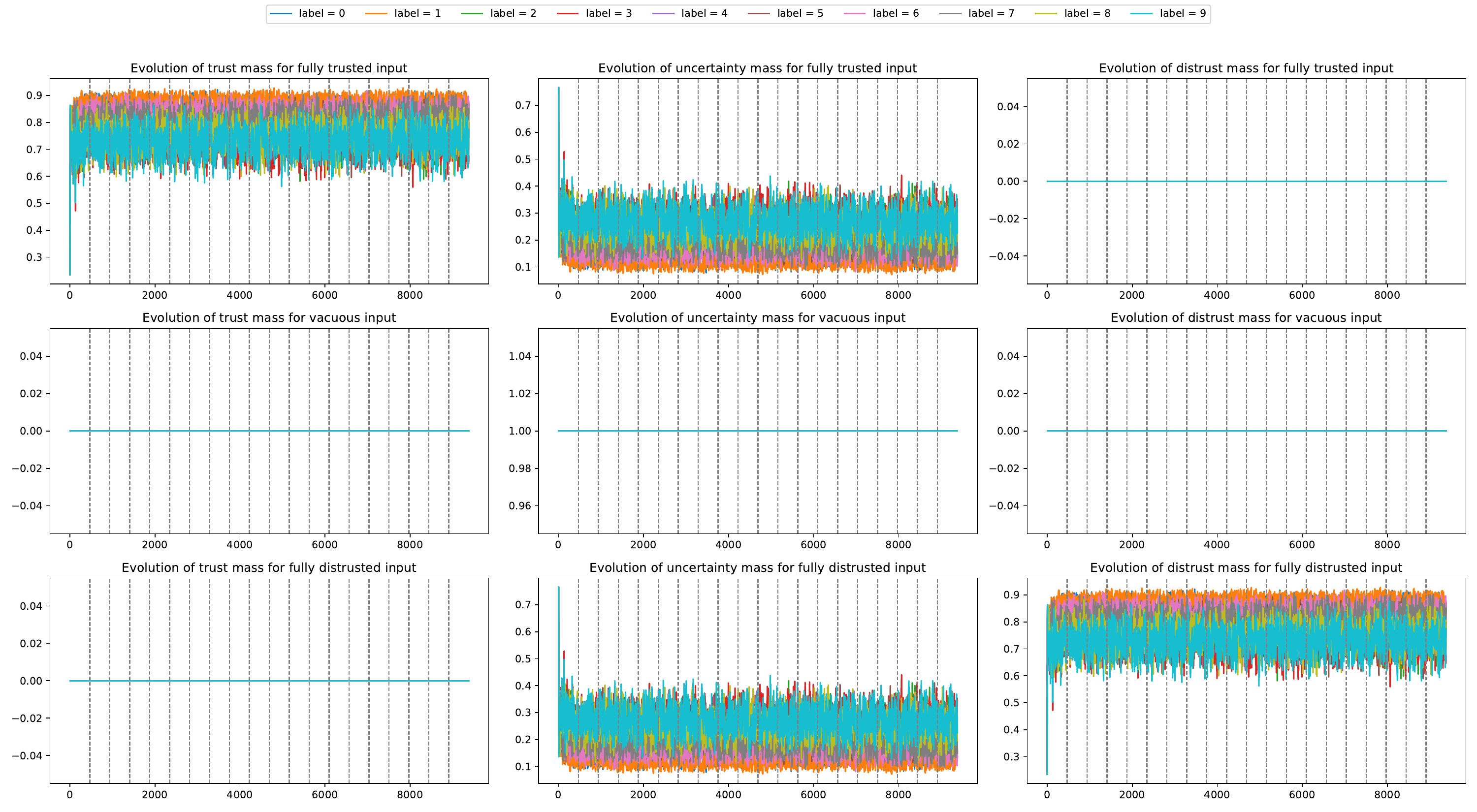}
  \caption{16 Hidden neurons with Fully Trusted Assessment}
\end{figure}

\subsection{MNIST poisoned 128 hidden neurons (\cref{exp:3})}\label{sec:resmnistpois}
\begin{figure}[H]
  \centering
  \includegraphics[width=\textwidth]{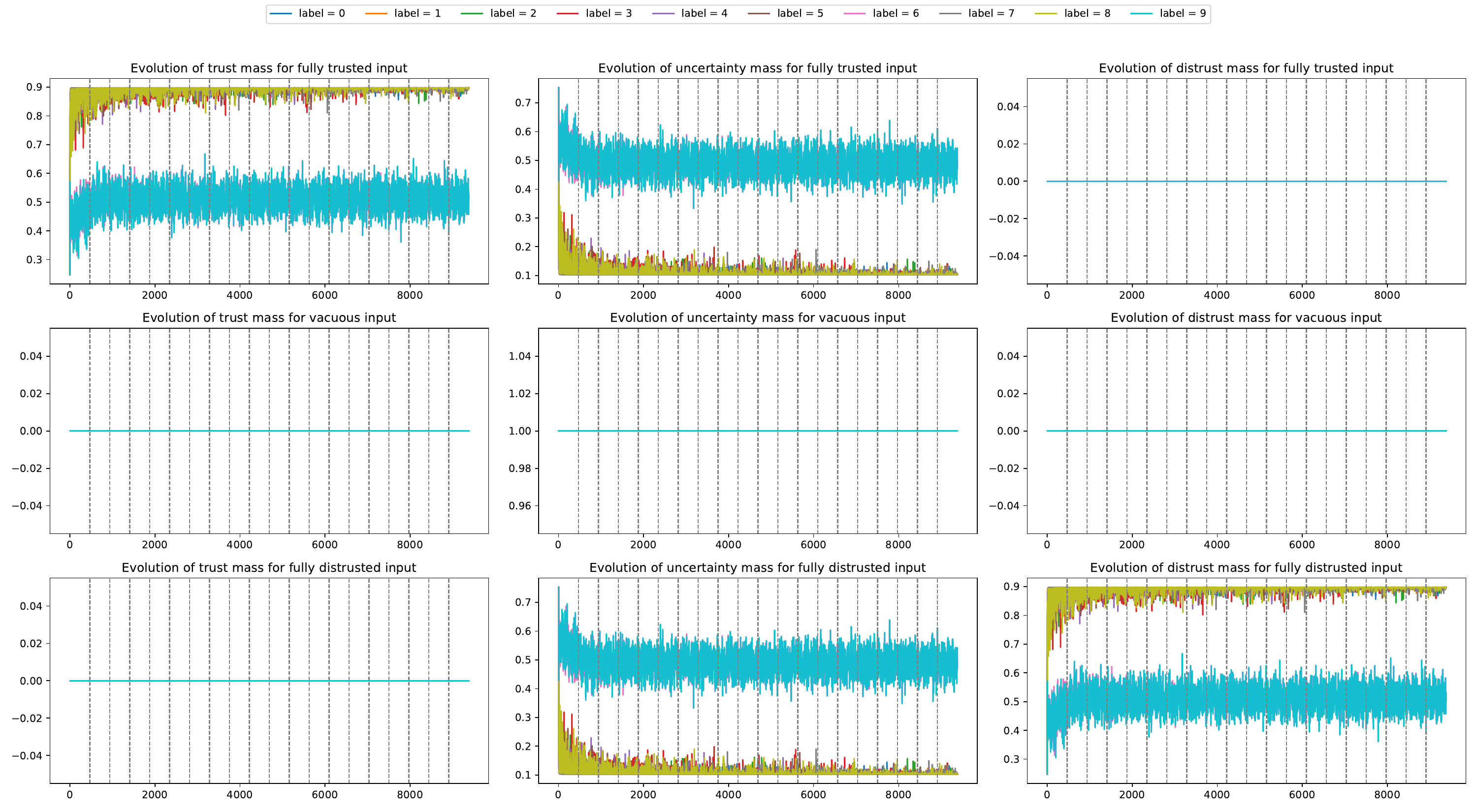}
  \caption{1 pixel}
\end{figure}
\begin{figure}[H]
  \centering
  \includegraphics[width=\textwidth]{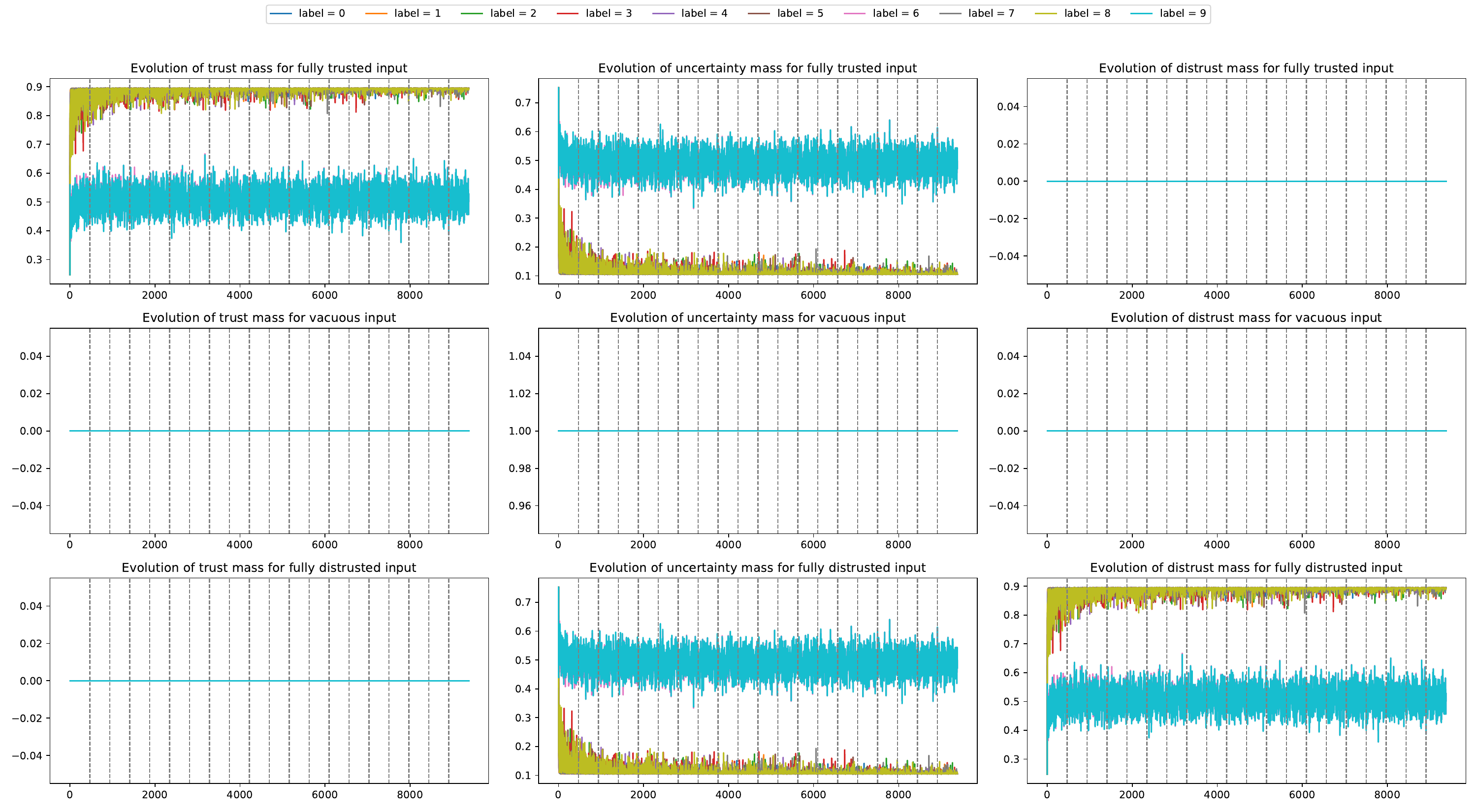}
  \caption{4×4 pixels}
\end{figure}
\begin{figure}[H]
  \centering
  \includegraphics[width=\textwidth]{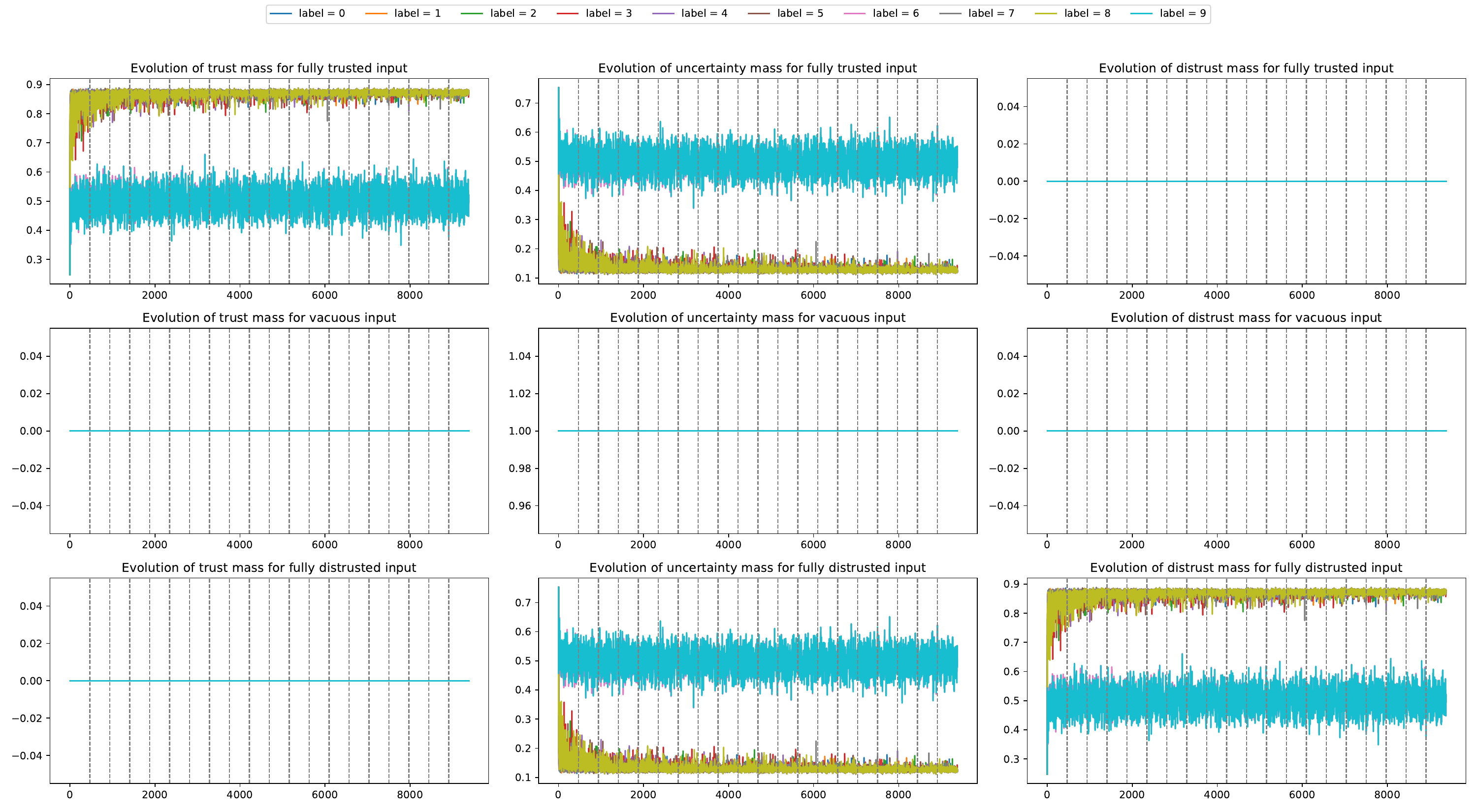}
  \caption{10×10 pixels}
\end{figure}
\begin{figure}[H]
  \centering
  \includegraphics[width=\textwidth]{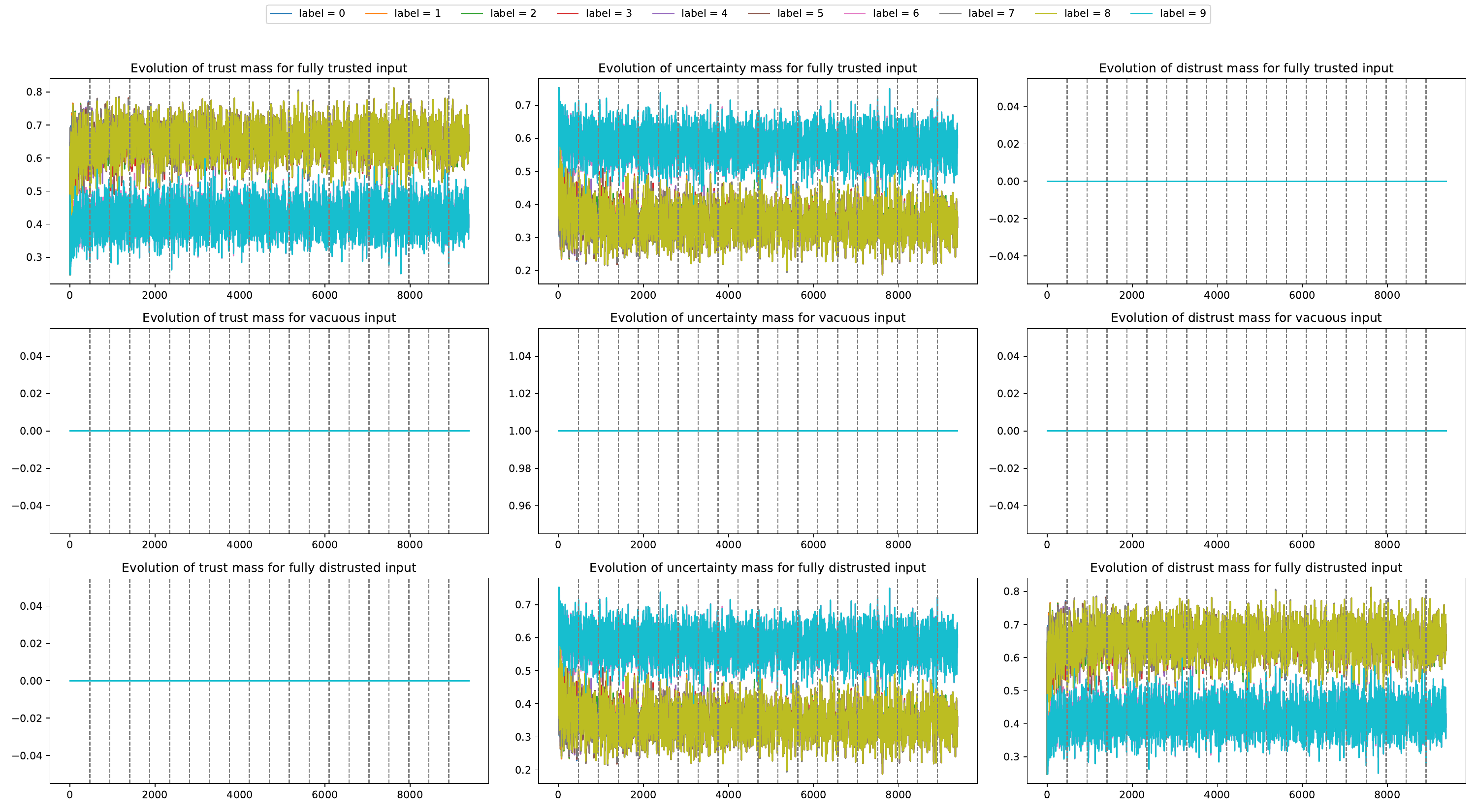}
  \caption{27×27 pixels}
\end{figure}
\subsection{Plots for the Comparative Trust Evaluation (\cref{exp:4})}\label{sec:resexp4}
\begin{figure}[H]
  \centering
  \includegraphics[width=0.62\textwidth]{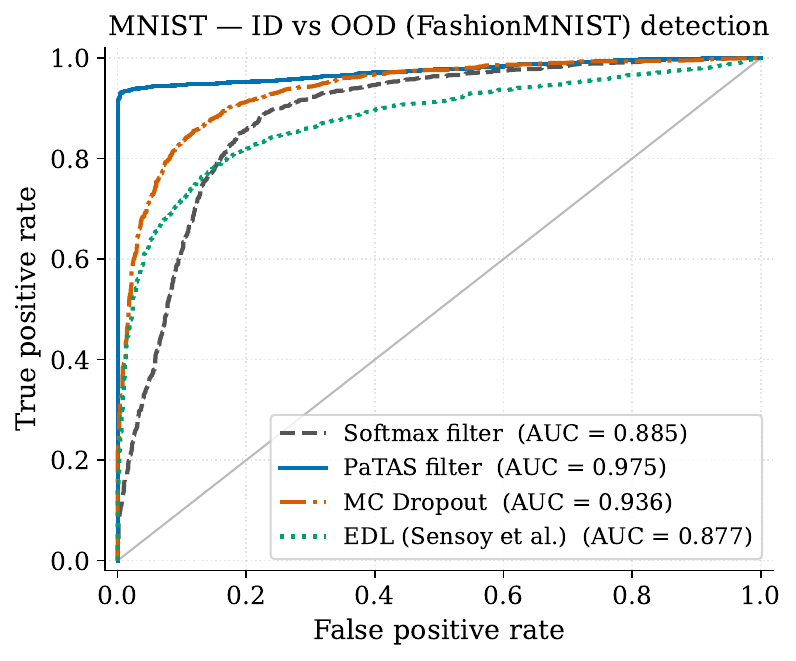}
  \caption{Out-of-distribution detection ROC (MNIST in-distribution vs.\ Fashion-MNIST OOD): each method's own score as the discriminator. Curves and legend AUC values are from a single representative evaluation seed; \cref{tab:comparative} reports mean $\pm$ std over three seeds.}
\end{figure}
\begin{figure}[H]
  \centering
  \includegraphics[width=0.62\textwidth]{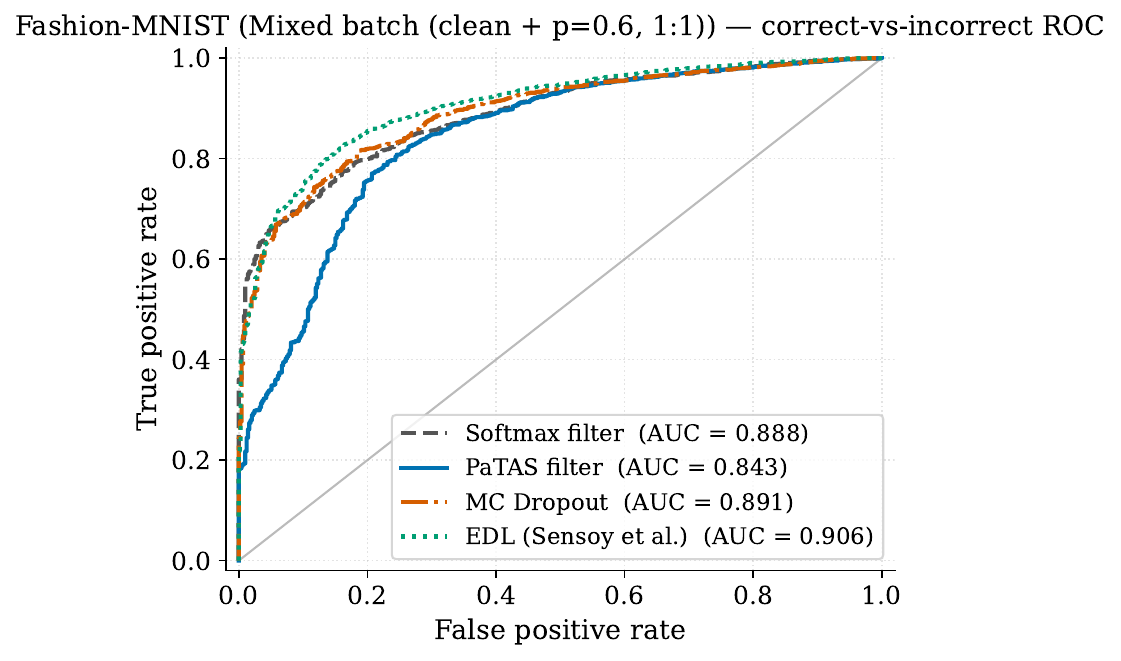}
  \caption{Correct-vs-incorrect ROC on the Fashion-MNIST mixed batch (clean + corrupted inputs pooled 1:1), single representative evaluation seed.}
\end{figure}
\begin{figure}[H]
  \centering
  \includegraphics[width=0.62\textwidth]{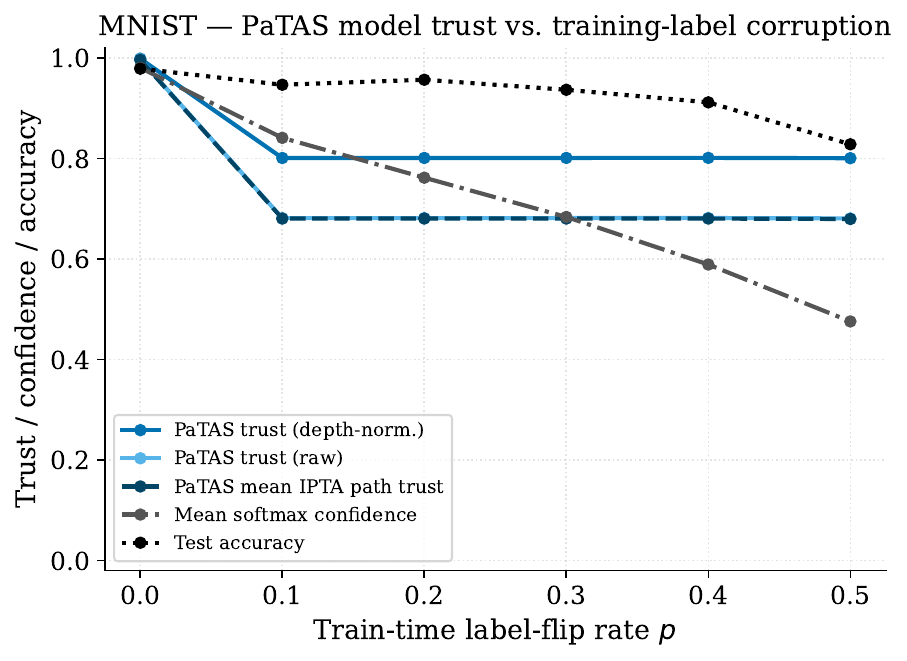}
  \caption{Training-quality audit (\cref{tab:audit}): PaTAS model-side trust, mean softmax confidence, and test accuracy as functions of the train-time label-flip rate $p$. The depth-normalized curve rescales the committed mass $b+d$ of the aggregated opinion to $(b+d)^{1/L}$, with $L$ the number of trust-propagation layers, making trust comparable across network depths. Trust drops as a tripwire at any non-zero rate without using test data; confidence tracks severity but requires clean held-out data.}
\end{figure}

\subsection{Example for random dataset trust assessment}\label{sec:resrand}

\begin{figure}[H]
\centering
\includegraphics[width=\textwidth]{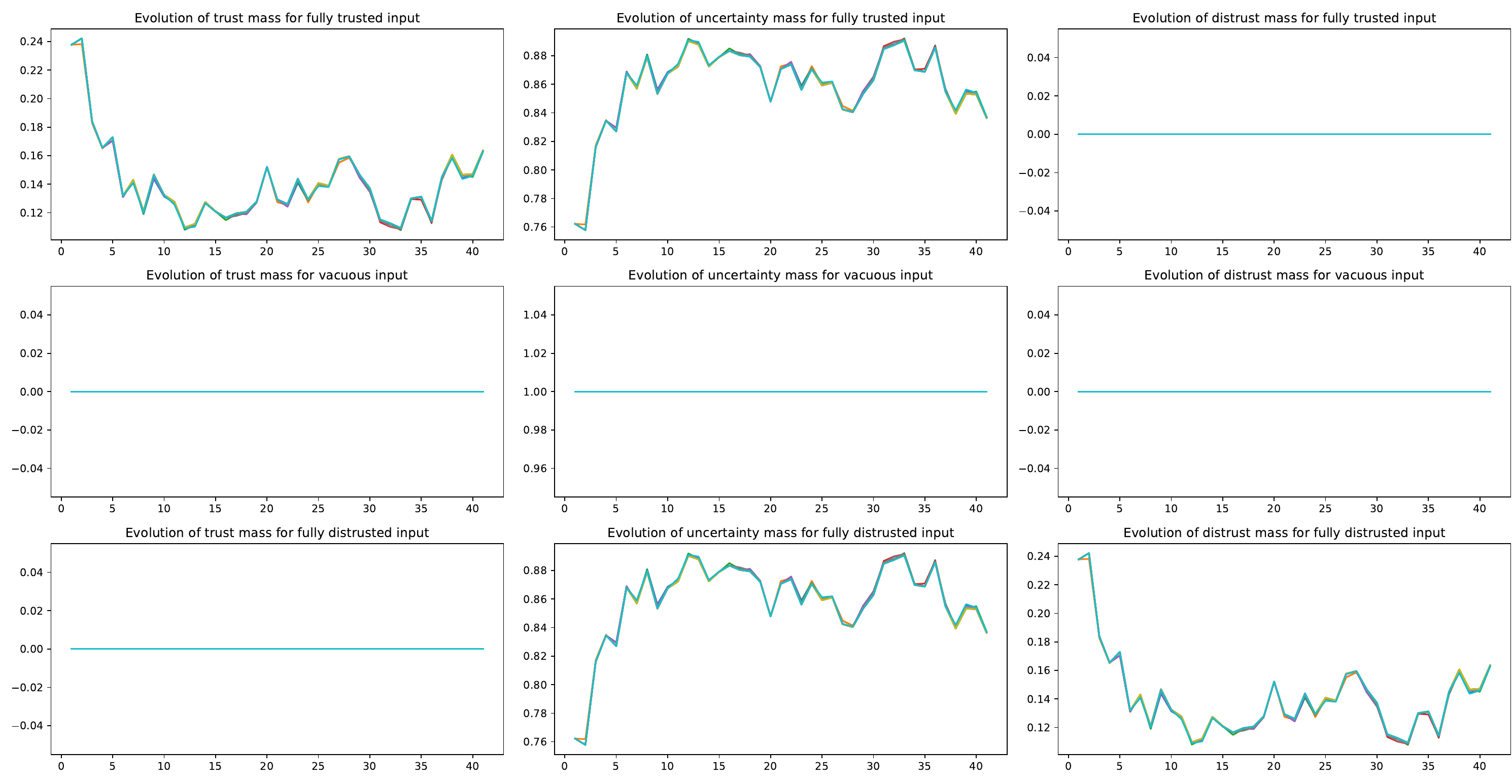}
\caption{MNIST randomized trust}
\label{fig:mnistrand}
\end{figure}

\fi

\end{document}